%% file: collas2022_conference.tex
\documentclass[table]{article} 
\usepackage{collas2022_conference,times}

\input{math_commands.tex}

\usepackage{hyperref}
\hypersetup{
    colorlinks=true,
    linkcolor=red,
    filecolor=magenta,      
    urlcolor=blue,
    citecolor=purple,
    pdftitle={Overleaf Example},
    pdfpagemode=FullScreen,
    }
\usepackage{times}
\usepackage{wrapfig}

\usepackage{epsfig}
\usepackage{graphicx}
\usepackage{amsmath}
\usepackage{amssymb}
\usepackage{subcaption}
\usepackage{multirow}
\usepackage{enumitem}
\usepackage{xcolor}
\usepackage{xspace}
\usepackage{soul}
\usepackage{tabularx}
\newcommand*{\transpose}{^\intercal}
\newcommand{\norm}[1]{\| #1 \|}
\newcommand{\myparagraph}[1]{\noindent\textbf{#1}}
\newcommand{\IN}{{\tt In}\xspace}
\newcommand{\Out}{{\tt{Out}}\xspace}
\newcommand{\F}{{\tt{Forg}}\xspace}
\newcommand{\mah}{\textit{Mahalanobis}\xspace}
\newcommand{\odin}{\textit{ODIN}\xspace}
\newcommand{\nav}{\textit{Softmax}\xspace}
\newcommand{\vae}{\textit{VAE}\xspace}
\newcommand{\fine}{{FineTune}\xspace}
\newcommand{\MAS}{{MAS}\xspace}
\newcommand{\OOD}{{OOD}\xspace}
\newcommand{\LwF}{{LwF}\xspace}
\newcommand{\ER}{{ER}}{\xspace}
\newcommand{\SSIL}{{SSIL}}{\xspace}
\newcommand{\tinyimg}{{TinyImageNet Seq.}\xspace}
\newcommand{\tasks}{{Eight-Task Seq.}\xspace}
\newcommand{\supp}{{ the Appendix}\xspace}
\newcommand{\Ba}{\textit{B1}\xspace}
\newcommand{\Bb}{\textit{B2}\xspace}
\makeatletter
\DeclareRobustCommand\onedot{\futurelet\@let@token\@onedot}
\def\@onedot{\ifx\@let@token.\else.\null\fi\xspace}
\def\latin{}
\def\eg{\latin{e.g}\onedot}

\def\ie{\latin{i.e}\onedot}

\def\vs{\latin{vs}\onedot}
\def\wrt{w.r.t\onedot}

\def\ND{\text{ND}}
\makeatother

\title{Continual Novelty Detection}

\author{Rahaf~Aljundi, Daniel~Olmeda~Reino, Nikolay~Chumerin   \\
Toyota Motor Europe \\
Belgium \\
\texttt{firstname.lastname@toyota-europe.com} \\
\And 
Richard E. Turner  \\
University of Cambridge \\
United Kingdom \\
\texttt{ret26@cam.ac.uk} \\
}

\collasfinalcopy 

\begin{document}

\maketitle
\input{sections/abstract.tex}
\input{sections/introduction.tex}
\input{sections/related_work.tex}
\input{sections/problem_description.tex}
\input{sections/evaluation.tex}
\input{sections/discussion.tex}
\input{sections/conclusion.tex}
\input{sections/ethics.tex}

\bibliography{collas2022_conference}
\bibliographystyle{collas2022_conference}

\appendix
\section{Appendix}
\input{sections/appendix.tex}

\end{document}

%% file: math_commands.tex
\usepackage{amsmath,amsfonts,bm}

\def\eqref#1{equation~\ref{#1}}

\def\1{\bm{1}}

\def\vs{{\bm{s}}}

\DeclareMathAlphabet{\mathsfit}{\encodingdefault}{\sfdefault}{m}{sl}
\SetMathAlphabet{\mathsfit}{bold}{\encodingdefault}{\sfdefault}{bx}{n}

\DeclareMathOperator*{\argmax}{arg\,max}

%% file: sections/abstract.tex
\begin{abstract}
\label{sec:abstract}
Novelty Detection methods identify samples that are not representative of a model's training set thereby flagging misleading predictions and bringing a greater flexibility and transparency at deployment time. However, research in this area has only considered Novelty Detection in the offline setting.  
Recently, there has been a growing realization in the machine learning community that applications demand a more flexible framework - Continual Learning -
where new batches of data representing new domains, new classes or new tasks become available at different points in time.
In this setting, Novelty Detection becomes more important, interesting and challenging.
This work identifies the crucial link between the two problems and
investigates the Novelty Detection problem under the Continual Learning setting. We formulate the Continual Novelty Detection problem and present a benchmark, where we compare several Novelty Detection methods under different Continual Learning settings. 
 We show that Continual Learning affects the behaviour of novelty detection algorithms, while novelty detection can pinpoint insights in the behaviour of a continual learner. We further propose baselines and discuss possible research directions. We believe that the coupling of the two problems is a promising direction to bring vision models into practice\footnote{\href{https://github.com/rahafaljundiTME/Continual-Novelty-Detection}{Our code is publicly available}. }.

\end{abstract}

%% file: sections/introduction.tex
\section{Introduction}
\label{sec:introduction}
\textit{Novelty} or Out of Distribution or Anomaly \textit{Detection} methods deal with the problem of detecting input data that are unfamiliar to a given model and out of its knowledge scope~\citep{hendrycks2016baseline} therefore avoiding misleading predictions and providing the model with a ``no'' option.
In spite of the significant potential of the novelty detection problem in real-life scenarios, most novelty detection methods are designed and tested for detecting input data drawn from datasets that are very different from those used for training (\eg, training on CIFAR-10 and detecting SVHN images~\cite{lee2018simple}) in which the discrepancy between the datasets can be exploited
rather than a difference in the represented objects.  
It has been noted that novel object detection is harder when they correspond to new categories from the same input domain~\citep{liang2017enhancing}.
Imagine for example an object detection or recognition model operating on input from driving scenes. For simplicity, let us say that the model is initially trained to recognize cars and pedestrians given available annotations. 
At deployment time, new objects could be encountered that are not covered by the initial training set \eg, bicycles and motorcycles. First, it is important that those objects are treated neither as cars nor as pedestrians since they behave and should be handled differently. Second, once these objects are spotted as novel, they can be later annotated and used to retrain the model expanding its current knowledge to include cars, pedestrians, cyclists and motorcycles. 
Even if  training data were intended to cover all possible objects, our world is constantly evolving and new objects will always appear \eg, monowheels and e-scooters did not exist a decade ago and now represent a form of transportation used in daily life.
The setting when a model is trained incrementally on a new set of classes or new tasks is referred to as lifelong learning, incremental learning or \textit{continual learning}. Continual learning provides machine learning models with the ability to update and expand their  knowledge over time with minimal computation and memory overhead. 
Research in this field is inspired by  biological intelligence, however, mammals in general are known for their distinctive ability to discriminate familiar objects from novel objects~\citep{antunes2012novel, broadbent2010object}. When faced with a novel object, they try to discover and acquire information on the novel object, thus adding it to the set of familiar known objects. Continual learning and novelty detection are essential interconnected components of intelligence yet they are currently studied as independent problems. 
Moreover, in continual learning, forgetting of some previously learned skills can be unavoidable and a model's knowledge is not only expanding but also shifting,  raising the question of how forgotten data will or should be treated by a novelty detection method.\par
In this work, we define and formulate the problem of Continual Novelty Detection, which arises when novelty detection methods are applied in a continual learning scenario where the concerned prediction model is continuously updating its existing knowledge. 

\begin{wrapfigure}{r}{0.5\textwidth}
 \vspace*{-0.5cm}
  \begin{center}
    \includegraphics[width=0.48\textwidth]{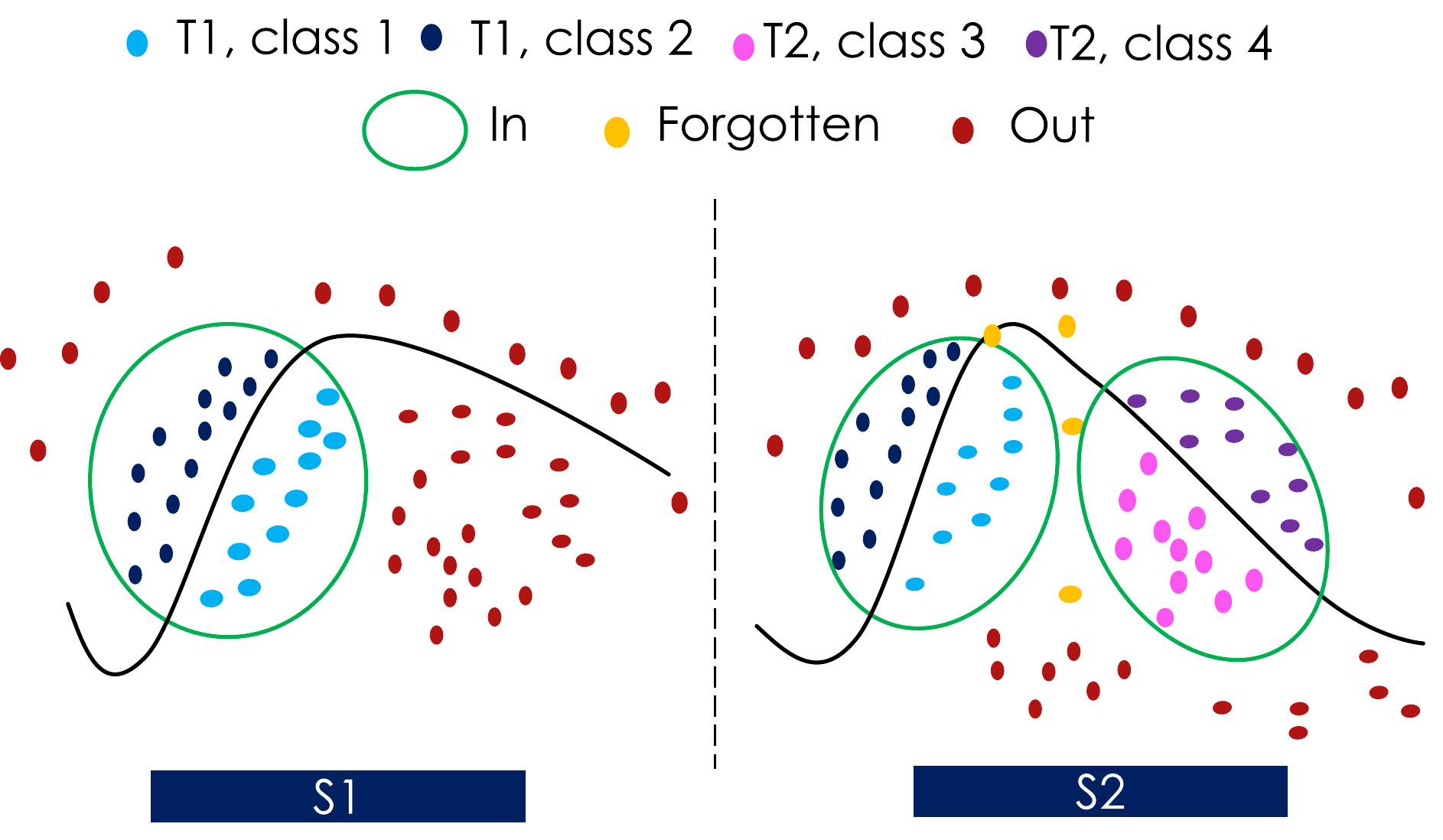}
  \end{center}
   \vspace*{-0.5cm}
  \caption{\small Illustration of three three different sets of samples constructed in the context of continual learning.}
   \label{fig:expanded seen set}
   \vspace*{-0.6cm}
\end{wrapfigure}
We divide the set of inspected samples into three subsets:~(\IN) samples covering those of previously and recently learned concepts, (\F) forgotten samples subject to mistaken predictions as a result of continual learning and, finally, the (\Out) samples, covering those of yet unseen concepts.
Our key goals are to understand (a)~the existing novelty detection methods behaviour in this setting; (b)~whether  novelty detection methods are capable of coping with a dynamic (\IN) set; (c)~how potentially forgotten samples are treated by the novelty detection methods. To this end, we propose a benchmark to evaluate several existing novelty detection methods.

We note a clear connection between continual learning and novelty detection and insights that novelty detection methods can provide about a continual learner's behaviour and vice versa. We also highlight the need for developing new novelty detection methods that are closely linked to the trained model and the deployed continual learning method.\\ 
Our contributions are as follows:
1)~we are the first to investigate the novelty detection problem under the continual learning setting~(\ref{sec:problem_description});
2)~we establish benchmarks for the described continual novelty detection problem and compare different existing methods~(\ref{sec:evaluation});
3)~To stimulate research on this problem, we highlight possible promising directions and propose  simple yet effective baselines~(\ref{sec:discussion}).

%% file: sections/related_work.tex
\section{Related Work}
\label{sec:related_work}
Our work lies at the intersection of continual learning, novelty detection and open world recognition. 
The main challenge faced by the \textbf{continual learning} methods is ``the catastrophic'' forgetting of previously learned information as a result the new learning episodes. 
Over the past years,  solutions have been proposed for different settings: the multi-head setting~\citep{aljundi2018memory, lee2017overcoming, kirkpatrick2017overcoming, finn17icml, nguyen2017variational} where each new learning episode (representing a separate  task) has a dedicated output layer, the shared-head setting~\citep{rebuffi2016icarl, zhao2020maintaining, wu2019large, hou2019learning} in which  the  learnt groups of classes at different  episodes share the same, single output layer,
and the online setting~\citep{Aljundi2019continualNoT,lopez2017gradient,aljundi2019online} where learning happens online on a stream of data. We refer to~\cite{de2019continual} for a survey. In this work, we consider multi-head and shared-head settings. Further, continual learning methods assume that new classes arrive in batches along with their labelled data while at test time only learned classes are assumed to be present.
Very recently,  methods were proposed to detect task boundaries~\citep{lee2020neural, zeno2018task, aljundi2018task}, \ie, when a new task different from the existing training data is encountered. However, this  detection assumes that samples of new tasks are received in batches along with their labels and mostly rely on the statistics of the loss function for such detection.\\
\textbf{Novelty Detection}
 or Out of distribution detection~(\OOD) methods \citep{simonyan2014very, hendrycks2016baseline, liang2017enhancing, hsu2020generalized, bodesheim2015local, schultheiss2017finding} tackle the problem of detecting test data that are different from  what a model has been trained on.
The detection is done offline where it is assumed that In distribution is final. Moreover, evaluation benchmarks don't usually consider novel objects of similar input domain.
Note that \OOD approaches are also concerned with detecting adversarial attacks, however, in this work we are mainly  interested in detecting relevant novel objects and not random input or adversarial attacks.\\
\myparagraph{Open World Recognition}
 describes the general problem of learning, identifying unknowns and the novel data that should be labelled and added to the existing training set, where new training steps can be performed~\citep{bendale2015towards}. Open world problems typically assume that the test examples which are identified as novel, are then labelled by an expert and used in new learning episodes. However in many scenarios labels are not available for these examples. In this work, data corresponding to the new learning steps are independent of the novelty detection results. 
\cite{boult2019learning} have critically reviewed open world methods demonstrating  unsuitably for deep neural networks.
To the best of our knowledge, more recent works on open world for classification are very scarce. There is only one extension of~\citep{bendale2015towards}, to deep networks~\citep{mancini2019knowledge} and an improved version~\citep{fontanel2020boosting} for web aided robotics applications, where training data are extracted from the web. This method simply relies on the nearest class mean distance for detecting novel input. Existing novelty detection methods for classification problems have not been evaluated out of the offline batch setting which is the aim of this work. With regards to the object detection problem, the open world detection methods~\citep{joseph2021towards,yang2021objects} are specific to object detectors where background class and multiple instances per image are essential in their formulation, hence not applicable to the general novelty detection problem for classification. These few works lack a proper validation of the novelty detection aspect and focus on the performance and accuracy of the learned classes under the open world setting.
Moreover, in the few existing open world works, there is no specific use or discussion of possible continual learning methods, usually  a simple stack of previous samples or their features is used to prevent catastrophic forgetting. As such, an evaluation of  existing continual learning and novelty detection approaches in such setting is still missing. In summary, the existing few works of open world lack a proper consideration of the main continual leaning challenge, catastrophic forgetting, and how it affects the novelty detection performance.
 Whilst the  open world setting is certainly interesting, we believe  it is a complicated problem to start with,  given the current state of continual learning and novelty detection. Differently, in this work instead of attempting at solving the whole open world problem at once, we start with the first essential step of studying how novelty detection  can be extended to a continual learning scenario.\\
Given the recent efforts and advances in continual learning:
We study how deploying a continual learning method to facilitate learning while preventing catastrophic forgetting affects the performance of existing novelty detection methods. Under this setting, we define 3 sets that emerge from the continual learning process (\IN, \Out and \F.)
We study novelty detection methods behaviour under various continual learning settings with and without example replay, different architectures and different benchmarks.

%% file: sections/problem_description.tex
\section{Problem Description}
\label{sec:problem_description}
We consider a continual learning scenario where batches of labelled data arrive sequentially at different points in time. Each new batch corresponds to a new group of categories that could form a separate task $\mathcal{T}_t$. Given a backbone model $\mathcal{M}$ \eg, a deep neural network, data arriving at time $t$ will be learnt in a stage $S_t$ updating the model status from $\mathcal{M}_{t-1} $ to $\mathcal{M}_t$, learning at stage $S_t$ will generally affect performance on all previous tasks $\mathcal{T}_1 \cdots \mathcal{T}_{t-1}$. Acquired knowledge will be accumulated and updated across time and access to previously seen data is prohibited or limited. 
To prevent  catastrophic forgetting of previously acquired knowledge, a continual learning method is applied during learning while maintaining a low memory and computational footprint. In between learning stages, the model will be deployed in a given test environment and should perform predictions on input covered by its acquired knowledge while identifying irrelevant input. 
A continual novelty detection method $\ND(\mathcal{M}_t)$ should be able to tell after each learning stage $S_t$ how familiar and similar a given input is to the current model $\mathcal{M}_t$ knowledge. The detection is carried online given each input.  Following existing novelty detection methods, the output of a ND method is a score $sc$ that is higher for In distribution samples  than for Out of distribution samples. 
Later on, novel inputs can be gathered and annotated, or other data can become available for training the model “continually” in a new learning stage $S_{t+1}$.

$\ND(\mathcal{M}_t)$  should reflect the incurred  change as $\mathcal{M}_{t-1}$ transition to $\mathcal{M}_t$ expanding the set covered by the trained model to contain samples of the previous and the current learning stages. Further, as forgetting of previous knowledge can occur,  we want firstly to understand how forgotten data are treated by the ND method and also   examine what the novelty detection behaviour could tell about the continual learning.  
For that purpose, after each stage $S_T$, we divide the test data  into three sets:
\begin{itemize}[noitemsep,topsep=0pt,parsep=0pt,partopsep=0pt]
    \item {\IN}: Samples that are represented by the current training stage or those from previous stages whose predictions remain correct (\ie, not forgotten). As in~\cite{hendrycks2016baseline} the  \IN set contains only correctly predicted test samples.
    \begin{equation}
        In_{t}=\{x: x,y\sim D^k_{x,y}	\wedge  k\in [1..t] 	\wedge \mathcal{M}_t(x)=y\}
    \end{equation}
    \item {\Out}: Samples of unknown tasks, classes or domains that are not covered by the current model, this set contains samples that could be covered in the future.
        \begin{equation}
        Out_{t}=\{x: x,y\sim D^k_{x,y}	\wedge  k\notin [1..t] \}
    \end{equation}
    \item{\F}: Samples of previously learned stages that  were predicted correctly and are predicted wrongly now. 
        \begin{multline}
        Forg_{t}=\{x: x,y\sim D^k_{x,y}	\wedge  k\in [1..t-1] 	\wedge \mathcal{M}_t(x)\neq y
        \wedge  \mathcal{M}_k(x)= y\}
    \end{multline}
    Forgetting is an essential phenomena when training models incrementally, within this work, we aim at understanding how this  phenomena affects novelty detection methods. Hence, we construct the \F set, to examine how forgotten samples are treated by a ND method, as \Out or \IN, or if forgotten samples can be potentially identified  as a separate set e.g., with a score lower than \IN and higher than \Out,  $sc$(\IN)$>sc$(\F)$>sc$(\Out). If identified, those samples can be further utilized to update the model, recovering previous important information.
\end{itemize}
Note that all sets are constructed from \textit{test} data not seen during the different training stages.  Figure~\ref{fig:expanded seen set} illustrates how these sets are constructed during continual learning. 

%% file: sections/evaluation.tex
\section{Empirical Evaluation of Existing methods}
\label{sec:evaluation}
\subsection{Datasets}
In order to evaluate existing ND methods in the setting of CL, we construct the following learning sequences that are commonly used to evaluate state of the art CL methods:\\
1)~TinyImageNet Sequence: a subset of 200~classes from ImageNet~\citep{tinyimgnet}, rescaled to size $64\times64$ with 500~samples per class. Each  stage represents a learning of a random set of 20~classes composing 10~stages~\citep{de2019continual}. Here, all training stages data belong to one dataset in which no data collection artifacts or domain shift can be leveraged in the detection;\\
2) Eight-Task Sequence: Flower~\citep{Nilsback08} $\rightarrow$ Scenes~\citep{quattoni2009recognizing} $\rightarrow$ Caltech USD Birds~\citep{WelinderEtal2010} $\rightarrow$ Cars~\citep{krause20133d} $\rightarrow$ Aircraft~\citep{maji13fine-grained}
$\rightarrow$ Actions~\citep{pascal-voc-2012} $\rightarrow$ Letters~\citep{deCampos09} $\rightarrow$ SVHN~\citep{netzer2011reading}. The samples of each task belong to a different dataset and each task represents a super category of objects. This CL sequence has been studied in~\citep{aljundi2018memory, de2019continual} under the multi-head setting. 
\begin{table*}[ht]
 \vspace*{-0.2cm}
\centering
\tabcolsep=0.11cm
\small
\resizebox{\textwidth}{!}{
\begin{tabular}{|c|c |c|c|c|c|c|c|c|c|c|c|c|c|  }
 \hline
&\multicolumn{10}{|c|}{TinyImageNet}& \multicolumn{3}{|c|}{\tasks} \\
 \hline
Setting &\multicolumn{4}{|c|}{ Shared-head }&\multicolumn{6}{|c|} {Multi-head} & \multicolumn{3}{|c|} {Multi-head}\\
\hline
Architecture&\multicolumn{4}{|c|}{ ResNet}&\multicolumn{3}{|c|} {ResNet}&\multicolumn{3}{|c|} {VGG} &\multicolumn{3}{|c|} {AlexNet}\\\hline
CL Method &ER 25s&ER 50s&SSIL 25s&SSIL 50s& LwF & MAS & FineTune & LwF & MAS & FineTune & LwF & MAS & FineTune\\ \hline
Acc. $\uparrow$&$26.20\%$&$32.09\%$&$31.26\%$&$33.02\%$& $69.39\%$&$61.59\%$&$37.97\%$ &$61.7\%$& $53.94\%$&$32.66\%$&$50.95\%$&$50.26\%$&$29.51\%$ \\\hline
Forg. $\downarrow$&$24.72\%$&$14.2\%$&$15.6\%$&$15.02\%$& $3.5\%$& $2.01\%$&  $36.47\%$&$4.01\%$&  $9.37\%$ &$37.54\%$&$5.0\%$& $2.90\%$& $26.38\%$\\
 \hline
\end{tabular}}
\vspace{-0.3cm}
\caption{\label{tab:cL-performance} \small  CL Average Accuracy and Average Forgetting at the end of the training sequence for each considered setting with  different  continual learning methods. Finetune forgets catastrophically, MAS is more conservative but less flexible compared to LwF. SSIL improves over ER in the low buffer regime.}
\vspace{-0.3cm}
 \end{table*}
 
\subsection{Continual Learning Setting}
In this work, we consider the setting where training data of each  stage $S_t$  is learnt offline with multiple epochs. 
In the context of image classification, we study both multi-head and shared-head settings explained in the sequel.\\
\myparagraph{Multi-head setting.} Data of each learning stage $S_t$ is learnt with a separate output layer (head). It is largely studied and it is less prone to interference between tasks as the last output layer is not shared among the learned tasks. However, knowledge of the task identity is required at test time to forward the test sample to the corresponding output layer.
In order to limit forgetting, we train the prediction model with a CL  method. We use two well established methods from two different families:
 Learning without Forgetting~(LwF)~\citep{li2016learning}
  uses the predictions of the previous model~$\mathcal{M}_{t-1}$ given the new stage $S_t$ data to transfer previous knowledge to  the new model $\mathcal{M}_t$ via a knowledge distillation loss.
 Memory Aware Synapses~(MAS)~\citep{aljundi2018memory} estimates an importance weight of each model parameter \wrt the previous learning stages. An L2 regularization term is applied to penalize changes on parameters deemed important for previous stages. 
These methods are lightweight, require no retention of data from previous tasks and are well studied in various CL scenarios. We also evaluate a fine-tuning baseline that does not involve any specific technique to reduce forgetting
, we refer to this as \fine.  Note that multi-head setting requires at test time  a ``task oracle'' to forward the test data to the corresponding head.\\
\myparagraph{Shared-head setting.}
Here, the last output layer is shared among categories learned at the different stages. This is a more realistic setting, however prune to severe forgetting of previous tasks. As such, most approaches use a  buffer of a subset of previously seen samples. We study shared-head continual learning with a fixed buffer of different sizes corresponding to  25~samples per class and 50~samples per class as in~\cite{de2019continual}.
We consider two continual learning methods:
 Experience Replay (ER)~\citep{David2018Experience} where buffer samples of previous classes are simply replayed while learning new classes and Cross Entropy loss is minimized on replayed and new samples together at each iteration.
Separated Softmax for Incremental Learning (SSIL)~\citep{ahn2021ss}  where a  Softmax and task wise knowledge distillation are applied to each task logits separately in order to reduce the interference with previous tasks logits activations.\\
\myparagraph{Architecture.}
We consider three different backbones, ResNet and VGG  for \tinyimg. For the \tasks, we use a pretrained AlexNet model as in~\cite{de2019continual, aljundi2018memory}. \\
\myparagraph{Training details.}
For  CL, early stopping on a validation set was used at each learning stage.  
Following~\cite{mirzadeh2020understanding} an appropriate  choice of regularization and hyperparameters could lead to a stable SGD training resulting in  smaller degrees of forgetting. Hence, we use SGD optimizer where dropout,
learning rate schedule, learning rate, and batch size were treated as hyper-parameters. Since our goal is to study {the behaviour of ND methods} in a CL scenario, we chose CL hyper-parameters that achieve the best possible CL performance, in order to reduce any possible effect of negative training artifacts on the ND performance. Thus, our CL results might be higher than those reported in~\cite{de2019continual}. 
Training was performed over ten random seeds and mean accuracy and forgetting are reported.
\subsection{Novelty Detection Setting}
\myparagraph{Deployed ND methods.}
For novelty detection, a method should be able to perform prediction given an input data point and a model. We do not consider methods that require simultaneous access to the training data due to the continual learning scenario under consideration. We select four different ND methods covering different detection strategies. 
1) \nav the output probability of the predicted class (an image classification scenario) after a softmax layer is used as a score for samples of the \IN set~\citep{hendrycks2016baseline}.
2) \odin~\citep{liang2017enhancing} adds a small perturbation to the input image and uses the probability of the predicted class after applying temperature scaling as a score.
3) \mah~\citep{lee2018simple} estimates the mean of each class  and a covariance matrix of all classes given  training data features. The Mahalanobis distances of different layers features are used by a linear classifier to discriminate \IN~\vs~\Out samples. The covariance matrix and the mean are updated when new classes are added. 
4) density models have been proposed as a way to perform OOD and we pick deep generative model trained as a \vae ~\citep{an2015variational} as an example.  
Note that in the context of continual learning, the \vae itself is prone to catastrophic forgetting~\citep{nguyen2017variational}. We treat the choice of \vae input (raw input or features of the model's last layer before classification) as a hyper parameter.
   \begin{figure*}[t]
    \vspace*{-0.5cm}
    \centering
    \subfloat[{\footnotesize C.AUC~($\%$),  \fine}]{{\includegraphics[width=.33\textwidth]{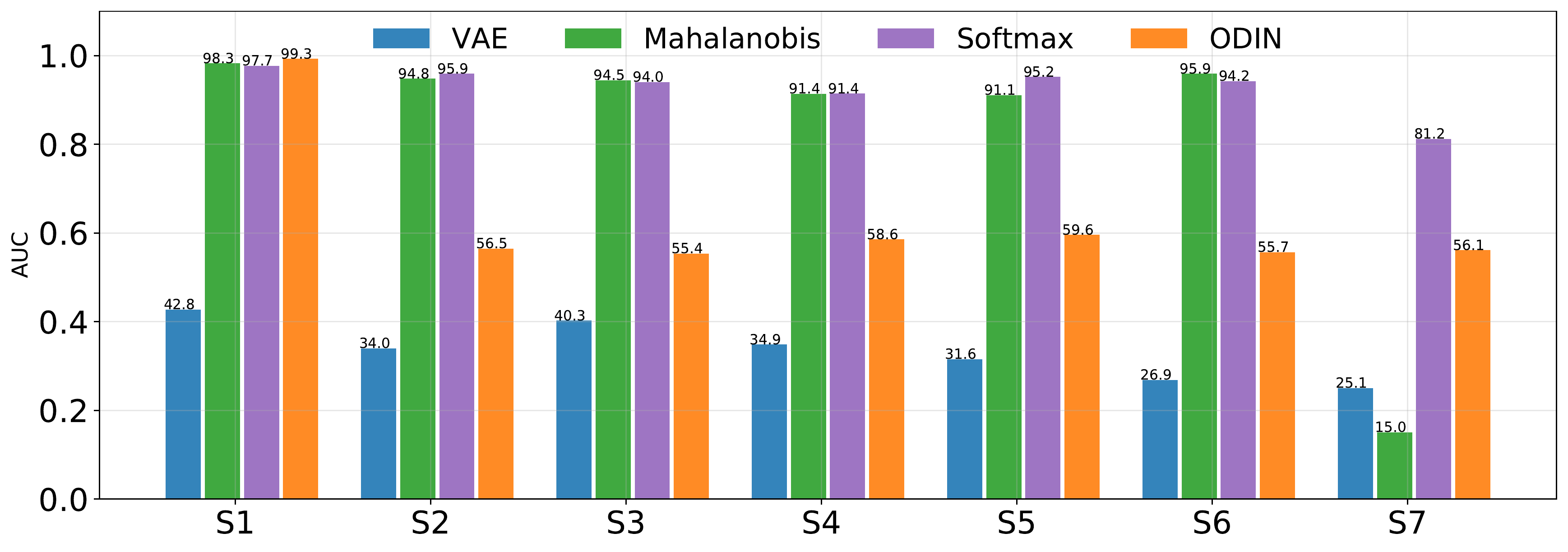}} }%
     \hfill
    \subfloat[{\footnotesize R.AUC~($\%$),  \fine}]{{\includegraphics[width=.33\textwidth]{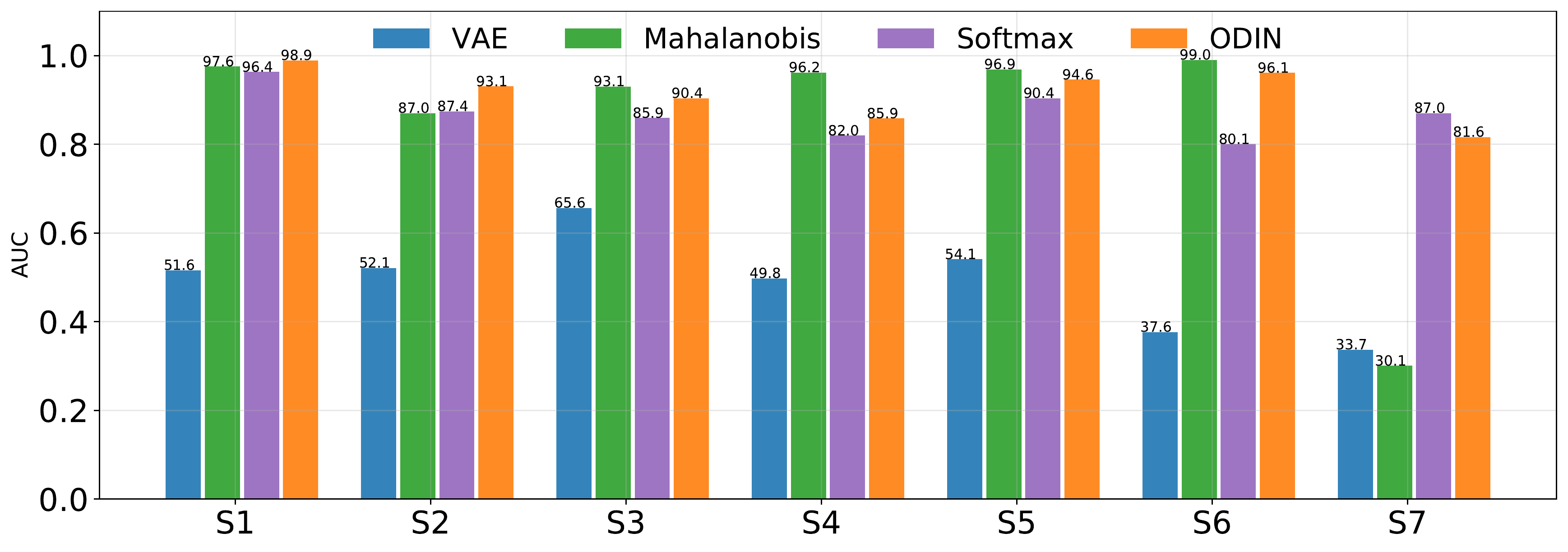} }}
        \hfill
    \subfloat[{\footnotesize P.AUC~($\%$),  \fine}]{{\includegraphics[width=.32\textwidth]{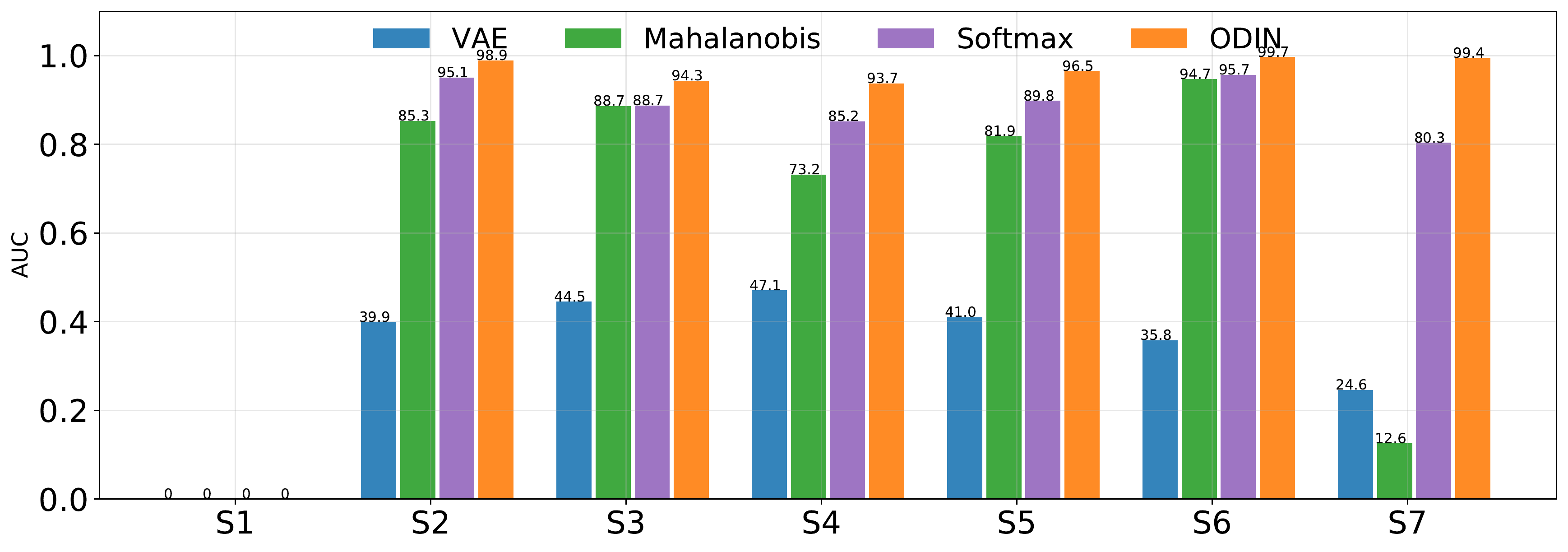}}}
 \vfill
    \centering
    \subfloat[{\footnotesize C.AUC~($\%$),  \MAS}]{{\includegraphics[width=.33\textwidth]{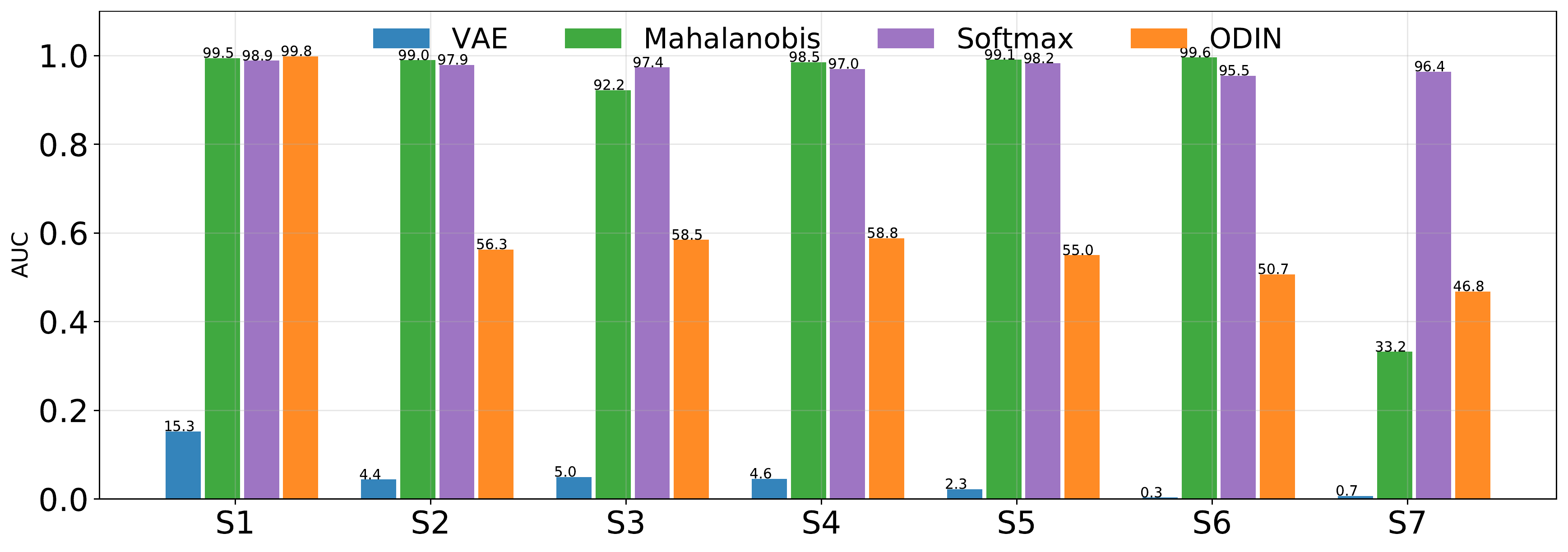}} }%
     \hfill
    \subfloat[{\footnotesize R.AUC~($\%$),  \MAS}]{{\includegraphics[width=.33\textwidth]{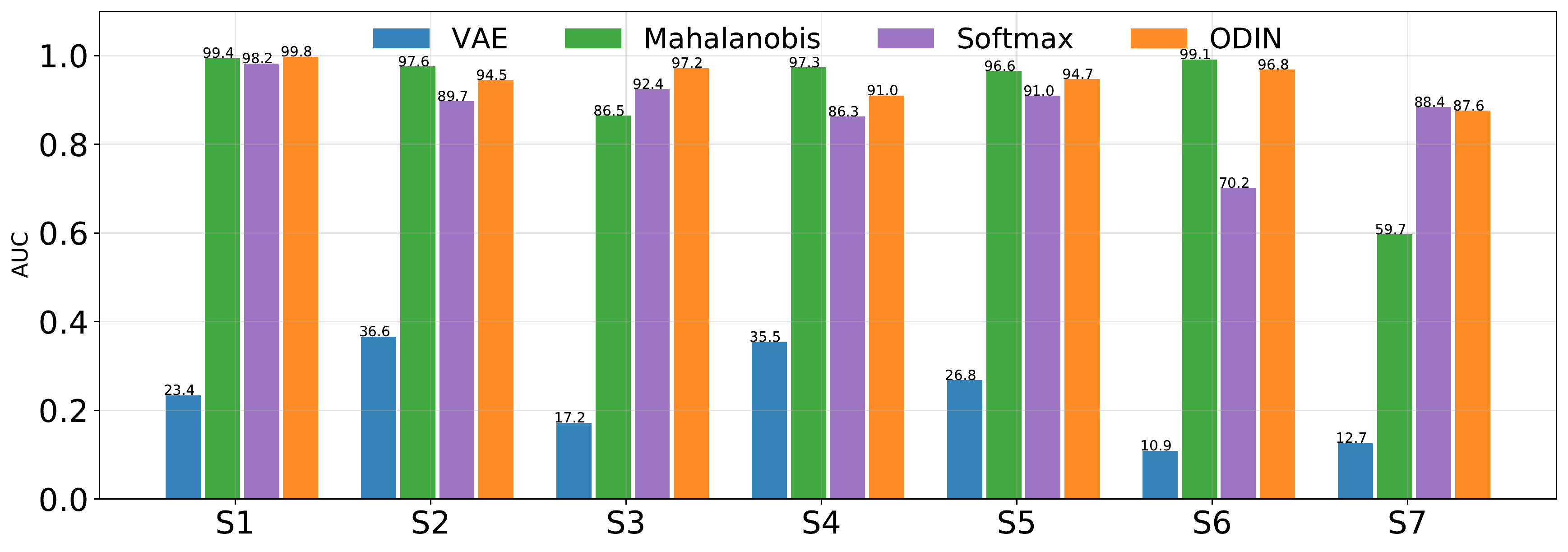} }}
        \hfill
    \subfloat[{\footnotesize P.AUC~($\%$),  \MAS}]{{\includegraphics[width=.32\textwidth]{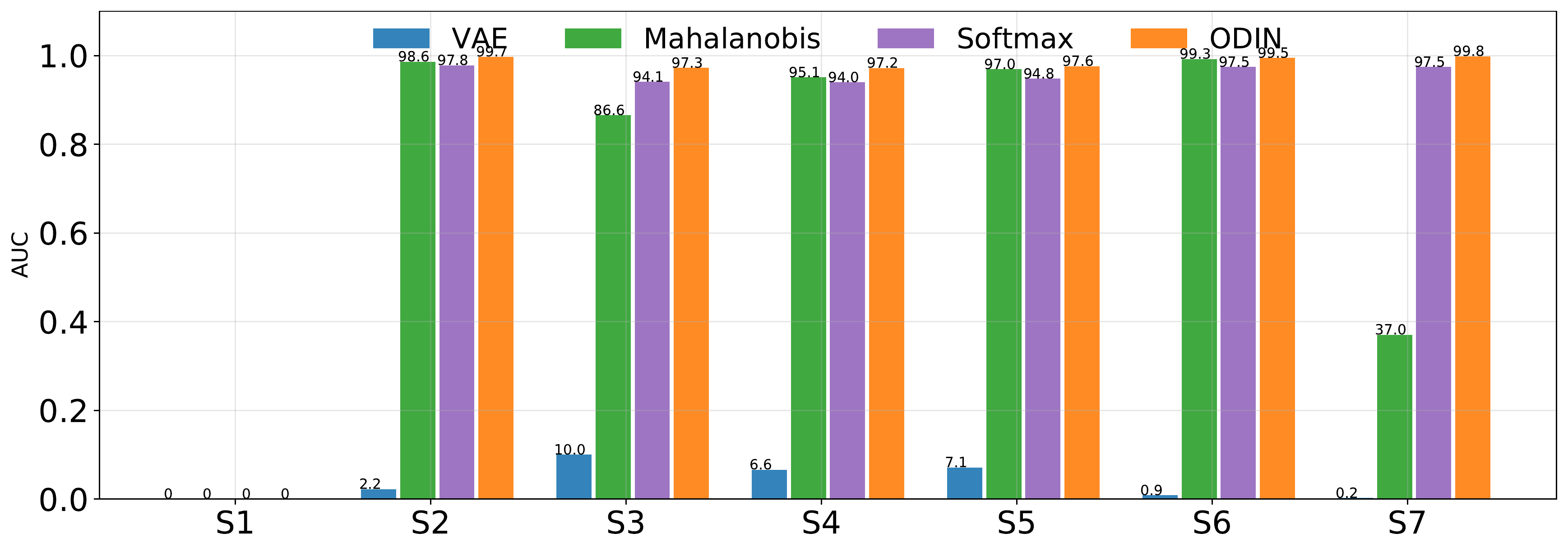}}}
\vfill
    \centering
    \subfloat[{\footnotesize C.AUC~($\%$),  \LwF}]{{\includegraphics[width=.33\textwidth]{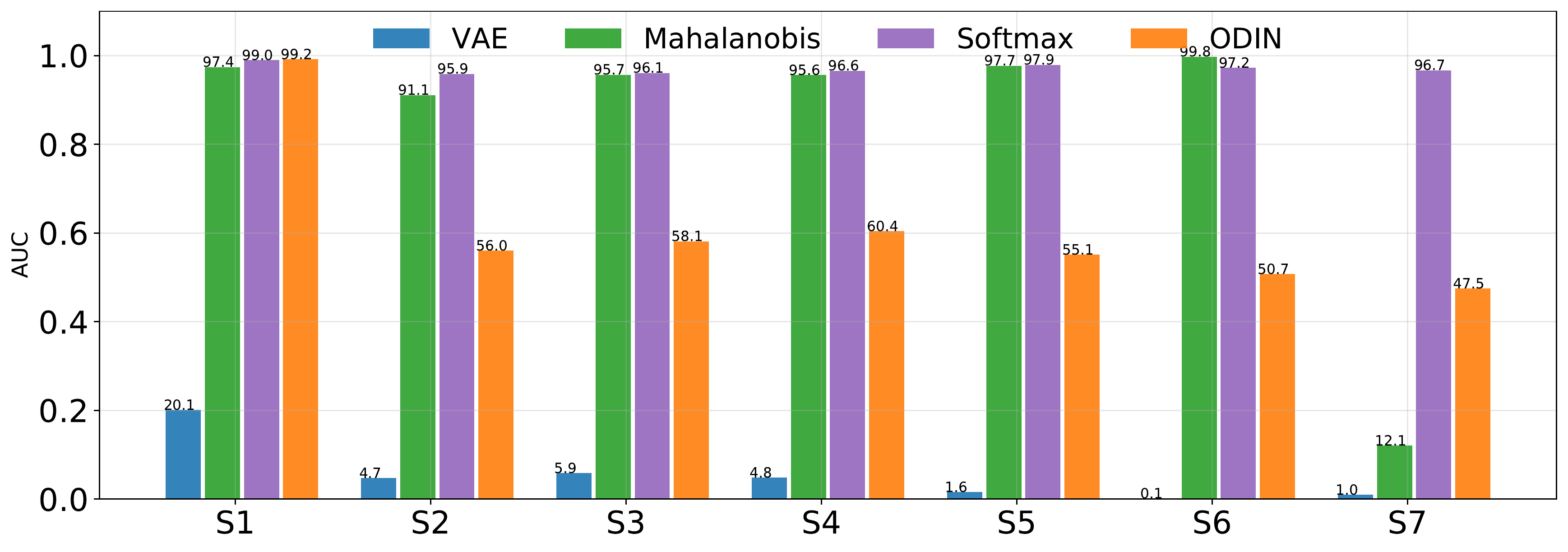}} }%
     \hfill
    \subfloat[{\footnotesize R.AUC~($\%$),  \LwF}]{{\includegraphics[width=.33\textwidth]{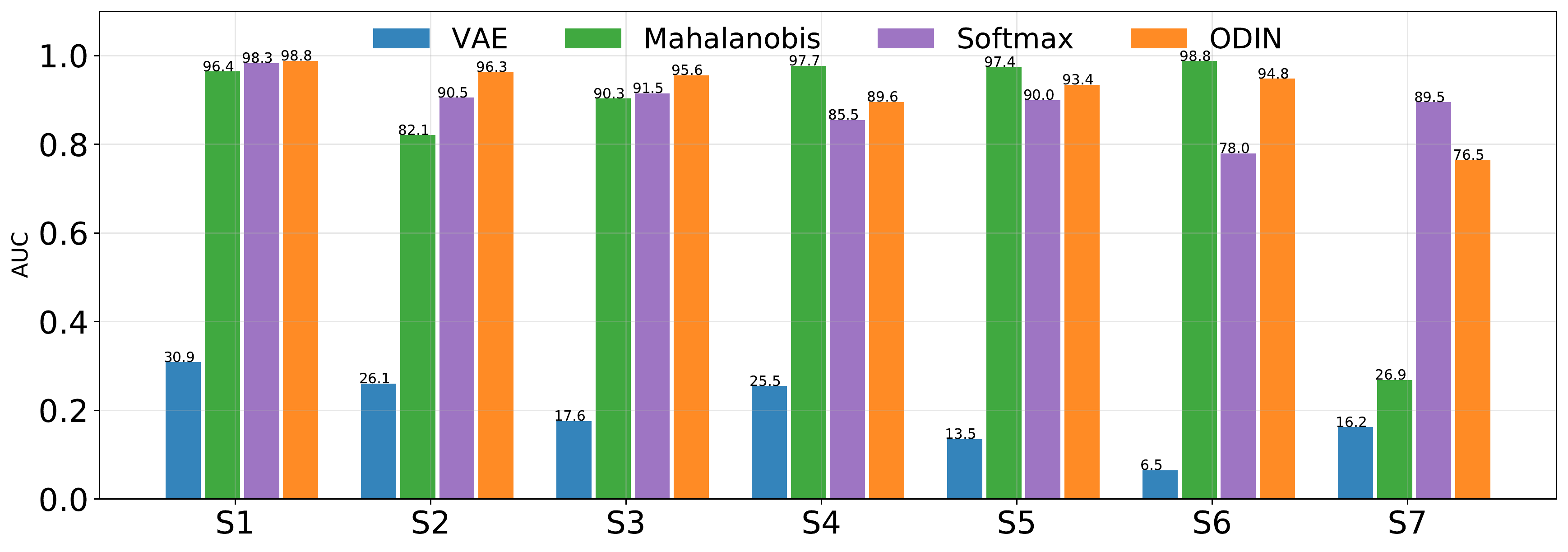} }}
        \hfill
    \subfloat[{\footnotesize P.AUC($\%$),  \LwF}]{{\includegraphics[width=.32\textwidth]{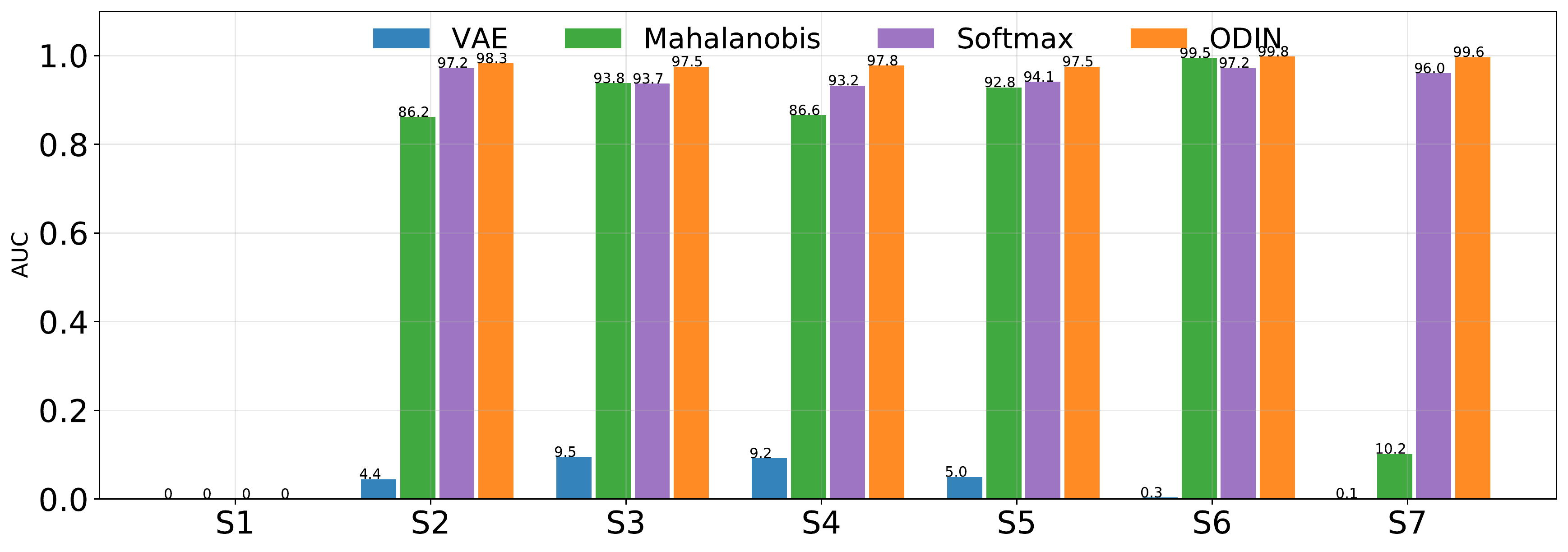}}}
     \vspace*{-0.3cm}
\caption{\small Novelty detection results for~\tasks in terms of Combined ~AUC~(C.AUC), Recent stage AUC~(R.AUC) and  Previous stages AUC~(P.AUC) ~with \fine~ \label{fig:8_tasks_fine} first row, \MAS second row~\label{fig:8_tasks_MAS} and  \LwF~\label{fig:8_tasks_LwF} third row.}%
     \vspace*{-0.8cm}
\end{figure*}
\\
\myparagraph{Evaluation.}
After each stage $S_t$ in the learning sequence, we evaluate the ND performance on the following detection problems constructed from the three sets mentioned earlier:\\
{\IN~/~\Out}: 
It serves to examine the detection ability of samples from unseen categories~(\Out) vs.~samples of  learned categories.
Note that the \IN set is changing after each learning stage to contain correctly predicted test samples from all the learning stages. Here {\Out} contains test samples classes that are yet to be encountered in the continual learning sequence. Hence the novelty detection evaluation is carried on until stage $T$-$1$ where $T$ is the length of the sequence.\\
{\IN~/~\F} \& {\F~/~\Out}  :  We want to examine if the forgotten samples are treated as novel, similarly to those in the~\Out set. In other words, can an ND method discriminate between a forgotten sample and a correctly predicted one? Moreover, does the predicted score for \Out differ from that of \F. Ideally, if an ND method could tell if a sample is forgotten  rather than completely novel,  those identified samples can later be treated differently either in the way they are re-annotated or re-learnt by the model. 

\myparagraph{Metrics.}
We consider the following metrics:
AUC: Area Under the receiver operating Characteristic curve~\citep{hendrycks2016baseline}. It  can be interpreted as the probability that an \IN sample has a greater score than an \Out sample.\\
AUPR-IN: Area Under the Precision Recall curve where \IN samples are treated as positives~\citep{hendrycks2016baseline}.
Detection Error (DER): The minimum possible detection error~\citep{liang2017enhancing}.\\
To examine closely the CL effects and how \IN samples belonging to previous learning stages are treated compared to those  of the most recent stage of learning, we propose  the following metrics:~1)~Combined AUC (C.AUC), measures the novelty detection performance for all \IN samples regardless of the activated head, this is to check if the predicted scores differ across tasks, 2) the most Recent  stage AUC (R.AUC), only concerned with  \IN samples of the latest stage, 3)~the Previous stages AUC (P.AUC) where \IN represents only data  relevant to previous stages.
 For each ND method we tune the hyper-parameters based on an out of distribution dataset, namely CIFAR-10.  
\subsection{Results}

\subsubsection{Continual Learning  Results}

\begin{figure*}[t]
  \vspace*{-0.6cm}
    \centering
    \subfloat[{\footnotesize C.AUC~($\%$),~ \fine}]{{\includegraphics[width=.33\textwidth]{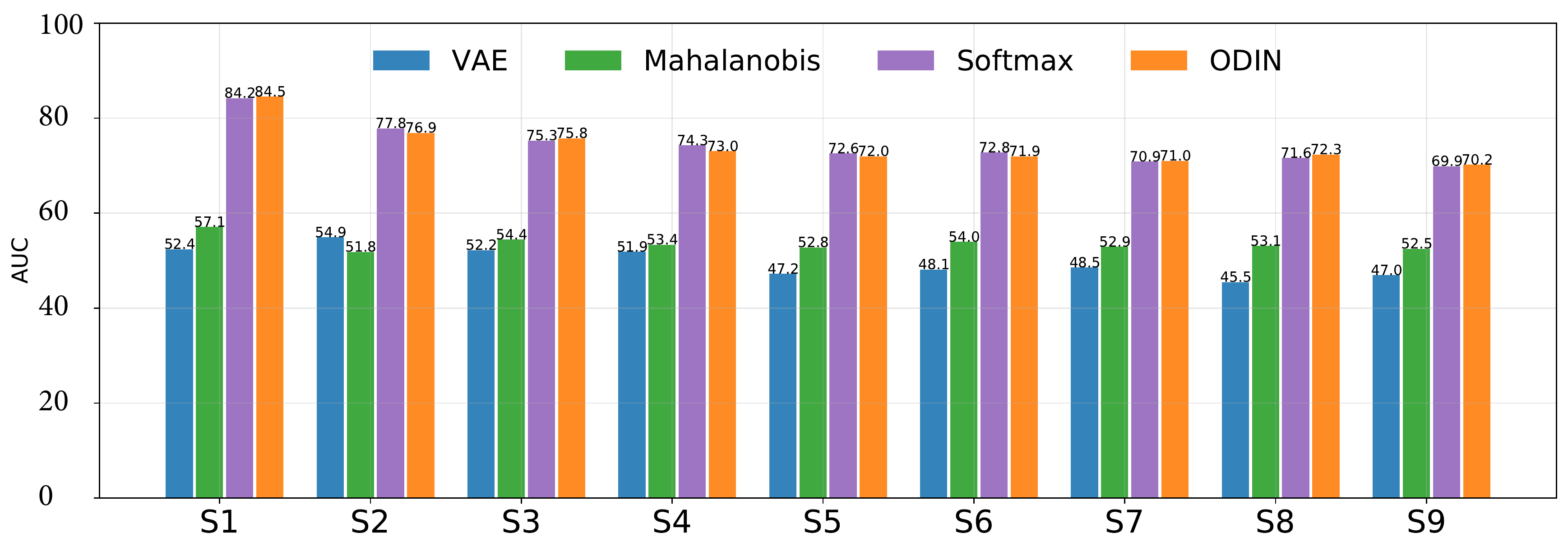}} }%
     \hfill
    \subfloat[{\footnotesize R.AUC~($\%$),~ \fine}]{{\includegraphics[width=.33\textwidth]{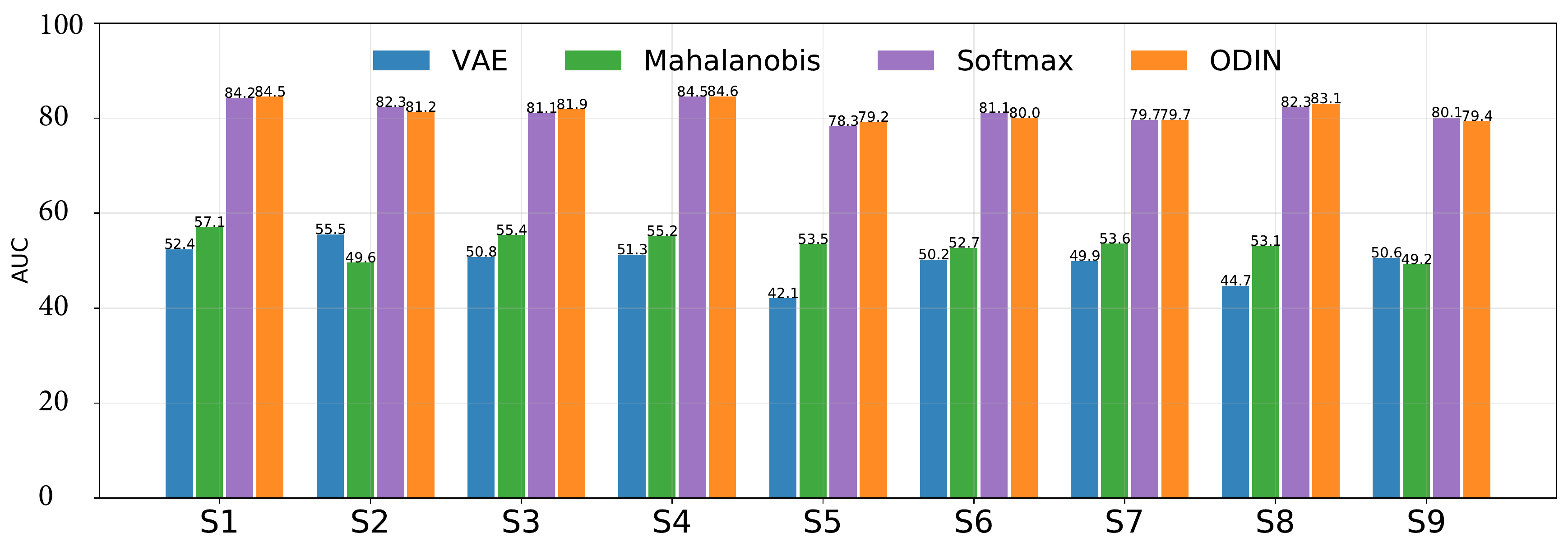}}}
        \hfill
    \subfloat[{\footnotesize P.AUC~($\%$),~ \fine}]{{\includegraphics[width=.32\textwidth]{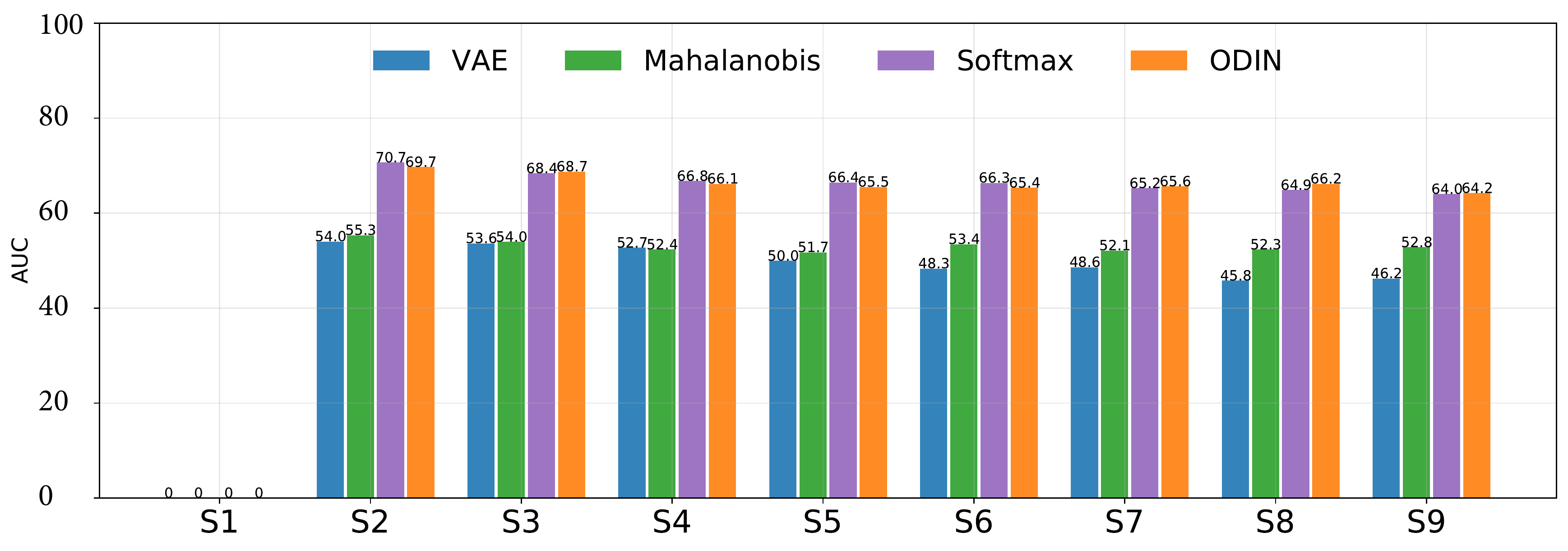}}}
\vfill
    \centering
    \subfloat[{\footnotesize C.AUC~($\%$),~ \MAS}]{{\includegraphics[width=.33\textwidth]{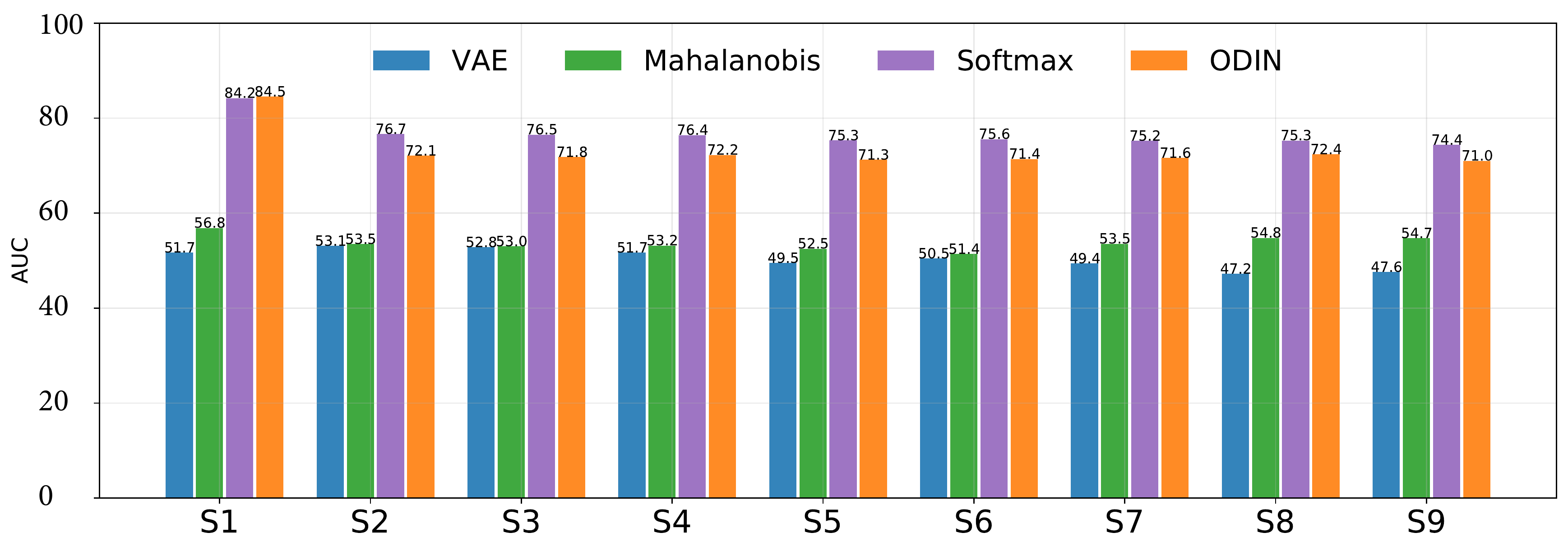}} }%
     \hfill
    \subfloat[{\footnotesize R.AUC~($\%$),~ \MAS}]{{\includegraphics[width=.33\textwidth]{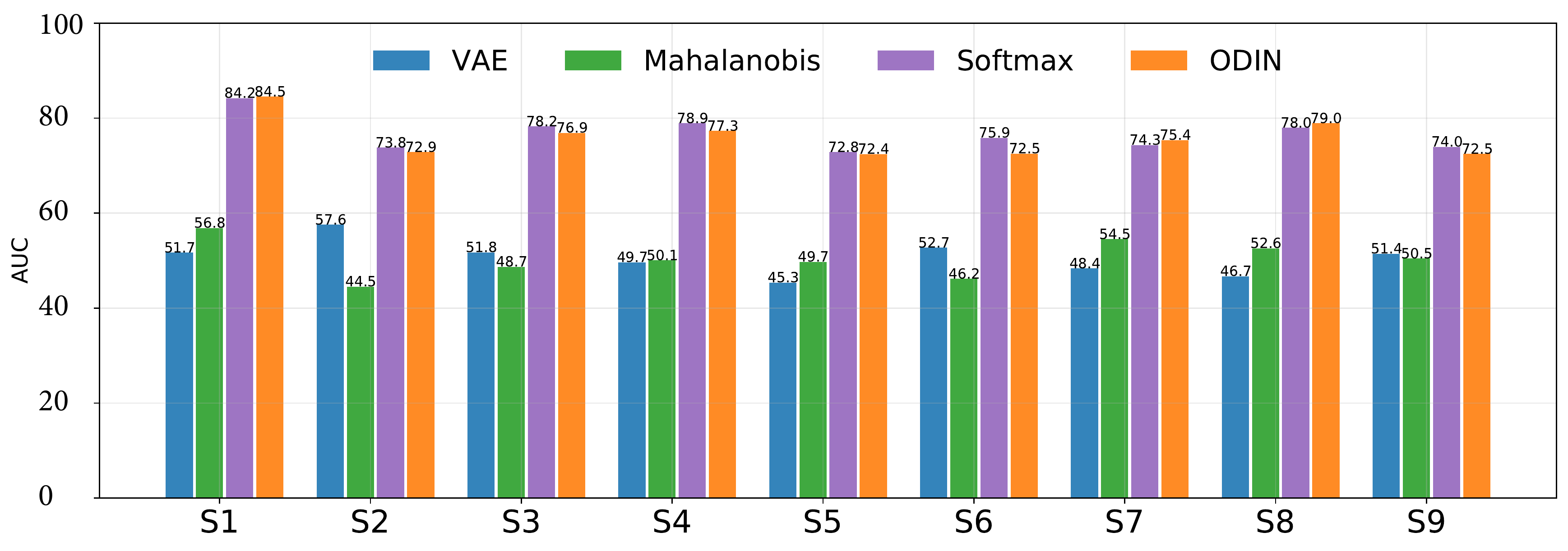}}}
        \hfill
    \subfloat[{\footnotesize P.AUC~($\%$),~ \MAS}]{{\includegraphics[width=.32\textwidth]{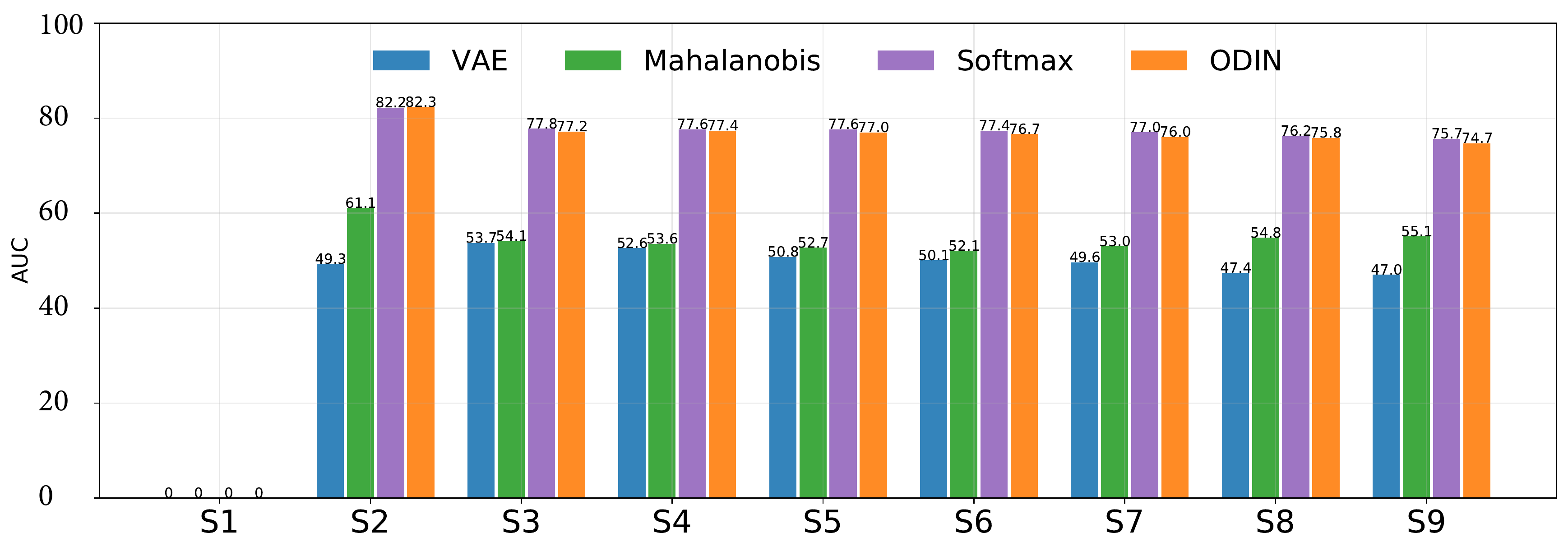}}}
\vfill
    \centering
    \subfloat[{\footnotesize C.AUC~($\%$),~ \LwF}]{{\includegraphics[width=.33\textwidth]{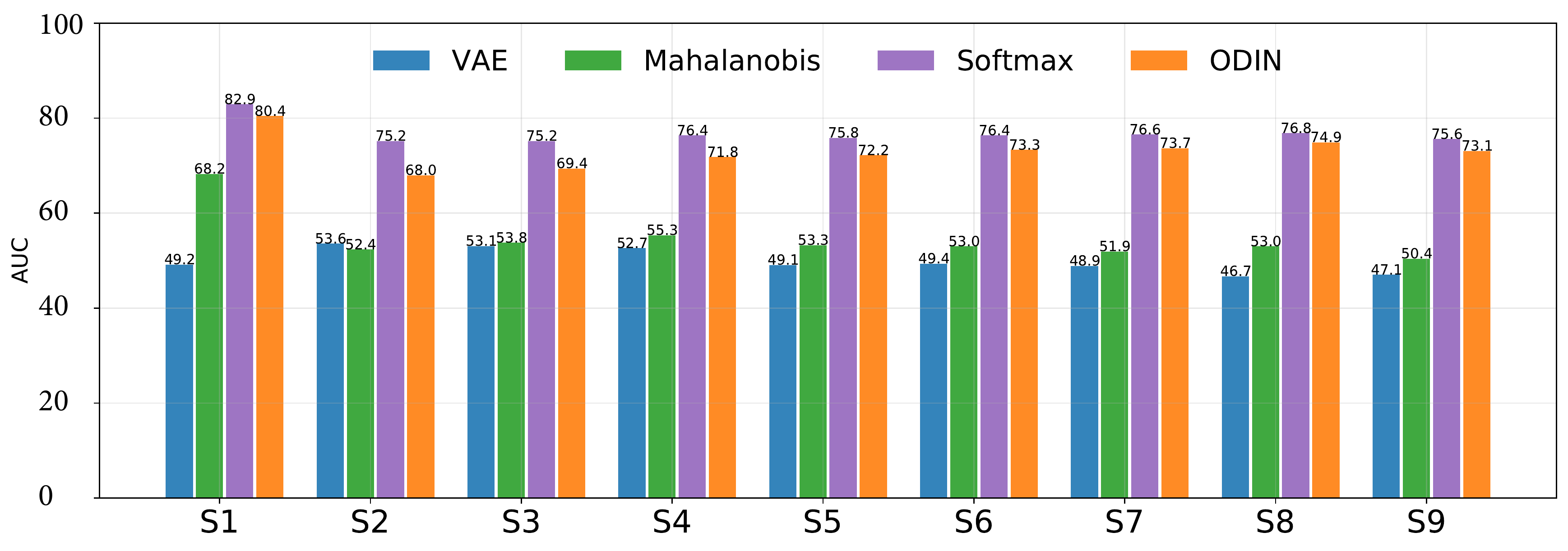}} }%
     \hfill
    \subfloat[{\footnotesize R.AUC~($\%$),~ \LwF}]{{\includegraphics[width=.33\textwidth]{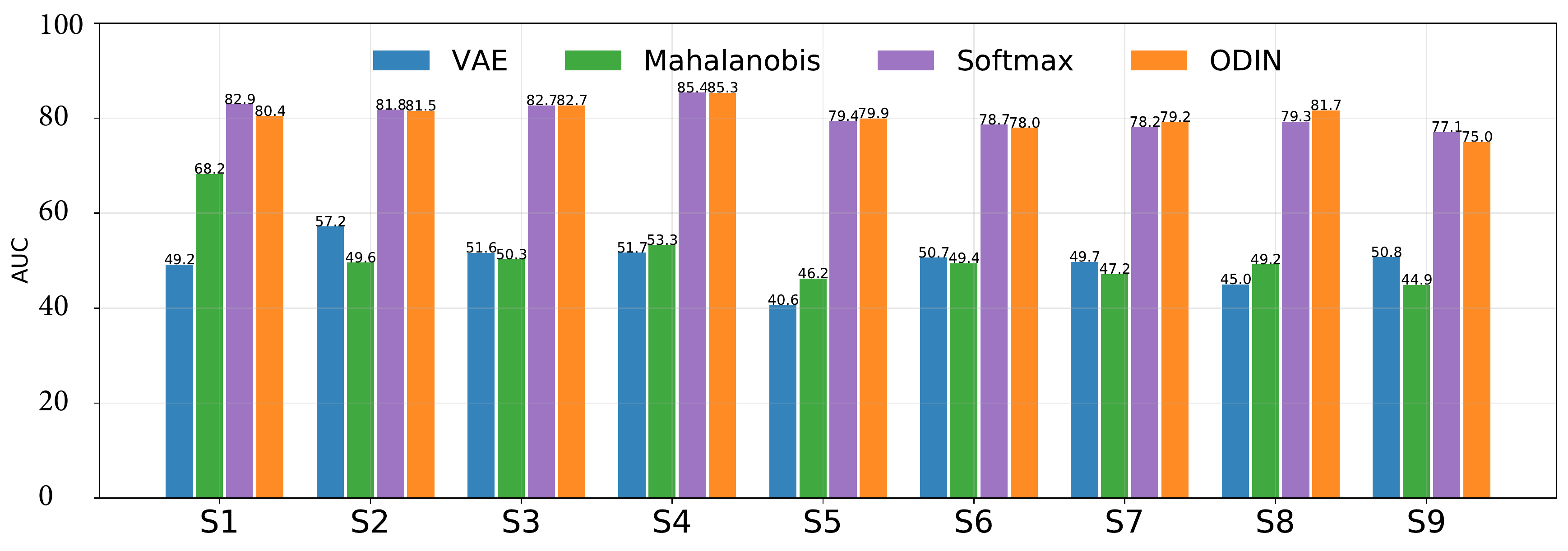}}}
        \hfill
    \subfloat[{\footnotesize P.AUC~($\%$),~ \LwF}]{{\includegraphics[width=.32\textwidth]{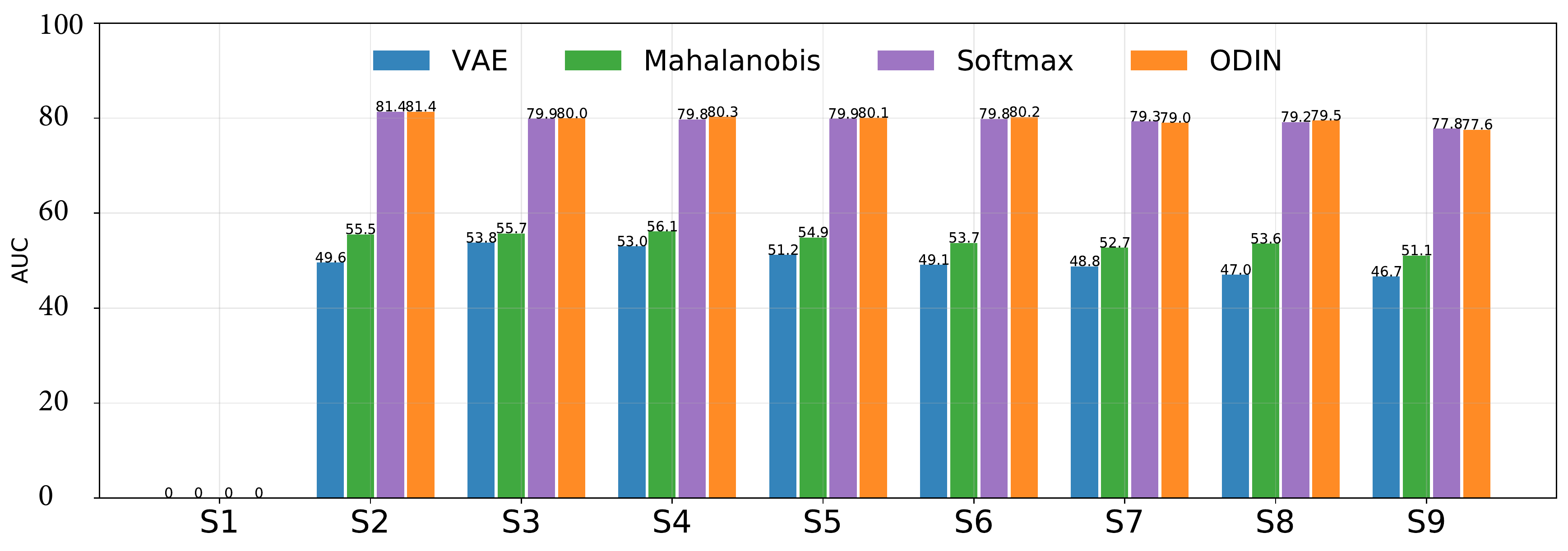}}}
    \vspace*{-0.3cm}
\caption{\small Novelty Detection results in terms of Combined ~AUC~(C.AUC), Recent stage AUC~(R.AUC) and  Previous stages AUC~(P.AUC) for  \tinyimg~with \fine~\label{fig:10_tasks_Fintune} first row, \MAS~\label{fig:10_tasks_MAS} second row, and \LwF~\label{fig:10_tasks_LwF} third row.}%
    \vspace*{-0.6cm}
\end{figure*}
We first report CL methods performance in order to note the methods' behaviour in the studied settings  and lay the ground for  a better understanding of the ND methods performance in the different cases.
Table~\ref{tab:cL-performance} reports the CL performance in terms of average accuracy and average forgetting at the end of each learning sequence under different settings. In \supp, we report each task performance.\\
\MAS applies a regularization term directly to the network parameters, penalizing their changes. As such, in general, MAS is more conservative than \LwF and  tends to largely preserve the performance of the very first learned classes. However the later stages struggle in the learning process, failing to reach the expected performance of each task. 
For \LwF no direct constraints are applied to the  parameters but knowledge distillation is deployed on the network output giving the later learning stages  flexibility in steering the parameters towards a better minimum. As such, \LwF shows lower accuracy on classes encountered in the early stages  and larger forgetting but higher accuracies on the later. 
\LwF performs best especially in \tinyimg. 
\fine does not control forgetting,   resulting in a low performance on previous stages and best performance on the most recent one.  For shared head settings, in general larger forgetting rates are shown as a result of the shared output layer. The larger the deployed buffer, the higher the previous tasks performance and the smaller the forgetting.  \SSIL~ succeeds in reducing tasks interference and hence forgetting compared to~\ER~ especially when a small buffer is deployed.
\\
\subsubsection{Novelty Detection Results}
Having discussed the continual learning performance across different sequences and different methods, here we study the obtained novelty detection performance and how the shown differences in the made CL tradeoffs affect the ND methods performance.
\\
\myparagraph{Eight-Task sequence ND results.}
Figure~\ref{fig:8_tasks_MAS}  shows the ND performance based on the AUC metric for the \tasks~with~\fine~(first row), \MAS~(second row) and \LwF~(third row). Due to space limit, performance based on other metrics is shown in~\supp. 
In each row, we show three sub figures:~Combined AUC (C.AUC), the current stage AUC (R.AUC) and the previous stages AUC (P.AUC). 
The achieved novelty detection performance is particularly high for \odin and \nav,  which leverage directly the multi-head nature by relying on the predicted class probability. 
For \mah, that operates in the learned feature space and hence does not leverage the separate output layers, the most recent task performance remains high through the sequence while the previous tasks AUC fluctuates as different tasks are being seen, possibly linked to the various degree of relatedness among the learned tasks since the \Out set is composed of yet to be seen tasks samples. Note that for \mah the mean and covariance estimated on previous tasks remain unchanged and hence might be outdated when significant changes happen on the features. Moreover, in this sequence, the last task is \textit{SVHN} that is visually similar to its precedent \textit{Letters}, which explains why the  ND performance of the last stage is lower than previous ones.
While \vae has a reasonable performance on the first learned
task, it rapidly drops as more tasks are learnt due to catastrophic forgetting in the VAE itself.
It is worth noting that, the  AUC of the first task is high  for the \Out sets belonging to the next 4 tasks. However for the last 3 tasks of Action, SVHN, Letter \vae fails to identify their samples as \Out.
On another note,  \odin performance seems the best on the R.AUC and P.AUC metrics, nevertheless,  on  C.AUC it is lagging behind \nav and \mah, indicating that the predicted novelty scores heavily rely on the used output head.
\\ 
  \begin{figure*}[t]
   \vspace*{-0.8cm}
  \centering
    \subfloat[\ER~with 25s R.AUC~($\%$)]{{\includegraphics[width=.49\textwidth]{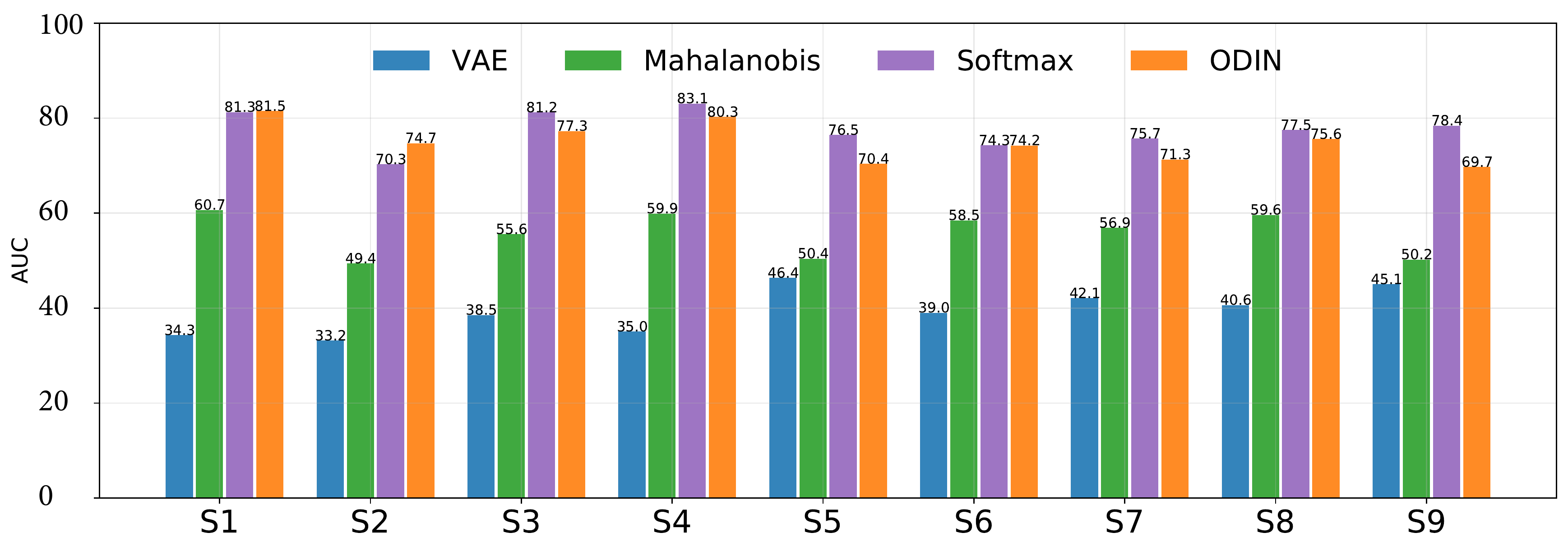}}}
        \hfill
    \subfloat[\ER~with 50s R.AUC~($\%$)]{{\includegraphics[width=.49\textwidth]{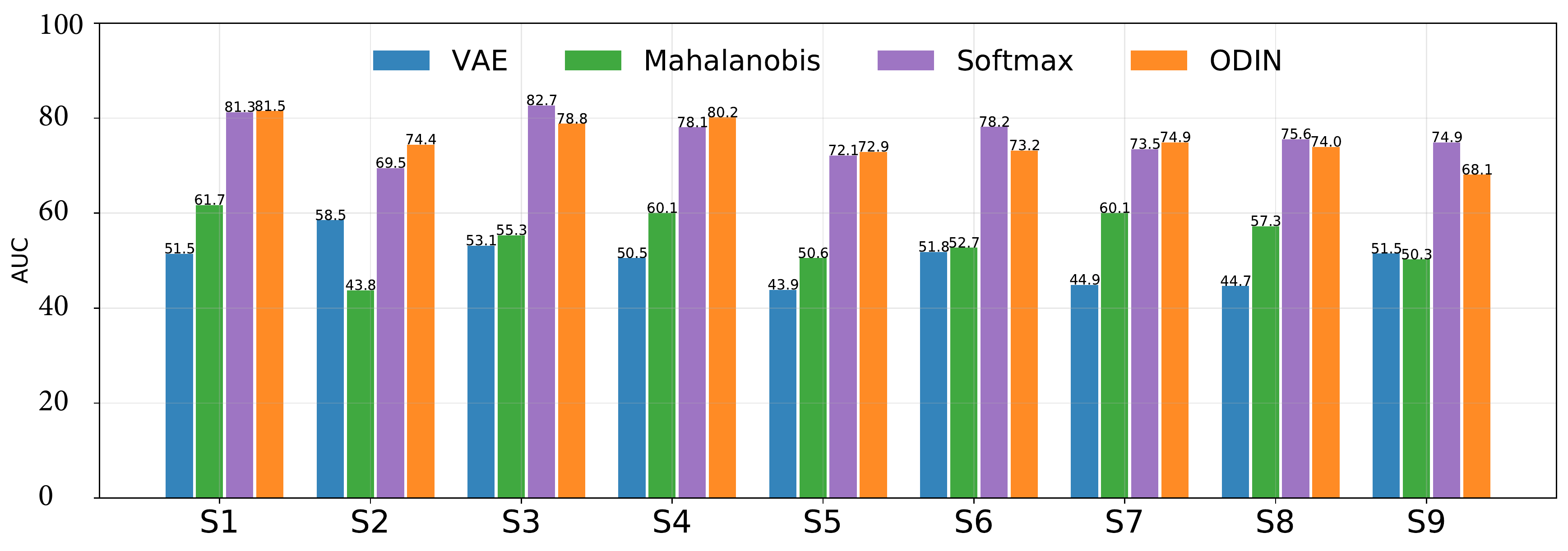}}}
    \vfill
    \centering
    \subfloat[\ER~with 25s P.AUC~($\%$)~]{{\includegraphics[width=.49\textwidth]{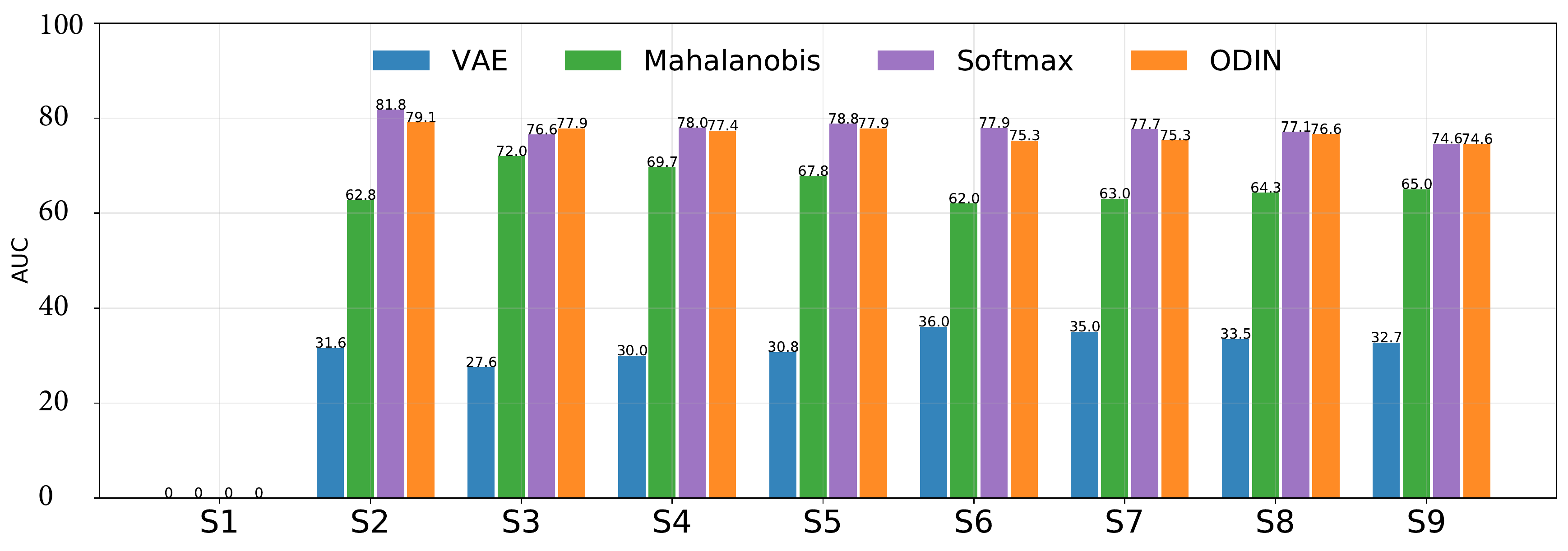}}}
        \hfill
    \subfloat[\ER~with 50s ~P.AUC~($\%$)~]{{\includegraphics[width=.49\textwidth]{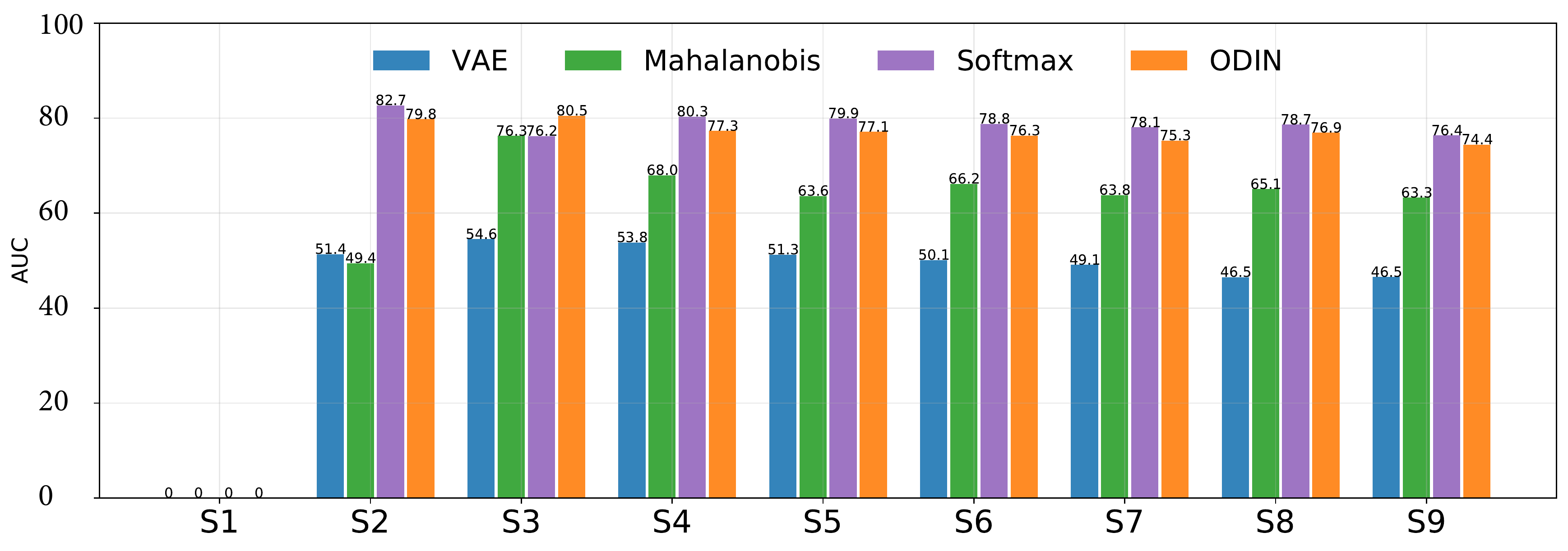}}}
     \vspace*{-0.3cm}
\caption{ \label{fig:shared_head_currenttasks}\small First row shows current stage {\footnotesize~(R.AUC)}, second row shows previous stages{\footnotesize~(P.AUC)} for \tinyimg~with ~\ER~ under shared head setting with two different buffer sizes, 25 samples and 50 samples per class.}
   
\end{figure*}
  \begin{figure*}[t]
   \vspace*{-0.3cm}
  \centering
    \subfloat[\SSIL~with 25s R.AUC~($\%$)]{{\includegraphics[width=.49\textwidth]{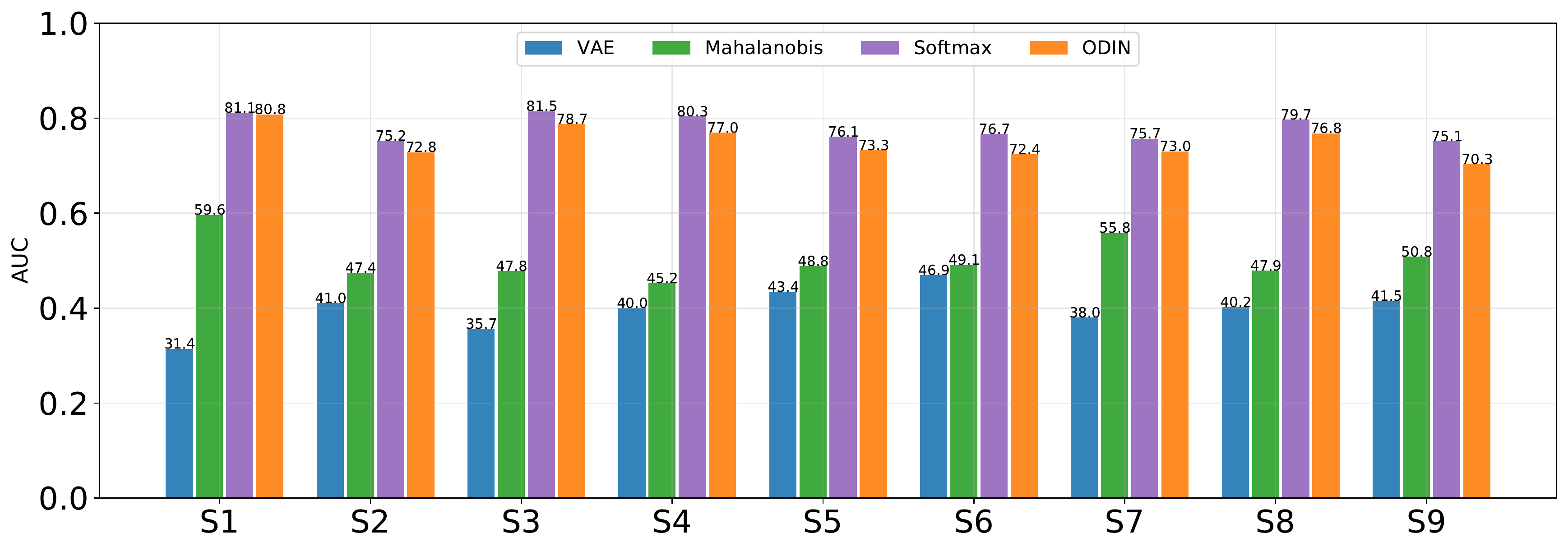}}}
        \hfill
    \subfloat[\SSIL~with 50s R.AUC~($\%$)]{{\includegraphics[width=.49\textwidth]{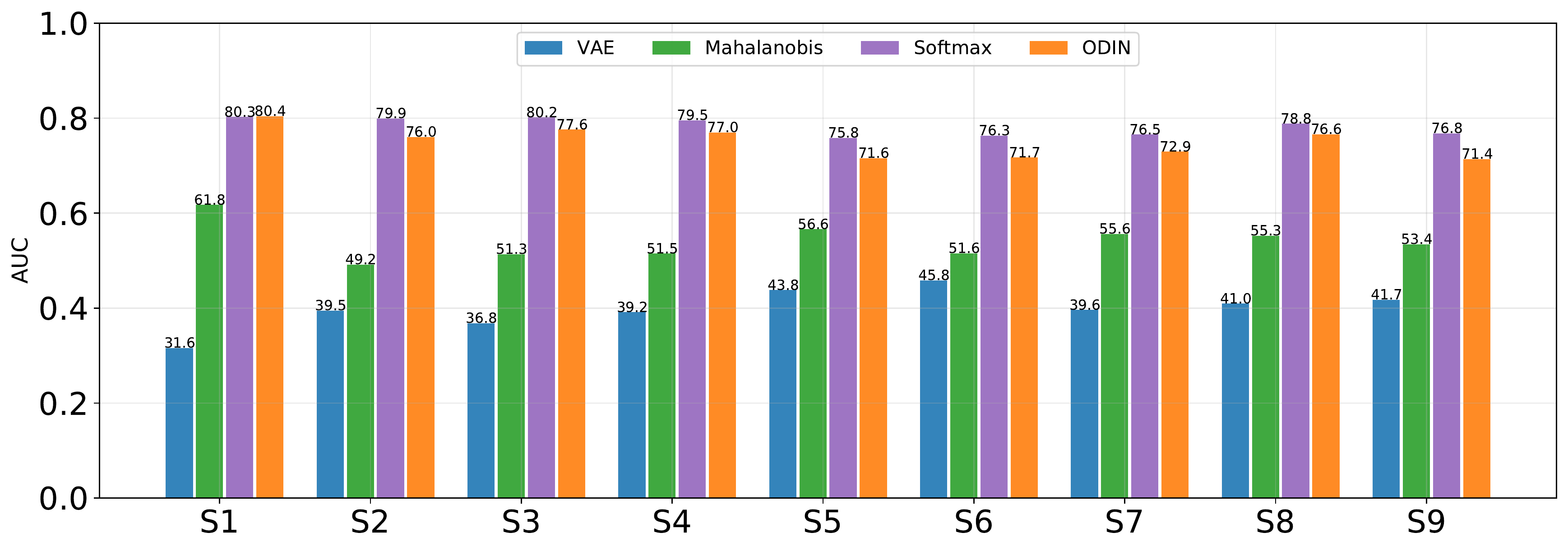}}}
    \vfill
    \centering
    \subfloat[\SSIL~with 25s P.AUC~($\%$)~]{{\includegraphics[width=.49\textwidth]{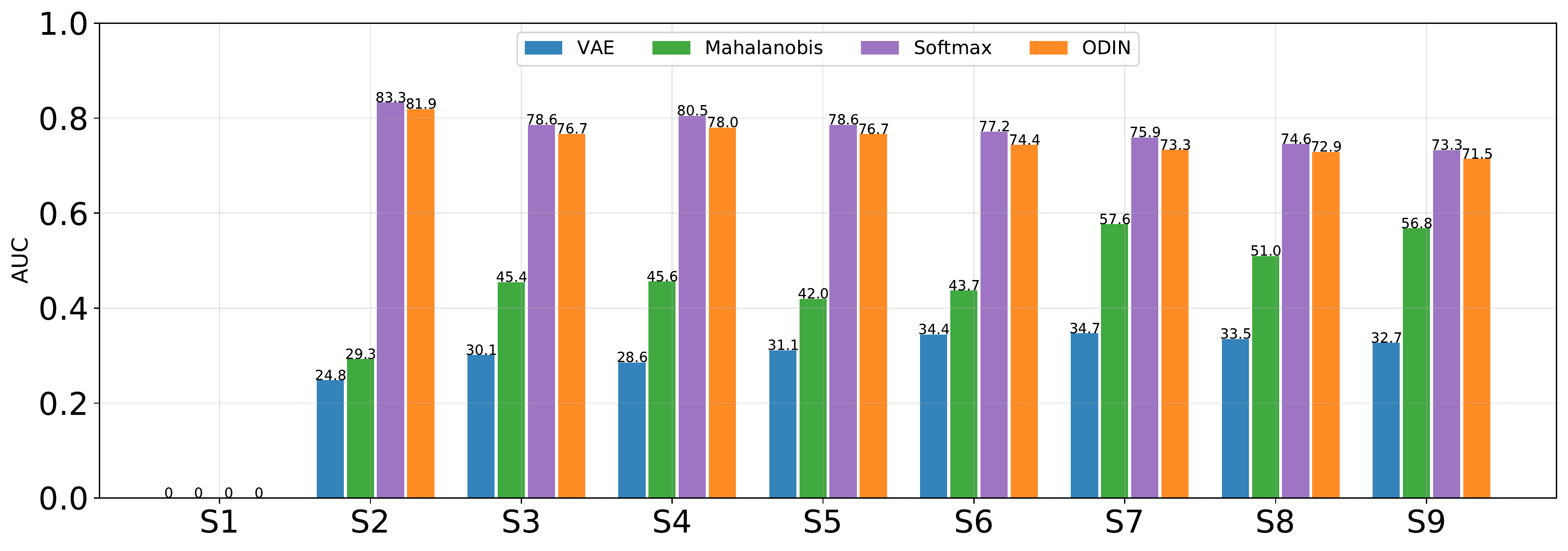}}}
        \hfill
    \subfloat[\SSIL~with 50s ~P.AUC~($\%$)~]{{\includegraphics[width=.49\textwidth]{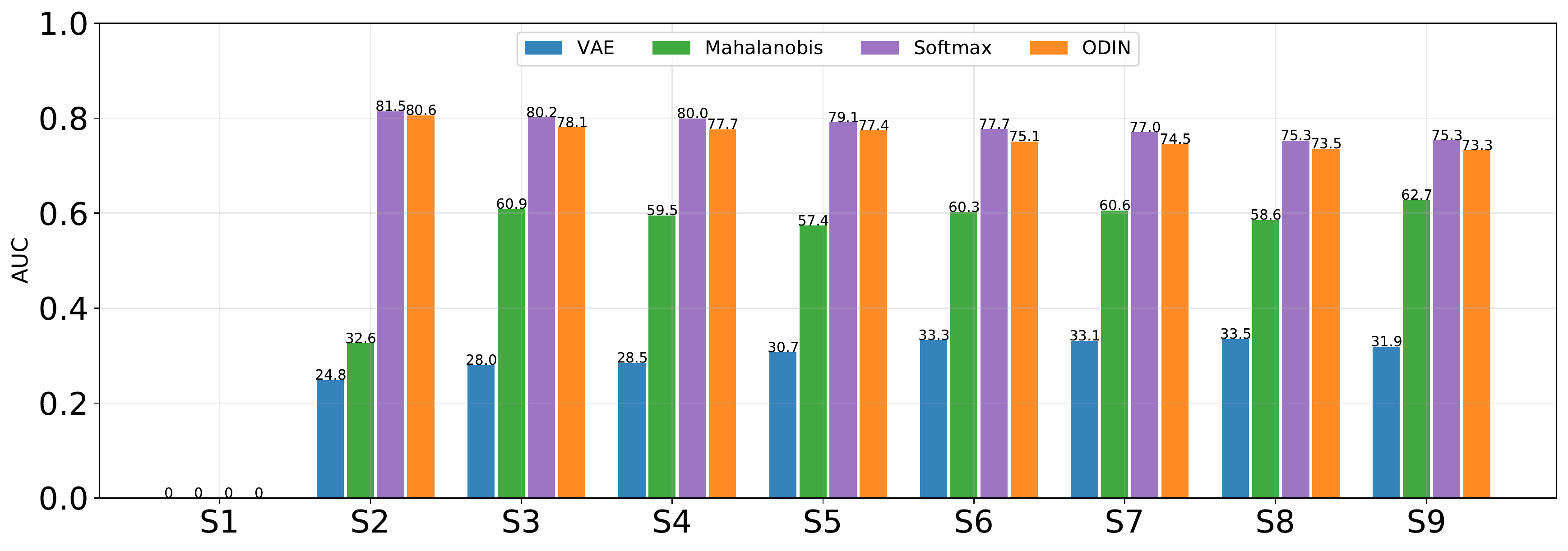}}}
     \vspace*{-0.3cm}
\caption{ \label{fig:shared_head_currenttasks_SSIL}\small First row shows current stage {\footnotesize~(R.AUC)}, second row shows previous stages{\footnotesize~(P.AUC)} for \tinyimg~with ~\SSIL~ under shared head setting with two different buffer sizes, 25 samples and 50 samples per class.}
 \vspace*{-0.6cm}   
\end{figure*}
\myparagraph{TinyImageNet  sequence.}
Here we examine ND methods performance on~\tinyimg~where new categories are learnt with no  shifts in the input  domain.
Figure~\ref{fig:10_tasks_Fintune} shows the C.AUC, R.AUC and P.AUC for \fine~(first row), \MAS~(second row), \LwF~(third row).\\
We make the following observations. First, the overall ND performance of all methods is lower than the \tasks, demonstrating the difficulty of this scenario.
With \fine,  the R.AUC score is quite similar at the different stages in the sequence. However, the P.AUC score declines as new classes are learnt and it is significantly lower than the R.AUC.  With \MAS and \LwF aiming at reducing catastrophic forgetting,  P.AUC scores of ND methods are significantly higher than for \fine. For example, \nav P.AUC is $75\%/77\%$ for \MAS~/~\LwF at the last  evaluation stage compared to $64\%$ with \fine. Second, the R.AUC scores with \MAS~/~\LwF are generally lower than with \fine (R.AUC~$74\%/77\%$  compared to $80\%$ with \fine).
This indicates that, while preserving the performance on previous classes, the current stage suffers  learning difficulties, translated into a decline in the  ND  performance. Another manifestation of the famous stability/plasticity dilemma. 
With \LwF, the R.AUC scores seem slightly better than MAS, which is indicative of the ability of LwF to adapt more to the new learning stage and to leverage the benefits of transfer learning, especially when the learned classes are similar. \\
When examining the different ND methods performance we can note the following:  the simplest method \nav performs the best especially in the C.AUC, \odin has consistently lower C.AUC score than \nav, as also seen in the \tasks~evaluation.
For \mah and \vae, the performance is considerably worse than methods operating on the output layer probabilities (\odin and \nav).  Note that by operating directly on the output probability they benefit from the CL method, while \mah and \vae suffer from the CL constraints. 
 In \supp~ND  experiments are shown with  VGG as a backbone model. Similar trends can be noted, our observations seem to be independent of the deployed backbone. 
 
Overall, in spite of the fact that the \IN sets are constructed only from correctly predicted samples, a clear degradation in ND performance  can be seen as more learning steps are carried.
Our results show that the quality of the learning and the characteristic of the continual learning method affect directly the performance of the novelty detection method. 
\newline
     \begin{wraptable}{r}{0.6\textwidth}
\vspace{-0.5cm}
\centering
\footnotesize
\input{latex/Forgtab}
\caption{\small Detection errors~($\%$)~of \F  (F) vs. \IN  (N)   and \F~(F)~vs.~\Out~(O) in the studied settings. Due to space limits, we only show the first and the last two stages  demonstrating the  change. Colors are used to highlight the change across the sequence.  In \supp, we report all stages and other metrics.}
 \label{tab:ForgottenvsInvsOut}
 \vspace*{-0.7cm}
 \end{wraptable}
\myparagraph{Shared-head results.}
We move to a setting where some previously seen samples are maintained and a shared output layer is used for all the learned classes.\\ 
Figures~\ref{fig:shared_head_currenttasks} and ~\ref{fig:shared_head_currenttasks_SSIL} show the R.AUC and the   P.AUC metrics,  with~\ER~ and ~\SSIL~ respectively for different buffer sizes. We  do not report the C.AUC as
 this is a shared-head setting.
The further in the sequence, the more  the P.AUC scores decline, although a larger buffer mitigates this effect.
In \supp, we show the performance on each \IN set at the different learning steps. 
It can be noticed that even when a replay buffer is used, novelty detection performance on the first learning stage  is the highest among all other stages. In \supp, we show that   the AUC of the first \IN set  is higher than the second \IN set and the gap increases with increasing  the deployed buffer size. This indicates difficulties with learning the new data while maintaining  the performance on replayed samples. This is the case for both CL methods, \ER~ and \SSIL.\\ 
On  the  ND methods performance, in general, \odin R.AUC scores here are lower than \nav. However, the P.AUC scores for \odin and \nav are quite similar. For \mah, although it does not leverage the replay buffer, better performance with~\ER~ is shown compared to multi-head setting, indicating a better feature preservation. 
This could suggest that larger rates of forgetting  and shift in the features space might be hidden in the multi-head setting. \\
However when~\SSIL~ is used as a CL method, \mah performance is close to random and \odin performance is consistently lower than \nav. We hypothesize that the focus on the preservation of the Softmax probabilities in ~\SSIL~ comes at a cost of a weaker features reservation.

\myparagraph{General observations.}
We highlight the following points on all the studied settings:
1)~\textit{The further in the sequence, the worse is the ND performance}. The incremental nature of the learning and the suffered forgetting affect even the correctly predicted samples and the way they are represented.
2)~\textit{ND methods also need a mechanism to control forgetting}. What is captured by the ND method of old data can be outdated as a result of the continuous model's updates, hence unless the ND method uses directly the model output, its performance radically drops.
3)~\nav~\textit{the simplest method works best}. In spite of being the beaten baseline in the typical ND benchmarks, using the predicted class probability outperforms more complex choices. However, a large gap remains between the ND performance in the \tinyimg and that of OOD benchmarks suggesting a big room for  improvements.
4)~\textit{ND performance on the ``first'' stage data is better than later stages}.   The first stage is learnt alone,
while other stages  are performed under constraints to prevent forgetting, and for shared-head setting the reminder of the classes  (when first learned)  have to be discriminated from already well learned classes. Such bias  hinders the learning in the later stages.

\myparagraph{What happens to the forgotten samples?}
 We want to examine how forgotten samples are treated by the ND method. Since those samples belong to a previous \IN set, are they treated similarly to the other \IN samples that are correctly predicted? Or are they now considered by the ND method as \Out samples? Can a novelty detection method possibly identify the  \IN\ \vs\ \F\ \vs~\Out?
 Table \ref{tab:ForgottenvsInvsOut} reports the detection error (DER) of \IN~vs.~\F  and \F~vs.~\Out in the different settings. 
First of all, similar to previous observations,  it can be seen that in later learning stages, the detection errors get higher.
For the \tasks, forgotten samples can be easily discriminated from unseen task samples (\F\ \vs~\Out). This is the case mostly for  \odin, \nav and \mah. It indicates a difference in the output probability and the feature distribution between \F and \Out. Note that the \Out samples belong to completely different tasks in this sequence.  However, in all other cases, it is extremely hard to discriminate samples of \Out sets from \F set. Does it mean that those are treated similarly? If we look closely, \odin is constantly better in all cases than \nav suggesting that the perturbation added to the input affects more \Out samples than \F samples. This could  provide a direction on how the separation of \F samples from \Out samples could be  approached. \\
Regarding \IN\ \vs\ \F, we can see that while at the second stage the detection error might be reasonable and those two sets can be differentiated,  as new learning stages are carried  the detection  gets harder. It is also worth noting that only methods that utilize the output probability could tell the difference, as opposed to those that operate in the feature space, such as \mah, suggesting that forgetting at first is more present at the output layer. 
For \tinyimg, the multi-head setting with a dedicated CL method (\MAS or \LwF), shows clearly better detection performance of \IN\ \vs\ \F for \nav and \odin when compared to \fine.
Finally, the shared head setting poses the most challenging detection of Forgotten samples. In spite of \SSIL~showing better CL performance than \ER, ~\F~ samples are harder to discriminate from \IN samples when \SSIL~is deployed compared to  \ER.
%
 Overall, we believe that the study of the novelty detection in the continual learning scenario brings extra insights on the learning characteristics that might be hidden when looking merely at the continual learning performance.

%% file: latex/Forgtab.tex
\resizebox{0.6\textwidth}{!}{
 \begin{tabular}{ |c|c|c|}
 \cellcolor{green!20} low DER $< 20$ &\cellcolor{yellow!20}medium DER $>20~\&<35$ &\cellcolor{red!20}high DER $>35$\\
\end{tabular}}
   \resizebox{0.6\textwidth}{!}{
\begin{tabular}{ |c|c|cc|cc||c|  }
 \hline
Setting& Method&\multicolumn{2}{|c|}{$S_2$}& \multicolumn{2}{|c|}{$S_{T-1}$ }&\multicolumn{1}{|c|}{$S_T$}  \\
 \hline
& &F. \vs N.&F \vs O.&F. \vs N.&F \vs O.&F. \vs N.\\
 \hline
 \multirow{4}{4em}{Eight-Tasks \fine}   &\nav& \cellcolor{green!20}19.2 & \cellcolor{green!20}17.5&\cellcolor{yellow!20}33.5&\cellcolor{red!20} 35.3&\cellcolor{red!20}36.6 \\
&\odin &\cellcolor{yellow!20}32.5 &\cellcolor{green!20} 7.6&35.7&\cellcolor{green!20} 4.2&\cellcolor{red!20}36.9   \\
&\mah  &\cellcolor{red!20} 49.9 & \cellcolor{green!20}16.8&49.1 & \cellcolor{red!20}46.6&\cellcolor{red!20}49.4   \\
&\vae & 50.0 & 50.0&48.8& 50.0&49.5   \\
\hline
\multirow{4}{4em}{Eight-Tasks MAS} & \nav &\cellcolor{green!20} 7.6 &\cellcolor{green!20} 12.4&21.3 & \cellcolor{green!20}14.6&\cellcolor{yellow!20}22.3  \\
  &\odin &\cellcolor{yellow!20} 26.9 &\cellcolor{green!20} 3.2&31.6 & \cellcolor{green!20}2.1&\cellcolor{yellow!20}31.2 \\
  &\mah &\cellcolor{red!20} 49.9 &\cellcolor{green!20} 10.2&48.4 &\cellcolor{red!20} 38.2&\cellcolor{green!20}11.8\\
  &\vae & 49.8 & 50.0&48.8 & 50.0&49.0   \\  
\hline
\multirow{4}{4em}{Eight-Tasks \LwF}   &\nav &\cellcolor{green!20} 15.4 & \cellcolor{green!20} 15.3&\cellcolor{yellow!20}20.0 &\cellcolor{green!20} 16.2&\cellcolor{yellow!20}24.7 \\
  &\odin &\cellcolor{yellow!20} 30.7 & \cellcolor{green!20}10.5&33.1 & \cellcolor{green!20}1.3&\cellcolor{red!20}35.0\\
  &\mah & \cellcolor{red!20}49.8 & \cellcolor{green!20} 19.7&48.8 & \cellcolor{red!20}49.9&43.3 \\  
   &\vae & 49.8 & 50.0&48.8 & 50.0&49.0      \\  
   \hline
\hline
\multirow{4}{4em}{TinyImag. Seq.  \fine} &\nav  & \cellcolor{yellow!20} 31.3 & 48.3&35.6 & 49.4&\cellcolor{red!20}36.4 \\ 

& \odin & \cellcolor{yellow!20}33.9 & 46.7&36.5 & 49.3&\cellcolor{red!20}36.2    \\ 

&\mah &   \cellcolor{red!20}44.8 & 46.2&45.9 & 48.0&\cellcolor{red!20}45.5  \\ 
&\vae &  46.7 & 47.6&47.2 & 47.9&46.7 \\ 
\hline
\multirow{4}{4em}{TinyImag. Seq. \MAS} & \nav  & \cellcolor{green!20} 15.4 & 49.7&18.9 & 48.6&\cellcolor{green!20} 16.3   \\ 

& \odin  &\cellcolor{yellow!20} 24.4 & 48.0&29.0 & 47.0&\cellcolor{yellow!20} 27.0   \\ 

&\mah & \cellcolor{red!20}42.9 & 43.8&43.2 & 44.4&\cellcolor{red!20} 43.0  \\ 
&\vae &  46.7 & 47.6&47.2 & 47.9&46.7  \\ 
\hline
\multirow{3}{4em}{TinyImag. Seq. \LwF} & \nav&\cellcolor{yellow!20} 20.6 & 49.5&8.5 & 48.3&\cellcolor{green!20}17.9   \\ 

& \odin & \cellcolor{yellow!20}24.2 & 49.4&26.6 & 46.1&\cellcolor{yellow!20}26.5  \\ 

&\mah &  \cellcolor{red!20}45.7 & 46.0&43.2 & 47.8&\cellcolor{red!20}44.9   \\ 
&\vae &  46.7 & 47.6&47.2 & 47.9&46.7 \\
   \hline
   \hline
   
\multirow{4}{4em}{TinyImag. Seq.  \ER~25s}&\nav &\cellcolor{yellow!20} 21.3 & 46.3&30.6 & 46.5&\cellcolor{yellow!20}30.8    \\  
&\odin &\cellcolor{yellow!20} 29.6 & 40.3&32.3 & 45.5&\cellcolor{yellow!20}32.1   \\  
&\mah & \cellcolor{red!20}37.5 & 46.1&39.4 & 47.0&\cellcolor{red!20}38.5 \\
&\vae & 45.3 & 45.1&47.2 & 48.6&46.8   \\ 
\hline
\multirow{4}{4em}{TinyImag. Seq.  \ER~50s} & \nav &\cellcolor{green!20} 19.9 & 46.4&27.3 & 46.9&\cellcolor{yellow!20}29.5   \\ 
&\odin & \cellcolor{yellow!20}28.2 & 45.5&32.4 & 45.2&\cellcolor{yellow!20}31.9   \\
&\mah & \cellcolor{red!20}47.3 & 43.8&40.0 & 45.4&\cellcolor{red!20}39.3  \\  
&\vae & 45.3 & 45.1&47.2 & 48.6&46.8   \\ 
\hline
\hline
\multirow{4}{4em}{TinyImag. Seq. \SSIL~25s} & \nav &  \cellcolor{yellow!20}27.7 & 43.1&32.9 & 42.1&\cellcolor{yellow!20}33.0 \\
&\odin & \cellcolor{yellow!20} 30.7 & 38.9&38.0 & 39.1&\cellcolor{red!20}37.2 \\
&\mah &\cellcolor{red!20} 49.1 & 49.6&45.5 & 48.4&\cellcolor{red!20}447.1 \\
&\vae & 45.3 & 48.6&49.0 & 44.6&47.3 \\
\hline
\multirow{4}{4em}{TinyImag. Seq. \SSIL~50s} 
&\nav &  \cellcolor{yellow!20}24.0 & 46.3&30.7 & 39.6&\cellcolor{yellow!20}34.1 \\
&\odin & \cellcolor{yellow!20} 28.7 & 41.5&36.4 & 37.9&\cellcolor{red!20}38.3  \\
&\mah & \cellcolor{red!20}49.1 & 48.4&46.4 & 48.0&42.3  \\
&\vae & 49.9 & 49.1&48.5 & 49.2&49.2  \\
\hline
\end{tabular}
 }

%% file: sections/discussion.tex
  \begin{figure*}[t!]
  \vspace*{-0.6cm}
  \centering
    \subfloat[{\footnotesize Eight-task: In vs. Out }]{{\includegraphics[width=.24\textwidth]{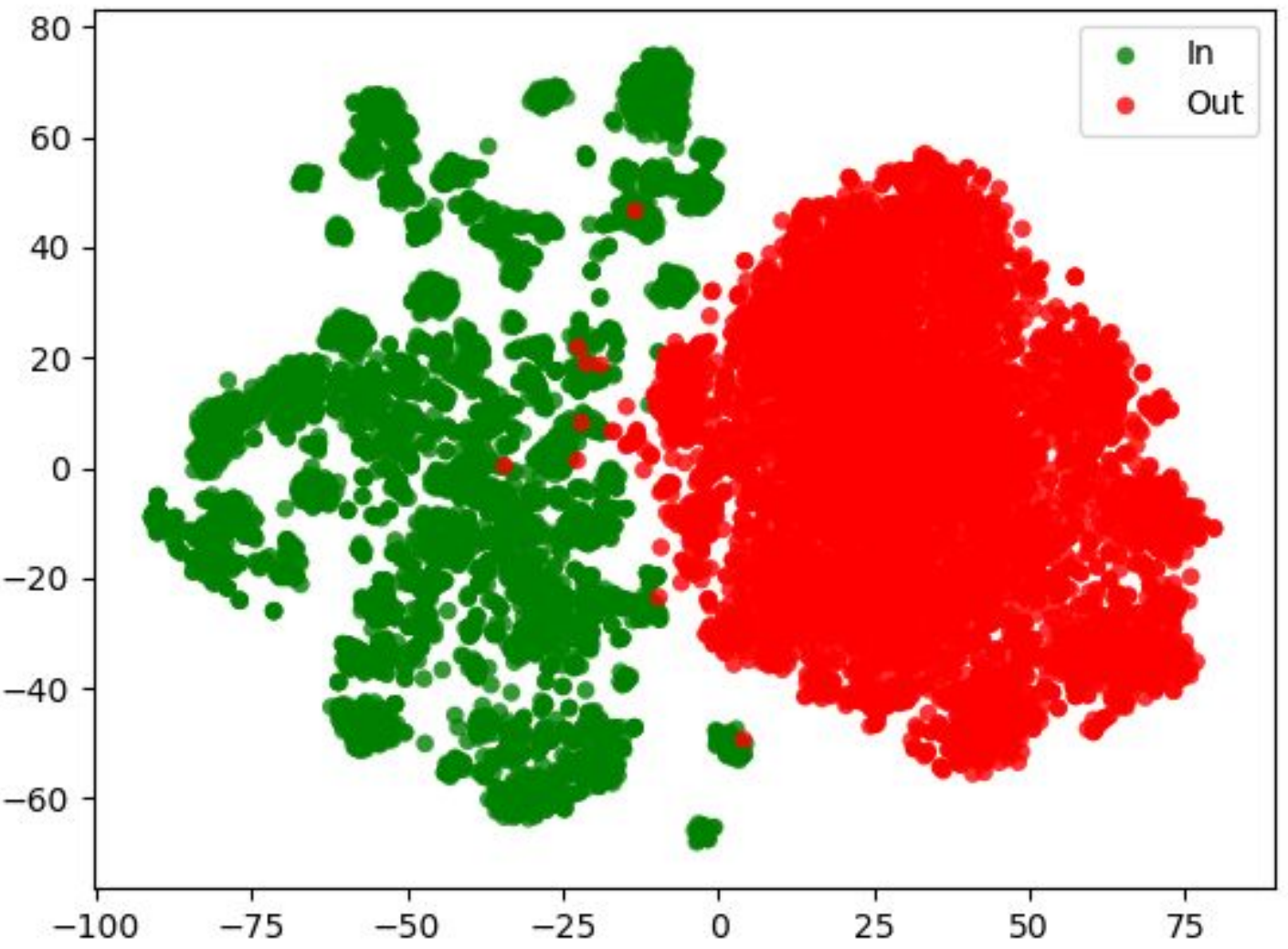}}}
        \hfill
    \subfloat[{\footnotesize Eight-Task: In vs. Forg. }]{{\includegraphics[width=.24\textwidth]{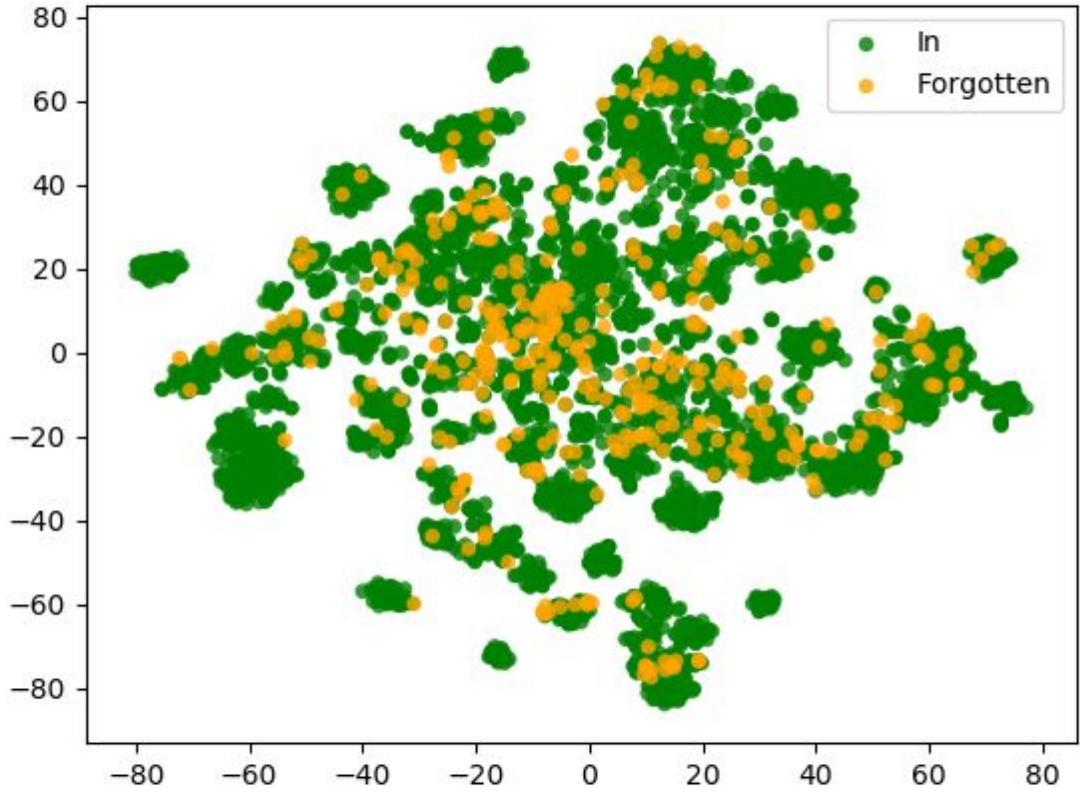}}}
    \hfill
    \subfloat[{\footnotesize TinyImageNet: In vs. Out}]{{\includegraphics[width=.24\textwidth]{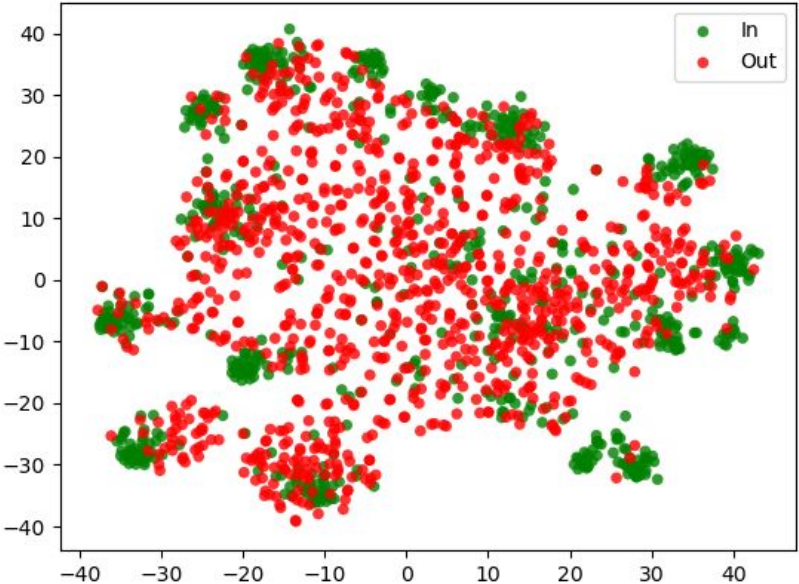}}}
        \hfill
    \subfloat[{\footnotesize~TinyImageNet: In vs. Forg. }]{{\includegraphics[width=.24\textwidth]{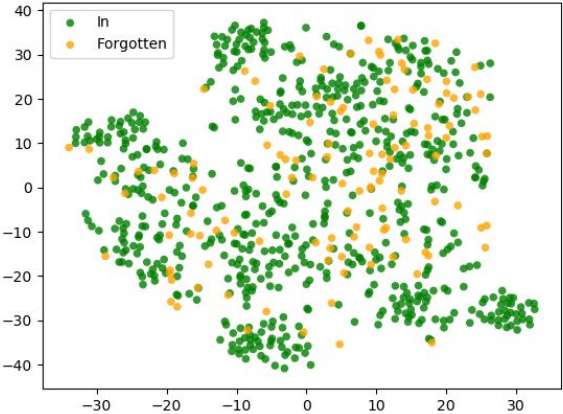}}}
\vspace*{-0.2cm}
\caption{\small tSNE 2D visualization. (a) and (c) show the projection of {\tt IN} and {\Out} samples based on $\mathcal{M}_1$, (b) and (d) show  the sets of {\tt IN} and {\F}  based on  $\mathcal{M}_2$ for multi-head \tinyimg and \tasks.} 
\vspace*{-0.6cm}
    \label{fig:tsne_visualization}%
\end{figure*}
\section{Discussion and Additional Baselines\label{sec:discussion}}
\myparagraph{tSNE visualization. } With the goal of understanding how the features of the 3 sets are distributed and whether this relates to the nature of the sequence (disjoint tasks vs. different groups of classes), we examine the 2D tSNE projections  of the (\IN, \Out, \F). 
Figure~\ref{fig:tsne_visualization} shows the tSNE projections of the first (\IN, \Out, \F) sets for \tasks~and multi-head   \tinyimg~based on the last layer features. It can be seen that in the later case, the \Out samples are projected in between the different clusters of learned categories as opposed to the {\Out} set, being a different dataset, and thus a different domain. This might explain the low ND performance in \tinyimg compared to \tasks. 
In both settings, the forgotten samples  appear on the boundaries of a predicted category cluster after a movement from one cluster to another that possibly caused  mistaken  predictions (forgetting). \newline
\myparagraph{A similarity based view.} 
~\cite{boudiaf2020unifying} shows that the optimal weights of the last classification layer  are proportional to the features mean. 
Similar analogies between the last layer class weights and the class feature mean are made in~\cite{lee2018simple,qi2018low}.
In continual learning, access to old data can be prohibited but the learned features of all categories are frequently updated. As such, we propose to exploit the analogy between the learned classes weights and the features, and treat the last layer class weights as prototypes of the learned classes. Instead of solely relying on the predicted class probabilities, we can directly leverage the similarities of an input sample to different class prototypes. 
We consider a neural network with  no bias term in the last layer.
Consider an input sample $x$ and a neural network  $\{\phi,\theta\}$ with $\phi$ the feature extraction layers and a fully-connected final classification layer represented by  $\theta\in\mathbb{R}^{k\times n}$, where $n$ is the dimensionality of features $\phi(x)$ and $k$ is the number of classes.
The $i^\text{th}$ column of $\theta$ is denoted as $\theta_i$.
The network output before the softmax is $\phi(x)\theta\transpose$ and the predicted class satisfies $c=\argmax_i(\phi(x)\cdot\theta_i)= \argmax_i(\norm{\phi(x)}\norm{\theta_i}\cos(\alpha_i))$, where $\alpha_i$ is the angle between  $\phi(x)$ and $\theta_i$. 
The norm of an extracted feature $\phi(x)$ and the cosine of the angle between $\phi(x)$ and the most similar class are \emph{data-dependent terms}~\ie, the scalar projection of $\phi(x)$ in the direction of $\theta_c$.
In \supp, we investigate  statistics of  $\phi(x)$  (mean, std., norm) for the studied sets at different stages. We note that after the first stage, the feature norm of \IN is significantly larger than that of \Out. However, in the later learning stages the difference disappears.  
This might be a possible cause for the \emph{degradation} of  \nav performance in the later stages. Particularly, \F samples show low features norm compared to \IN samples. 
In the following, we suggest two additional baselines leveraging the similarities between the features and the class prototypes.\\
\myparagraph{Baseline~1.}
Given an input  $x$, the estimated class probability vector is $p_x=[p_x^1,p_x^2,\dots, p_x^k]$, with $p_x^c=\frac{e^{\phi(x)\cdot\theta_c}}{\sum_i{e^{\phi(x)\cdot\theta_i}}}$.
We hypothesize that \IN samples can be better explained by the class prototypes than those of \Out samples.
We thus estimate vector $z=p_x\theta^*$, where $\theta^*\in\mathbb{R}^{k\times n}$ is the matrix of normalized class prototypes.
Denoting the normalized sample feature by $\phi^*(x)\in\mathbb{R}^n$, we define a scalar score $s=\phi^*(x)\cdot z$,
which we refer to as \Ba and use it for novelty detection. Note that $s$ is equal to the weighted sum of cosine similarities between $\phi(x)$ and the prototypes from $\theta^*$.\\
\myparagraph{Baseline~2.}
To further exploit the features/prototypes  similarities, we leverage the difference between the estimated similarities to the $(n)$ closest class prototypes.
We suggest adding a perturbation to the features vector $\phi(x)$ corresponding to the gradient of the cross-entropy loss given a candidate class as a label.
With $y_n$ being  the  $n^\text{th}$ closest class, the estimated gradient (perturbation)~is~$ g_n=\nabla_{\phi(x)}\ell(\phi(x),y_n)$ 
and the updated feature vector is $\phi^\prime_x=\phi(x)- g_n$.
Then we estimate the difference $\Delta$ in the cosine similarity between  the predicted class $c$ and the $n^\text{th}$ closest class as $\Delta=\cos(\alpha_c)-\cos(\alpha_n)$.
We add this score to the score of \Ba and refer to it as \Bb.\\
   \begin{wraptable}{r}{0.55\textwidth}
\vspace{-0.6cm}

\input{latex/BaselinetabShort}
  \vspace*{-0.2cm}
 \caption{\label{tab:baseline} \small ND performance~(TinyImageNet Seq.) expressed in the mean of each measure~$\%$~ w.r.t. \nav and our  baselines (\Ba,\Bb).}
 \vspace*{-1.3cm}
  \end{wraptable}
\myparagraph{Results.} We retrained the models in \tinyimg with no bias term and obtained similar CL performance, we refer to \supp for  additional details and notes on the shared-head setting.
Table~\ref{tab:baseline} compares the ND performance of our proposed baselines to \nav, the best  studied method. Our baselines show better performance than \nav in the different settings especially in the case of severe forgetting with \fine.
Our proposed baselines serve to  stimulate investigating diverse directions.
We believe that the ideal solution directly couples  an ND module and a CL model, which could better exploit any discriminative patterns of the \IN set.

%% file: latex/BaselinetabShort.tex
\resizebox{0.55\textwidth}{!}{
\small
\begin{tabular}{ c|c|c|c|c||c|c  }
Setting & Method & m.P.AUC $\uparrow$ &m.R.AUC$\uparrow$   & mAUC$\uparrow$ &F/IN$\downarrow$ &F/Out$\downarrow$\\ \hline
\multirow{3}{4em}{Multi-head\\\fine} 
&\nav& 66.1 & 81.6 &  71.0 & 31.0 & 43.8  \\ 
&\Ba& 67.7 & 83.0 &  72.5  & 30.8 & 43.6  \\
& \Bb  & 68.1 & 83.2  & 72.8  & 30.2 & 43.8   \\  
\hline
\multirow{2}{4em}{Multi-head\\MAS} 
&  \nav& 78.1 & 76.7  & 78.2  & 14.5 & 43.4  \\
&\Ba  & 78.4 & 77.1  & 78.5  & 22.8 & 40.8   \\
&\Bb & 79.1 & 77.7  & 79.2 & 20.6 & 41.9\\
\hline
\multirow{2}{4em}{Multi-head\\LWF}
 &\nav  & 79.7 & 80.5  & 80.3 & 17.1 & 42.9   \\
 &\Ba  & 80.3 & 82.2  & 81.1  & 22.6 & 39.8       \\
 &\Bb  & 80.8 & 82.4  & 81.6  & 20.4 & 41.0 \\
 \hline

   \multirow{2}{4em}{Shared-head\\ER-25s}
  & \nav & 78.4 & 70.5  & 76.6 & 22.8 & 42.2  \\ 
 & \Ba  & 78.7 & 69.7 & 76.4  & 25.5 & 41.0 \\
 &\Bb&  79.8 & 71.4  & 77.7  & 23.7 & 41.4  \\
 \hline
  \multirow{2}{4em}{Shared-head\\ER-50s}
  &\nav & 80.1 & 69.8  & 77.8  & 22.2 & 41.0   \\
  &\Ba & 79.1 & 66.1  & 76.1  & 25.9 & 39.2 \\
  &\Bb  & 80.7 & 68.9  & 78.0  & 23.5 & 40.1     \\ \hline
   \multirow{2}{4em}{Shared-head\\SSIL-25s}
 &\nav & 77.6 & 78.6 & 77.9  & 27.9 & 36.9\\
 &\Ba & 76.5 & 73.0 & 75.8  & 32.1 & 34.6\\
&\Bb & 78.0 & 70.6 & 76.5  & 29.3 & 35.9\\
 \hline
\multirow{2}{4em}{Shared-head\\SSIL-50s}
 &\nav & 78.1 & 77.6  & 78.0  & 27.9 & 37.2\\
 &\Ba & 77.5 & 72.0  & 76.4  & 32.2 & 34.6\\
&\Bb & 78.5 & 69.2  & 76.7  & 29.8 & 36.1\\

 \end{tabular}}

%% file: sections/conclusion.tex
\vspace*{-0.2cm}
\section{Conclusion}
\label{sec:conclusion}
We introduce the problem of continual novelty detection, a problem that emerges when novelty detection  techniques are deployed in a continual learning scenario. We divide the detection problem into three sub-problems, \IN~\vs~\Out, \IN~\vs~\F and \F~\vs~\Out. We examine several  novelty detection techniques on these sub-problems in various  continual learning settings. We show an increase in the detection error of \IN~\vs~\Out as a result of continual learning. Surprisingly,   the most simple ND method,~\nav, works best, yet poorly, in the studied  settings opening the door for more sophisticated methods to tackle this challenging problem. When investigating how the \F set is treated, it appears that as  learning stages are carried, the confusion with \IN set increases. However, unless the \Out set is composed of a different dataset,  it is extremely hard for current ND methods to discriminate  \IN from \F.
We present baselines improving over  \nav  in an attempt to stimulate further research in this problem.

%% file: sections/ethics.tex
\section{Ethics in Continual Novelty Detection}
\label{sec:ethics}
This work introduces the notion of applying novelty detection techniques in the setting of continual learning. As such, ethical implications of both aspects need to be pondered upon in this setting. Regarding novelty detection, the implications are rather positive, as it serves to bring more transparency to whichever knowledge has been embedded in the model. This, in turn, may help to alleviate biases introduced in the training set, or by the unbalanced nature of the data.
As for the continual learning aspect of it, we show that the rate of samples that are forgotten as the model learns new tasks gets invariably larger. In our study, we assume that the different tasks share the same importance. However, we foresee the need for methods that are able to continually adjust the relative importance of each task. How this importance is assigned, and the nature of the tasks, may have ethical implications.

%% file: sections/appendix.tex
\subsection{Introduction}
In this appendix, we show the following: 1) For each continual learning method and under the different considered settings, we report each task Accuracy at the end of the sequence, Subsection~\ref{sec:cl_per}.  2) We report the novelty detection performance on \tinyimg multi-head when using VGG as a backbone model, Subsection~\ref{sec:vgg}. 3) We report the novelty detection performance with AUPR-IN and DER metrics, Subsection~\ref{sec:othermetrics}. 4) We report the novelty detection performance on each  \IN set  before the last learning stage, Subsection~\ref{sec:laststage}. 5) We report the full \F~\vs~\IN and \F~\vs~\Out scores with different metrics, Subsection~\ref{sec:forg}. 
6) We report statistics of the learnt features under the different settings, Subsection~\ref{sec:discussion}.
\subsection{ CL Per Task Performance\label{sec:cl_per}} 
\begin{figure*}[t]
  \vspace*{-0.2cm}
    \centering
    \subfloat[{\footnotesize ~\tinyimg: Multi-head ResNet}]{{\includegraphics[width=.49\textwidth]{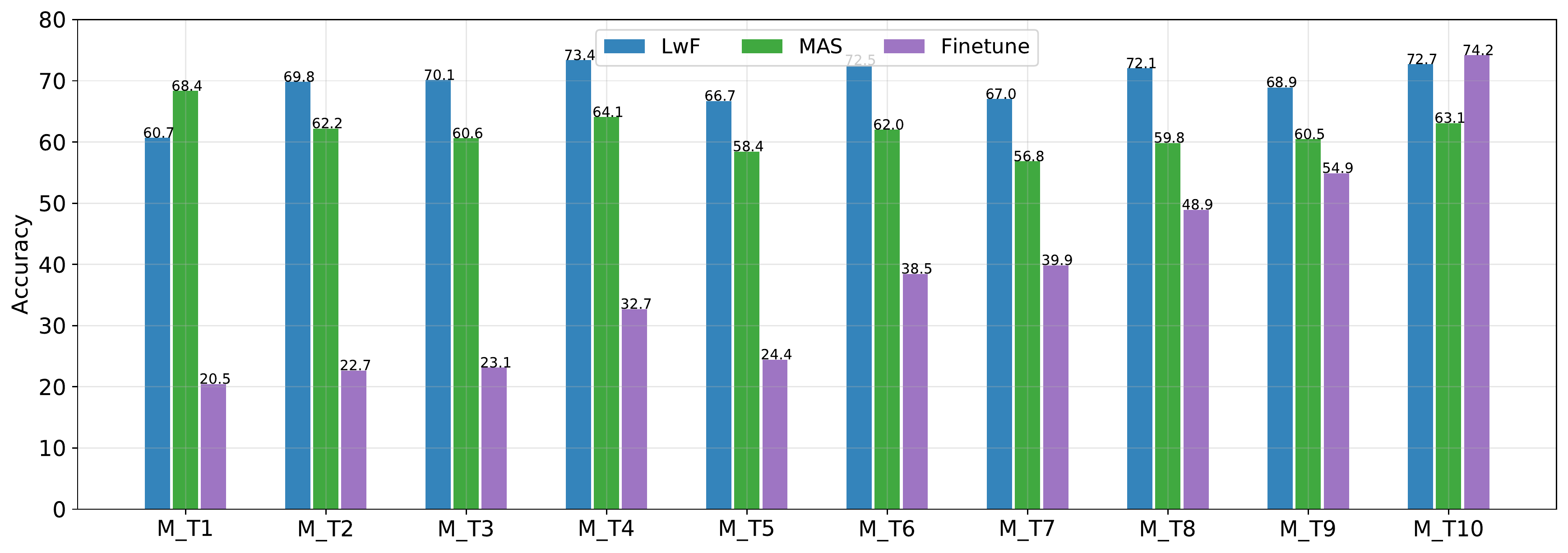}} }%
     \hfill
    \subfloat[{\footnotesize ~\tinyimg: Multi-head VGG}]{{\includegraphics[width=.49\textwidth]{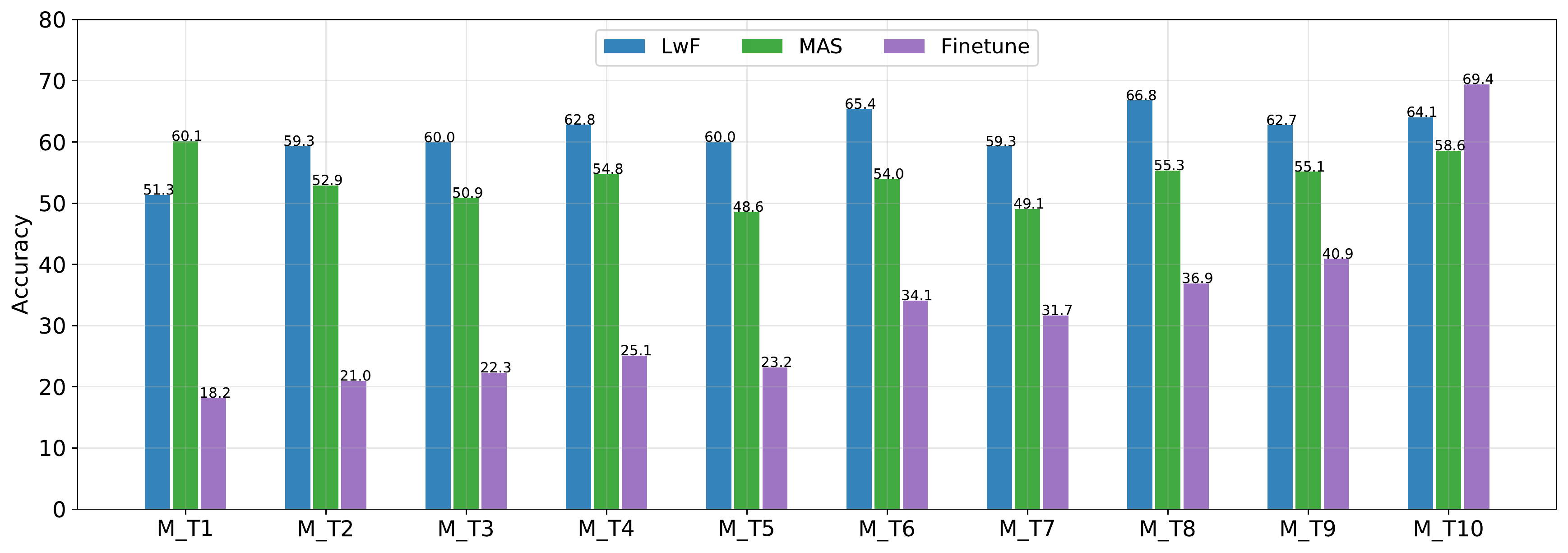}}}
\caption{\footnotesize   Seen tasks accuracy at the end of~\tinyimg with ResNet and VGG as backbones.}%
    \label{fig:multihead_CL}%
    \end{figure*}

\begin{figure*}[h!]
  \vspace*{-0.2cm}
    \centering
    \subfloat[{\footnotesize Shared-head ResNet}]{{\includegraphics[width=.49\textwidth]{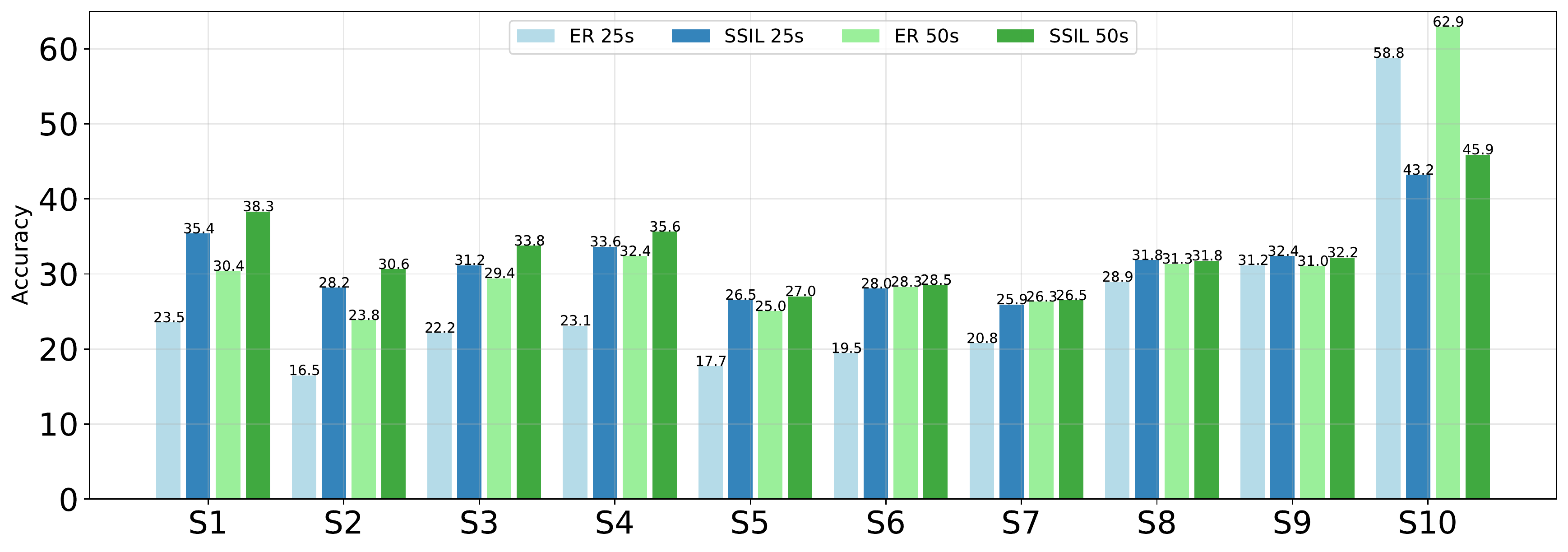}} }%
     \hfill
    \subfloat[{Multi-head \tasks}]{{\includegraphics[width=.49\textwidth]{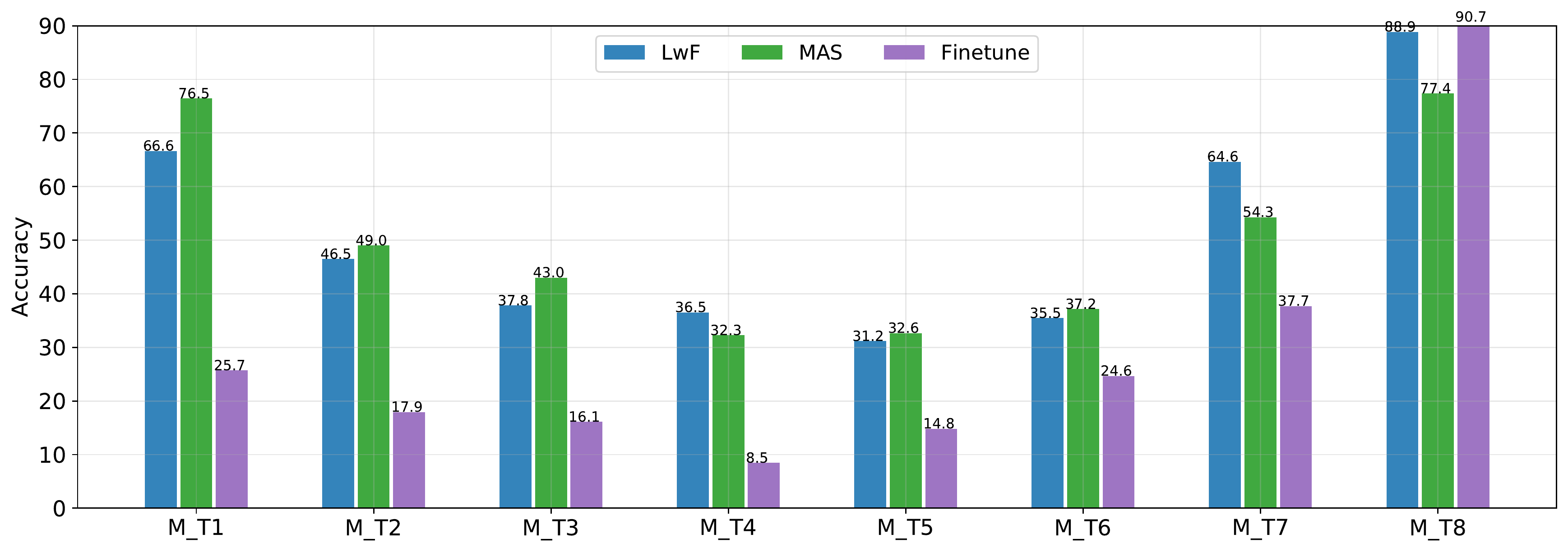}}}
\caption{\footnotesize  (a) Seen tasks accuracy at the end of~\tinyimg with ResNet as a backbone and shared-head setting. (b) Seen tasks accuracy at the end of~\tasks with multi-head and AlexNet as a backbone.}%
    \label{fig:shared_CL}%
    
    \end{figure*}

In the main paper, we have reported the average accuracy and the average forgetting of the various CL settings and deployed methods. Here we report the average accuracy of each task at the end of the sequence for the different settings (Figures~\ref{fig:multihead_CL}~\&~\ref{fig:shared_CL}).  These results support the discussions and the insights on the differences between MAS and LwF as different deployed continual learning methods. MAS is more conservative and tend to preserve the performance on the very first tasks while LwF is more flexible allowing more forgetting of the first tasks but obtaining better accuracy on average.
    
 \subsection{ ND Performance on Another Backbone.\label{sec:vgg}}    
\begin{figure*}[t]
  \vspace*{-0.2cm}
    \centering
    \subfloat[{\footnotesize C.AUC)~($\%$),~ \fine}]{{\includegraphics[width=.33\textwidth]{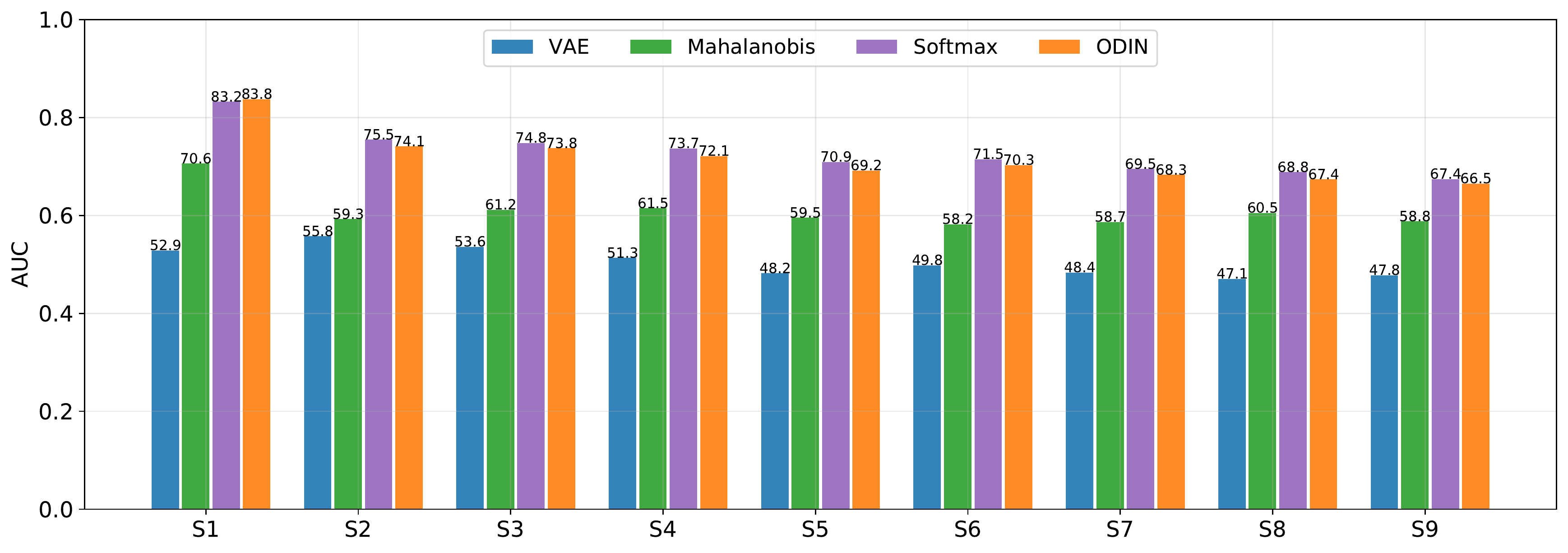}} }%
     \hfill
    \subfloat[{\footnotesize R.AUC~($\%$),~ \fine}]{{\includegraphics[width=.33\textwidth]{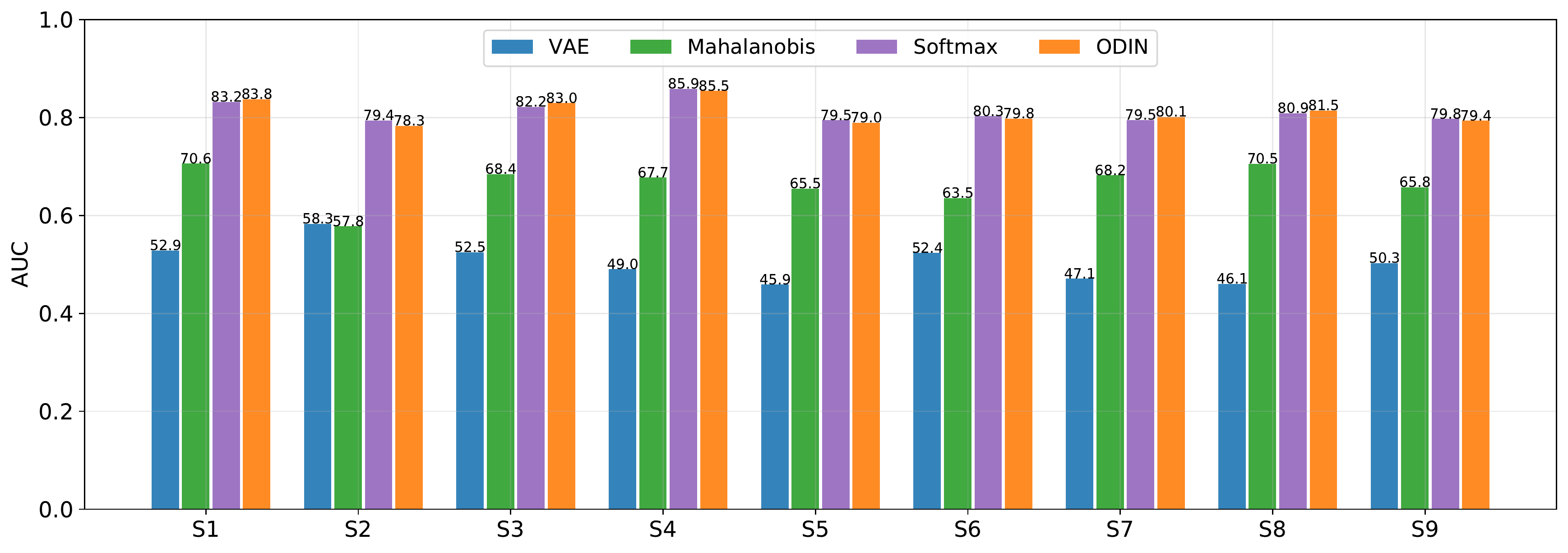}}}
        \hfill
    \subfloat[{\footnotesize P.AUC~($\%$),~ \fine}]{{\includegraphics[width=.32\textwidth]{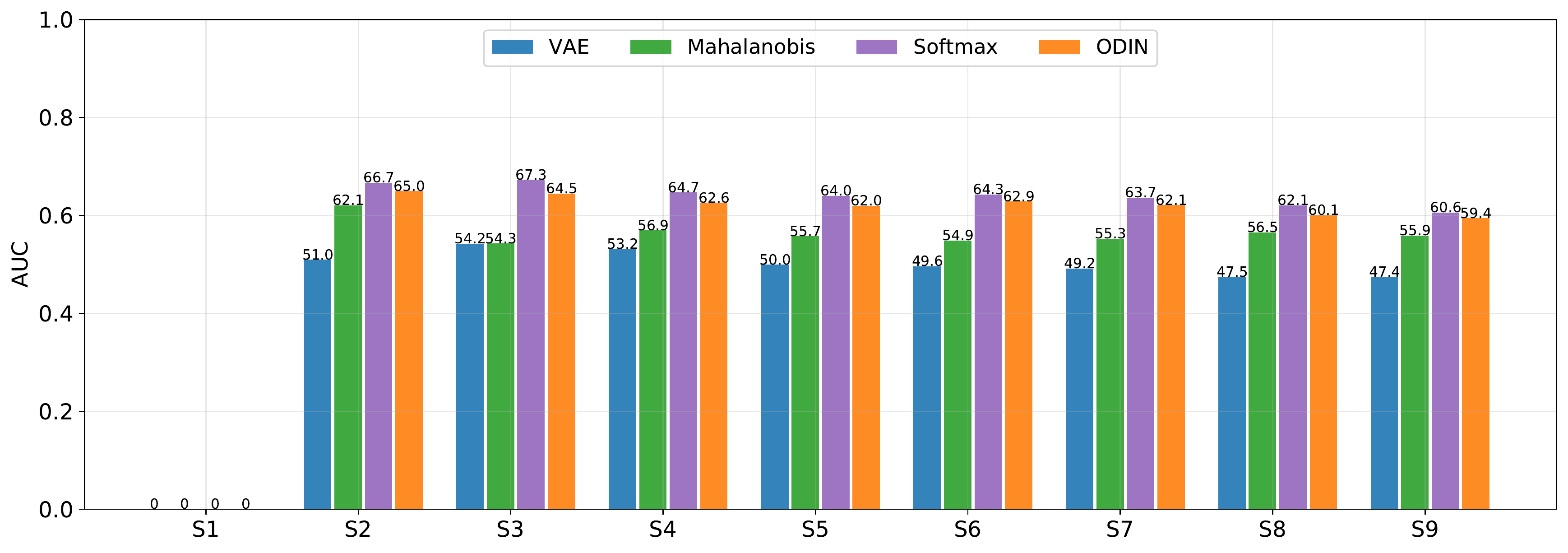}}}
\caption{\footnotesize Novelty detection  results in terms of Combined AUC~(C.AUC), Recent stage AUC~(R.AUC) and Previous stages AUC~(P.AUC) for \tinyimg with \fine and VGG network as a backbone.}%
    \label{fig:10_tasks_Fintune}%
    \end{figure*}

\begin{figure*}[h]
  \vspace*{-0.2cm}
    \centering
    \subfloat[{\footnotesize C.AUC~($\%$),~ \MAS}]{{\includegraphics[width=.33\textwidth]{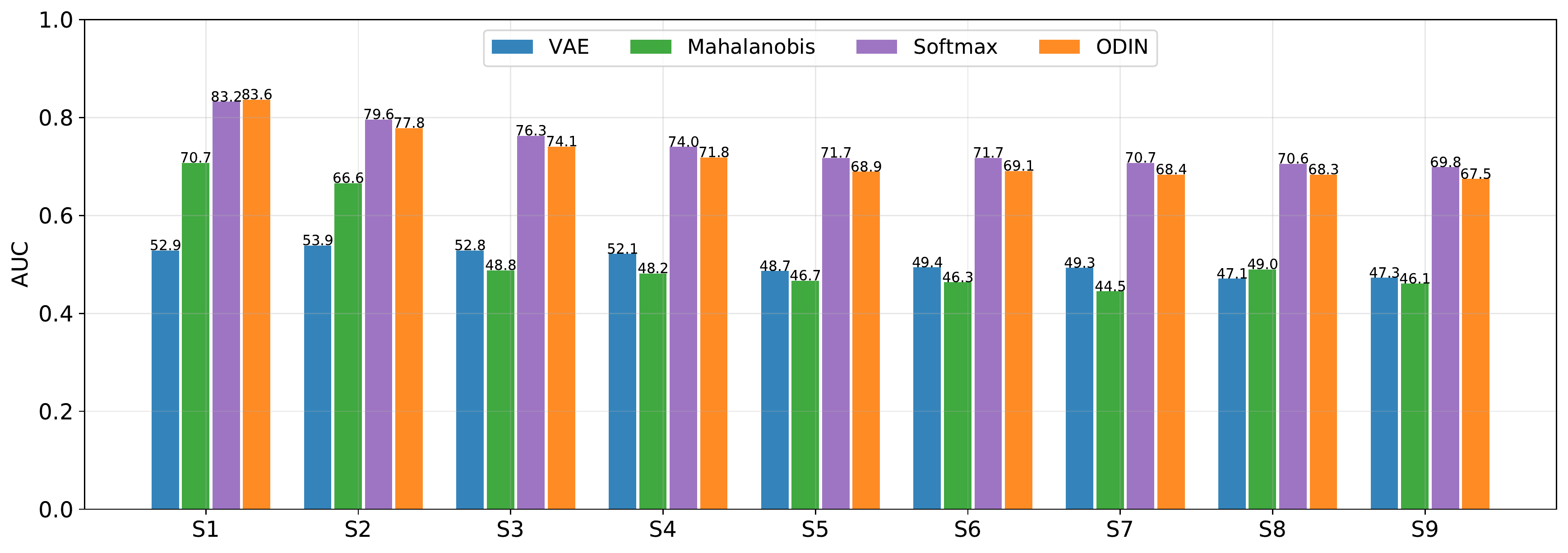}} }%
     \hfill
    \subfloat[{\footnotesize R.AUC~($\%$),~ \MAS}]{{\includegraphics[width=.33\textwidth]{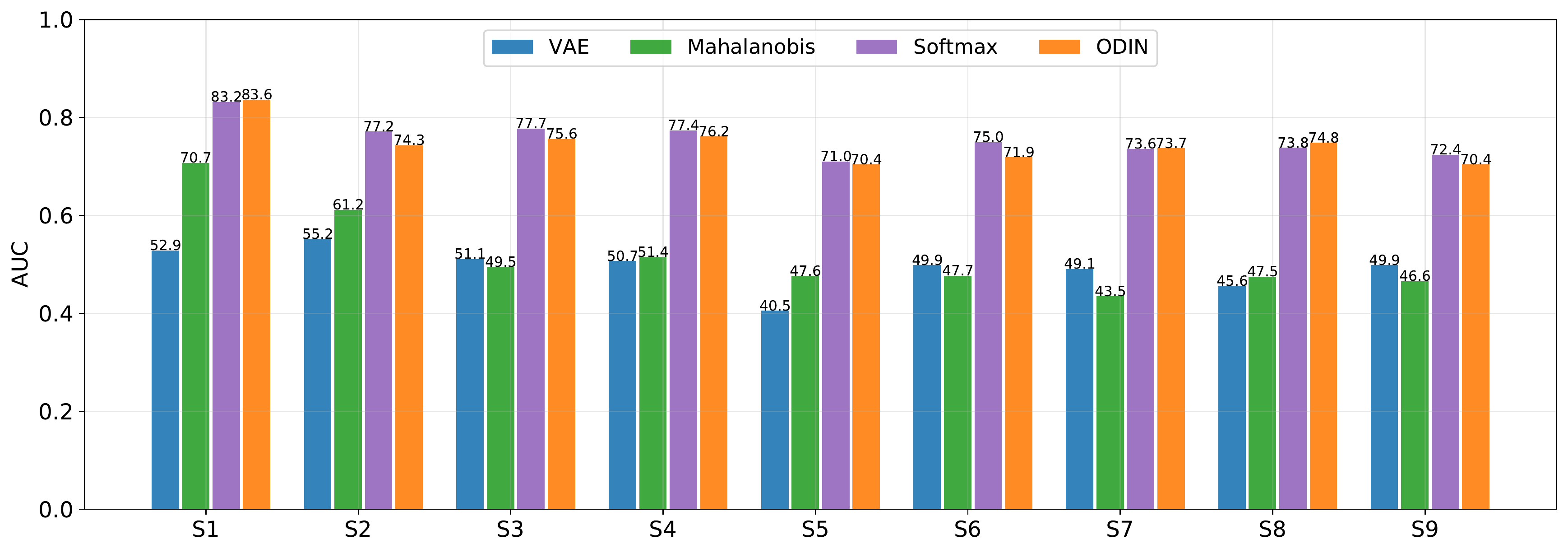}}}
        \hfill
    \subfloat[{\footnotesize P.AUC~($\%$),~ \MAS}]{{\includegraphics[width=.32\textwidth]{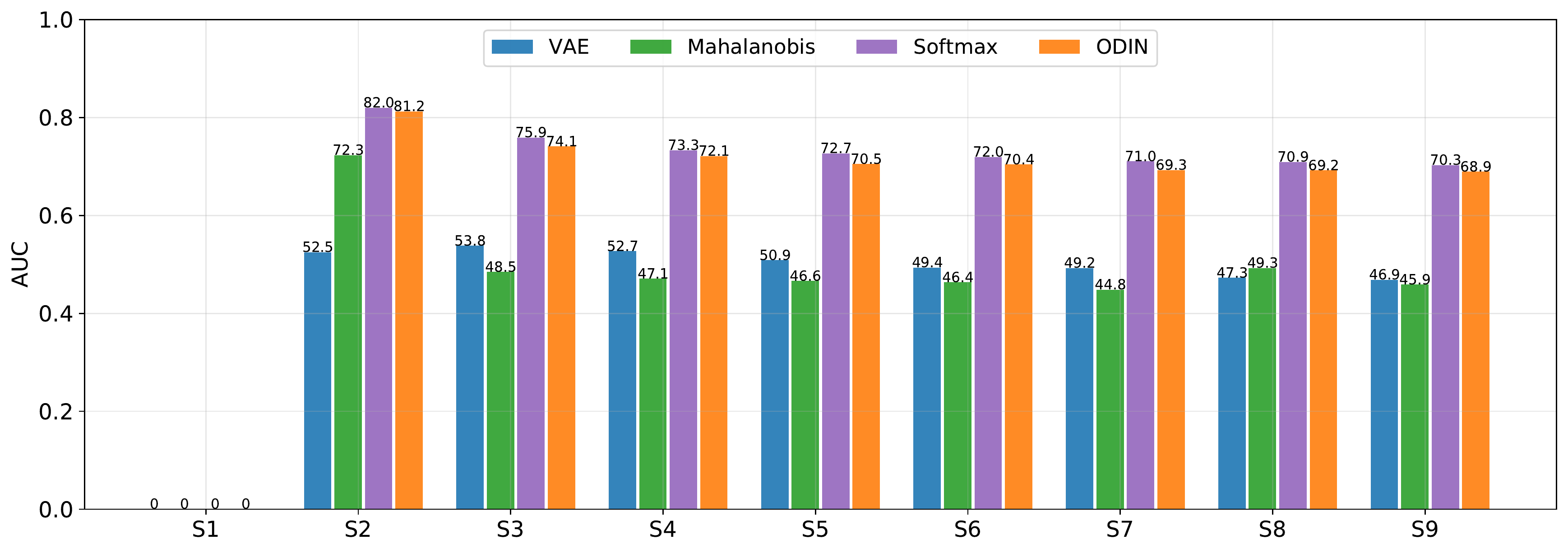}}}
\caption{\footnotesize Novelty detection  results in terms of Combined AUC~(C.AUC), Recent stage AUC~(R.AUC) and Previous stages AUC~(P.AUC) for  \tinyimg with \MAS and VGG network as a backbone.}%
    \label{fig:10_tasks_MAS}%
    \end{figure*}

\begin{figure*}[h]
  \vspace*{-0.2cm}
    \centering
    \subfloat[{\footnotesize C.AUC~($\%$),~ \LwF}]{{\includegraphics[width=.33\textwidth]{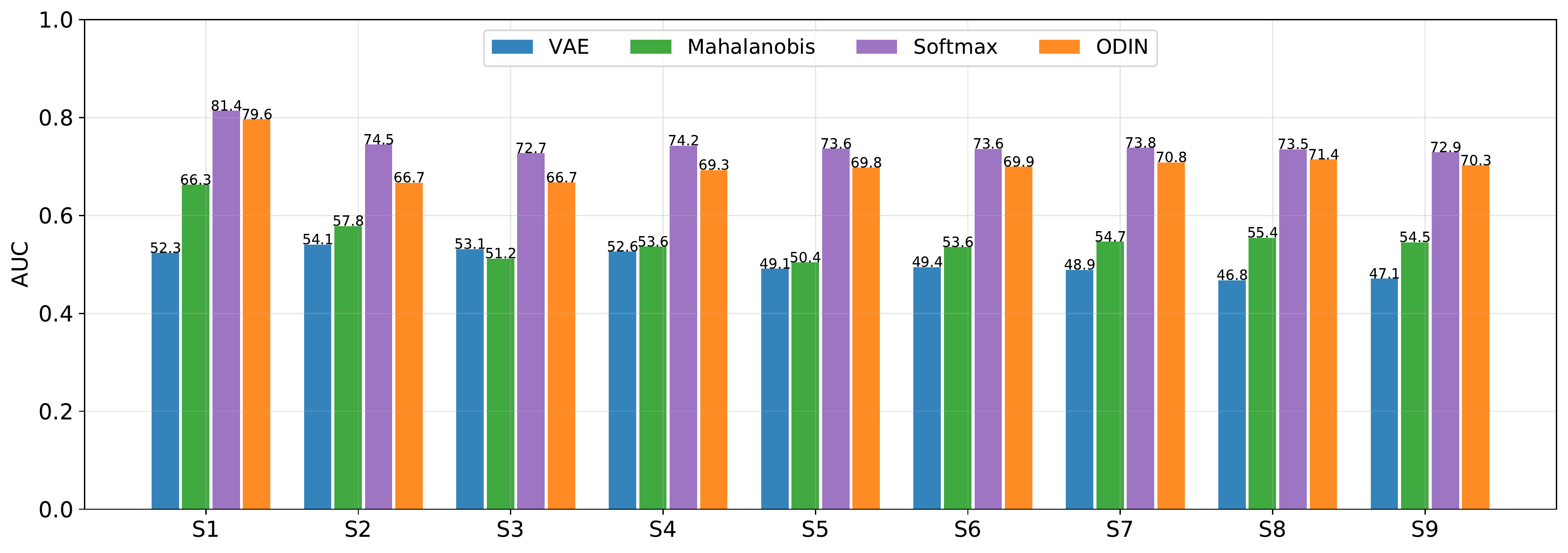}} }%
     \hfill
    \subfloat[{\footnotesize R.AUC~($\%$),~ \LwF}]{{\includegraphics[width=.33\textwidth]{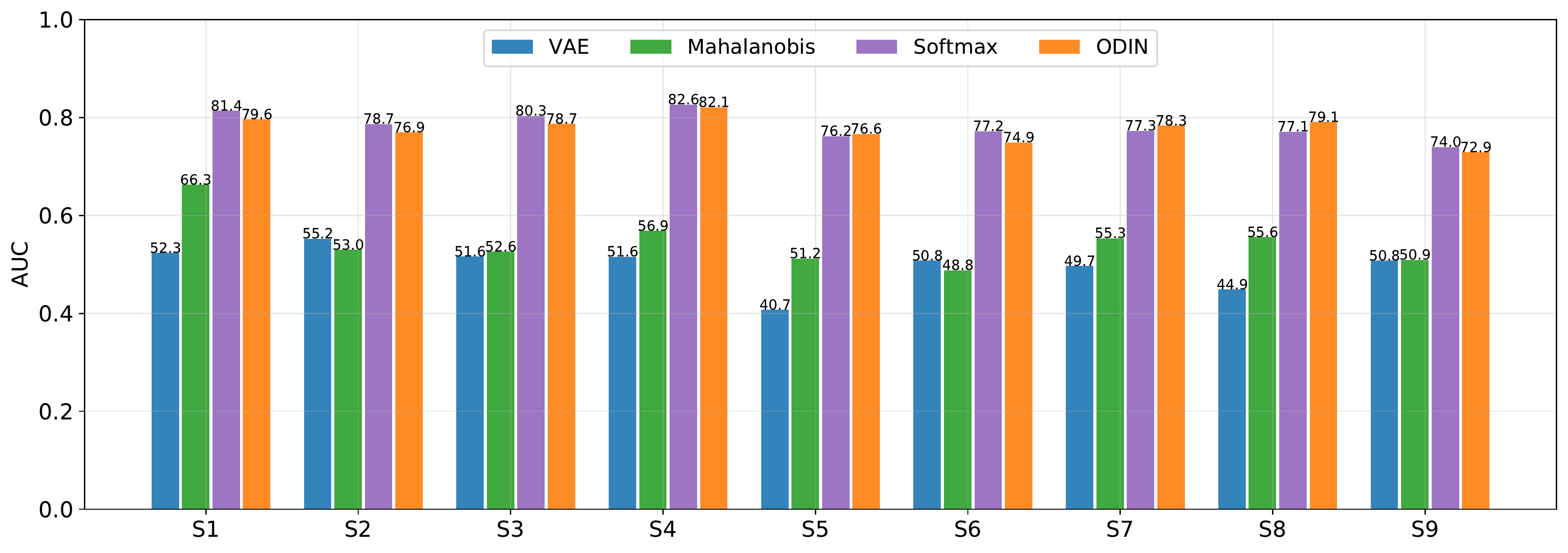}}}
        \hfill
    \subfloat[{\footnotesize P.AUC~($\%$),~ \LwF}]{{\includegraphics[width=.32\textwidth]{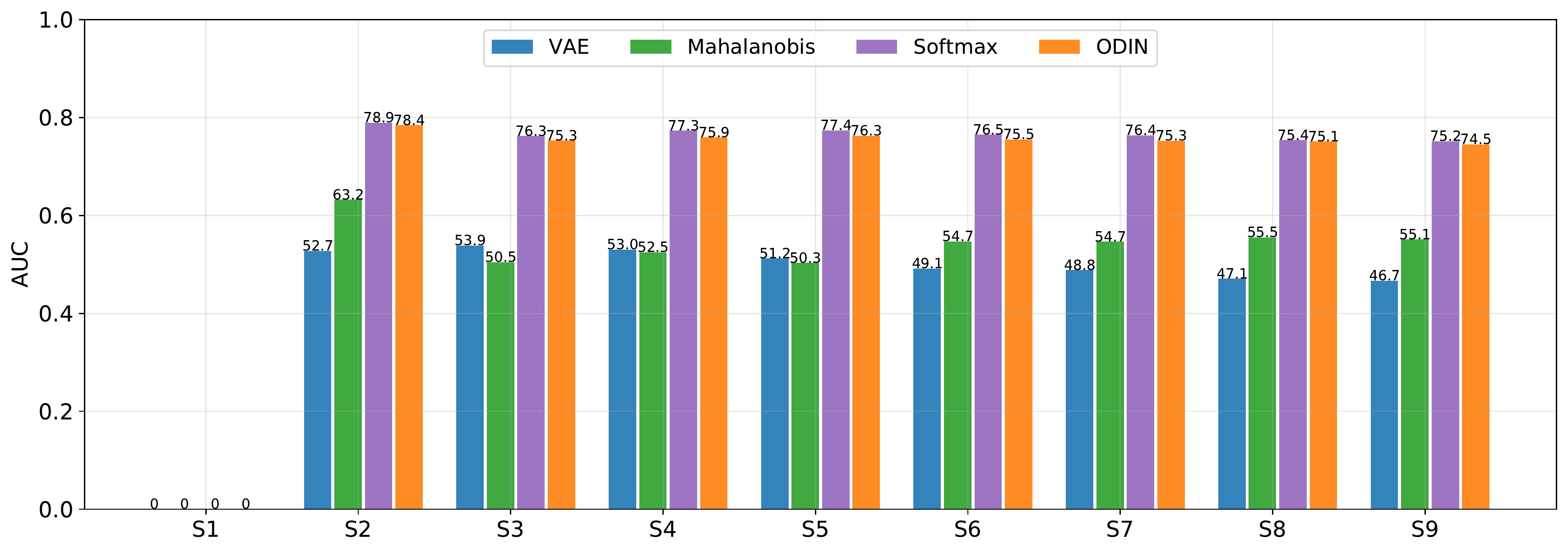}}}
\caption{\footnotesize Novelty detection  results  in terms of Combined AUC~(C.AUC), Recent stage AUC~(R.AUC) and Previous stages AUC~(P.AUC) for \tinyimg with \LwF and VGG network as a backbone.}%
    \label{fig:10_tasks_LwF}%
    \end{figure*}

    In the main paper, we have reported the ND performance for \tinyimg multi-head setting with only ResNet as a backbone model. Here, we report the ND performance when using VGG as a backbone model.
    Figures~\ref{fig:10_tasks_Fintune},~\ref{fig:10_tasks_MAS},~\ref{fig:10_tasks_LwF} report the Combined AUC (C.AUC), Recent Stage AUC (R.AUC) and Previous stages AUC (P.AUC) metrics for \fine, \MAS and \LwF  respectively. 
   It can be seen that the ND performance deteriorates as more learning stages are passing, this is more severe when no continual learning method is applied (\fine). \nav and \odin are the best methods, with \nav performing best when looking at the Combined AUC (C.AUC) metric. We note that our observations are consistent for both backbones, ResNet and VGG.

       
       \subsection{ Novelty detection  Performance with Other Metrics\label{sec:othermetrics}}
\begin{figure*}[t]
  \vspace*{-0.2cm}
    \centering
    \subfloat[{\footnotesize AUPR-IN~($\%$),~ \fine}]{{\includegraphics[width=.33\textwidth]{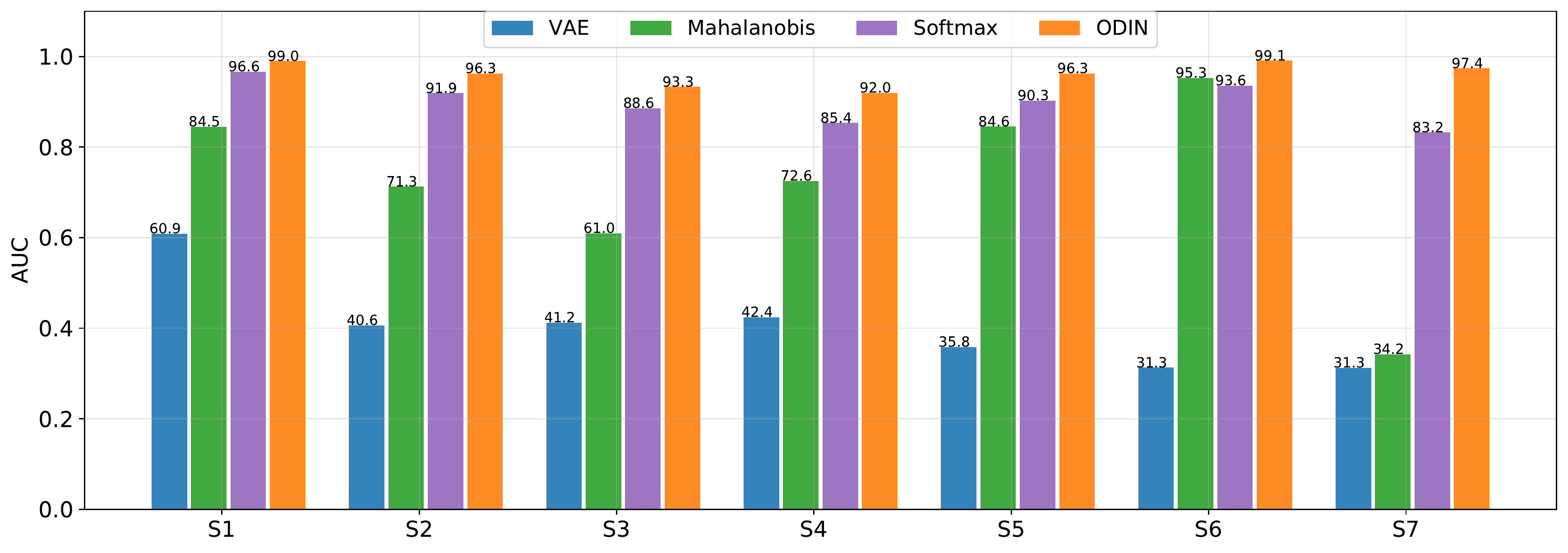}} }%
     \hfill
    \subfloat[{\footnotesize AUPR-IN~($\%$),~\LwF}]{{\includegraphics[width=.33\textwidth]{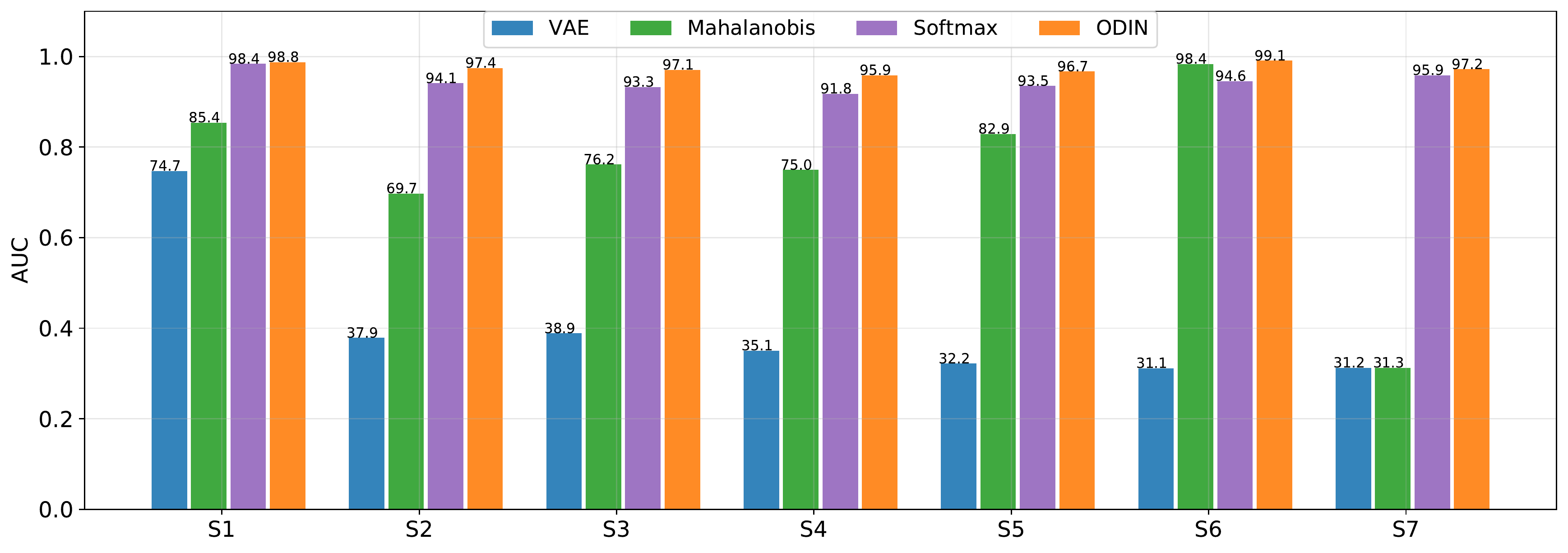}}}
        \hfill
    \subfloat[{\footnotesize AUPR-IN~($\%$),~ \MAS}]{{\includegraphics[width=.32\textwidth]{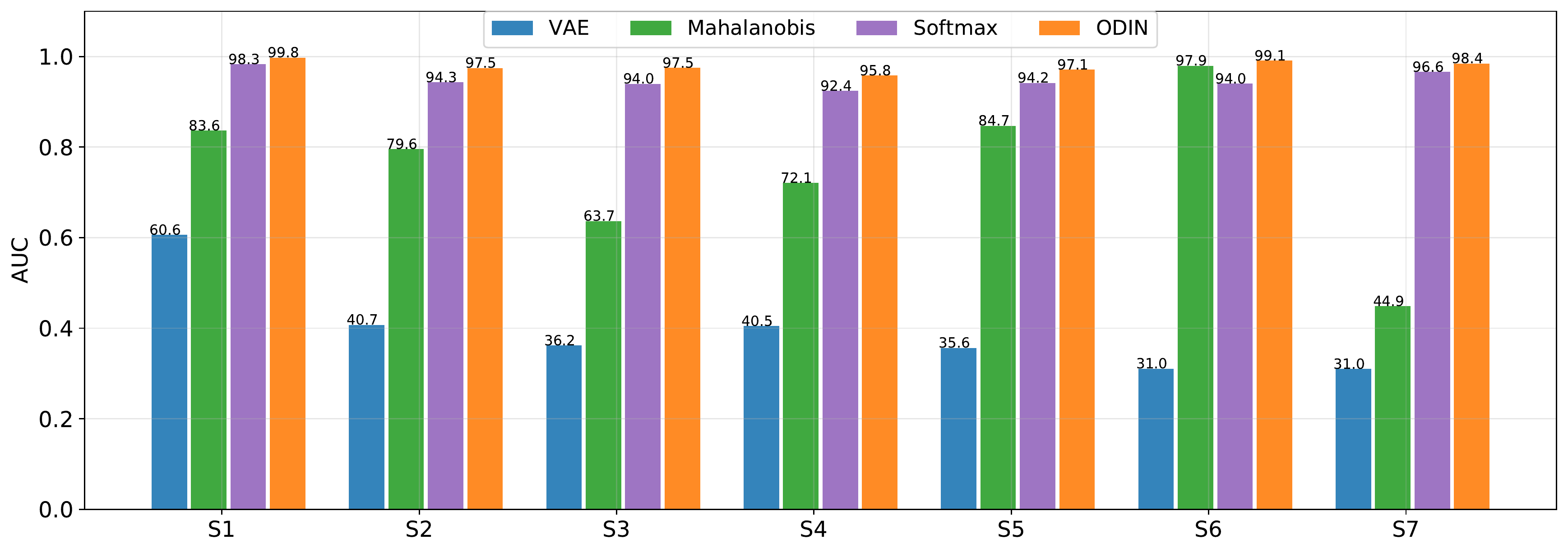}}}
\caption{\footnotesize Novelty detection performance measured by AUPR-IN and estimated after each learning stage and averaged over all possible (\IN,~\Out) pairs, for    \tasks.}%
    \label{fig:8task_auprin}%
    \end{figure*}

\begin{figure*}[h]
  \vspace*{-0.2cm}
    \centering
    \subfloat[{\footnotesize AUPR-IN~($\%$),~ \fine}]{{\includegraphics[width=.33\textwidth]{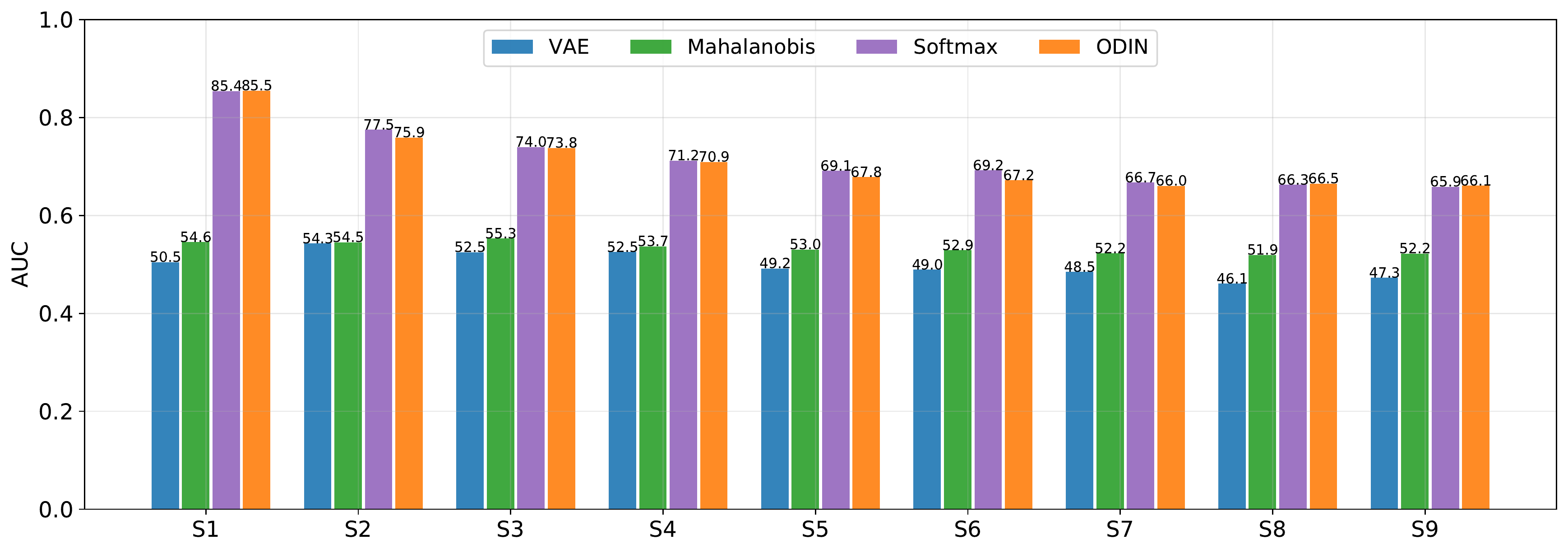}}} %
     \hfill
    \subfloat[{\footnotesize AUPR-IN~($\%$),~\LwF}]{{\includegraphics[width=.33\textwidth]{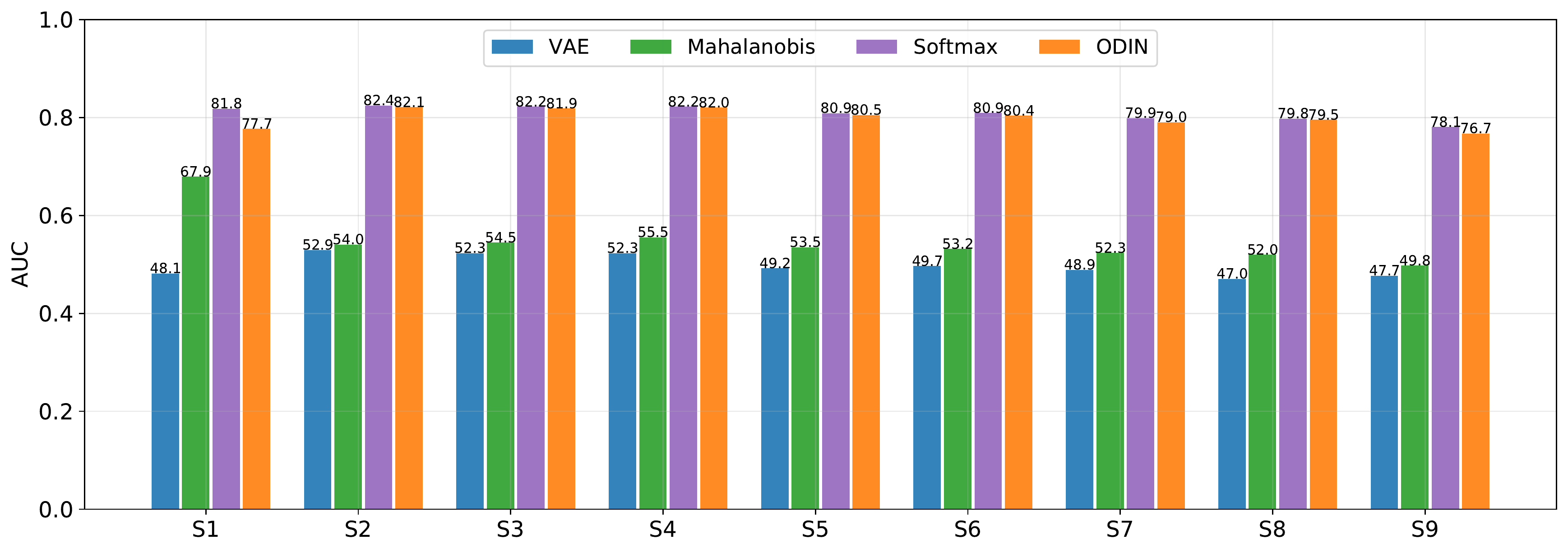}}}
        \hfill
    \subfloat[{\footnotesize AUPR-ININ~($\%$),~ \MAS}]{{\includegraphics[width=.32\textwidth]{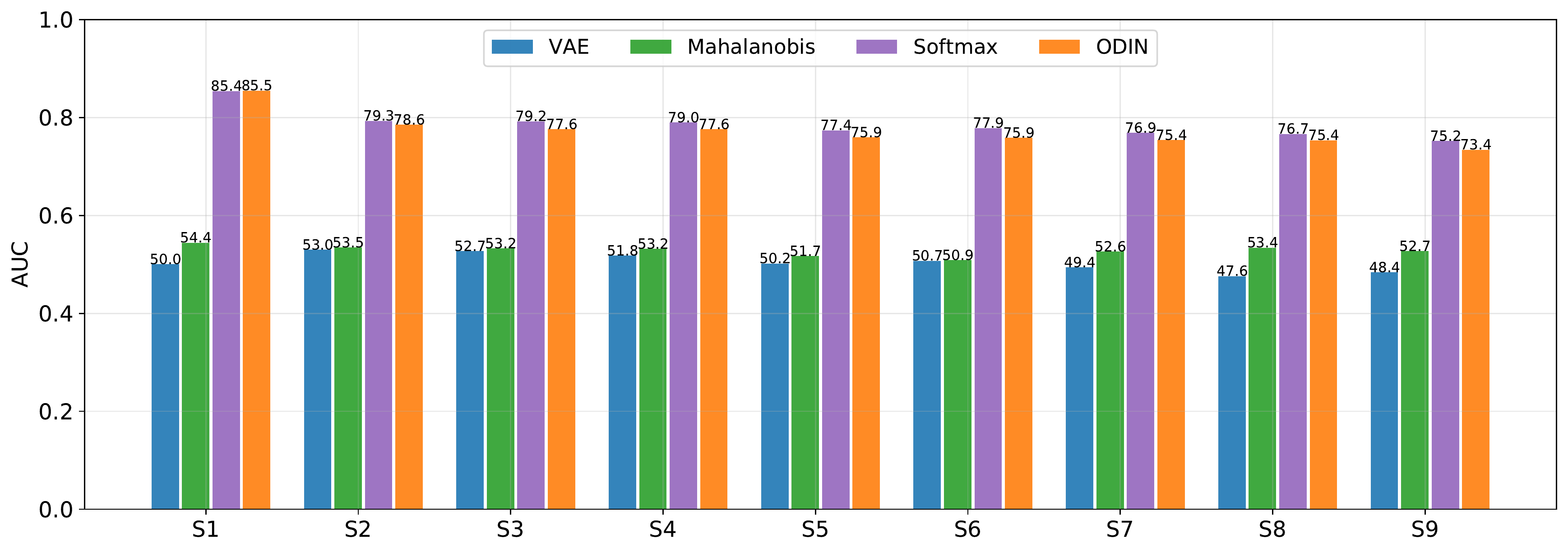}}}
\caption{\footnotesize Novelty detection performance measured by  AUPR-IN and estimated after each learning stage and averaged over all possible (\IN,~\Out) pairs, for multi-head \tinyimg with ResNet.}%
    \label{fig:tiny_multihead_resnet_auprin}%
    \end{figure*}

\begin{figure*}[h]
  \vspace*{-0.2cm}
    \centering
    \subfloat[{\footnotesize AUPR-IN~($\%$),~ \fine}]{{\includegraphics[width=.33\textwidth]{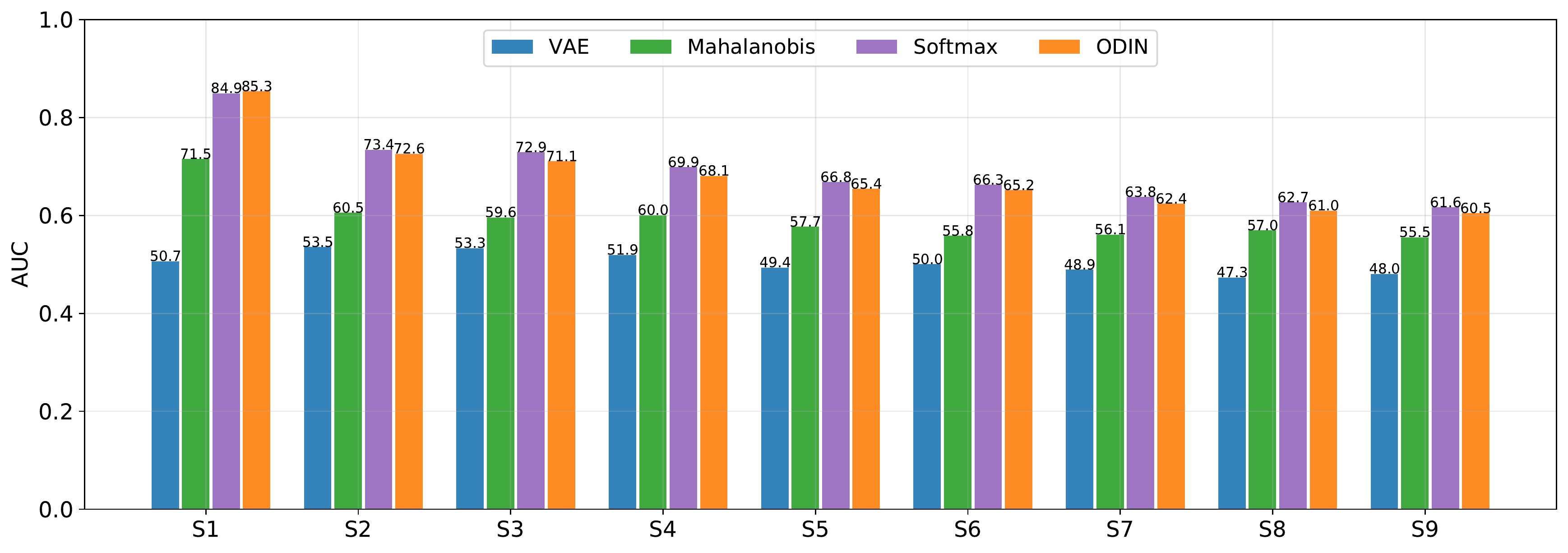}}} %
     \hfill
    \subfloat[{\footnotesize AUPR-IN~($\%$),~\LwF}]{{\includegraphics[width=.33\textwidth]{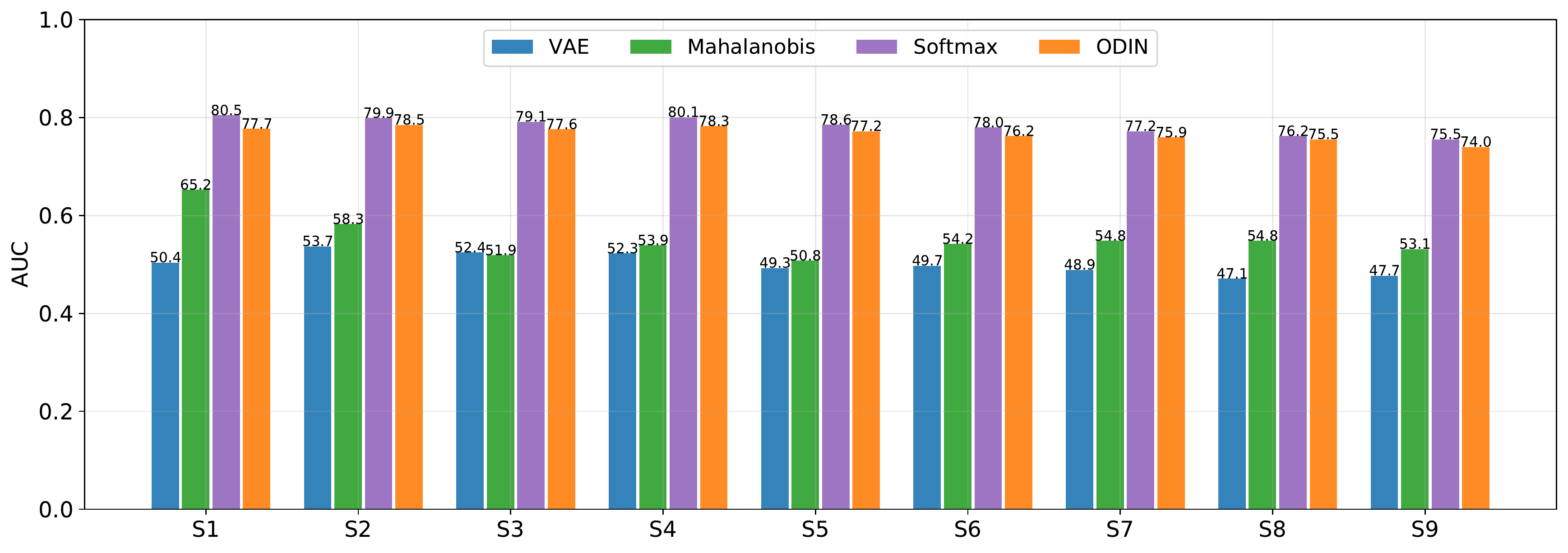}}}
        \hfill
    \subfloat[{\footnotesize AUPR-IN~($\%$),~ \MAS}]{{\includegraphics[width=.32\textwidth]{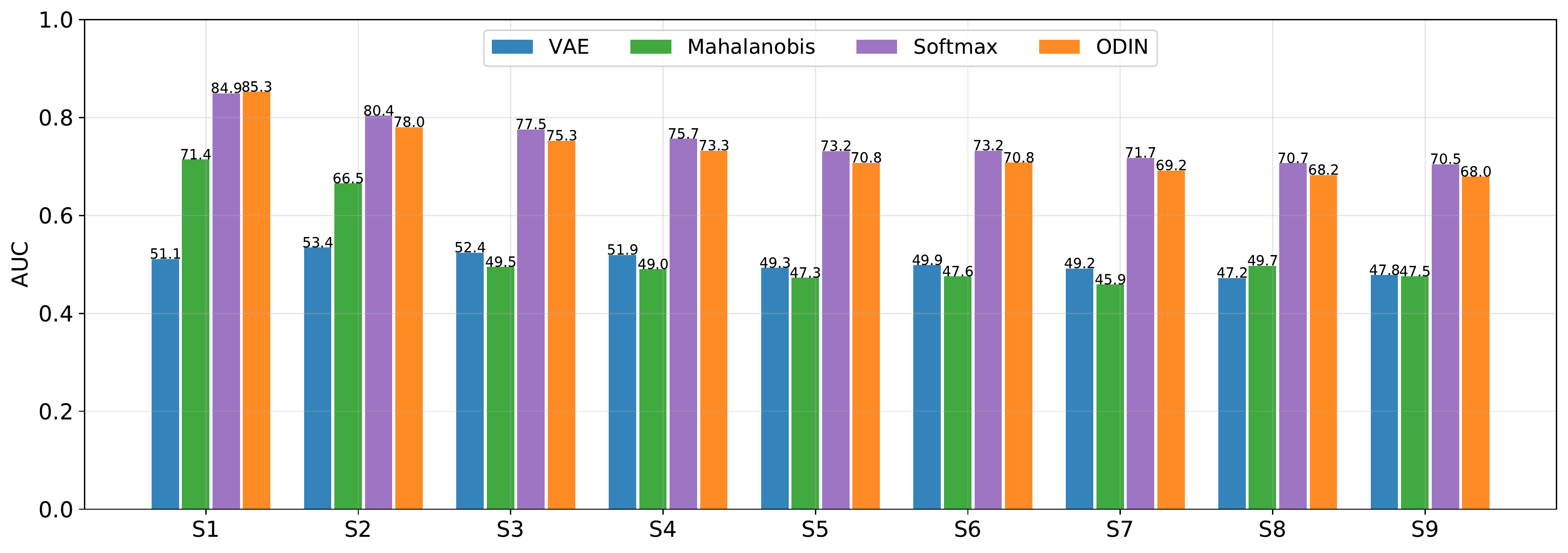}}}
\caption{\footnotesize Novelty detection performance measured by AUPR-IN and estimated after each learning stage and averaged over all possible (\IN,~\Out) pairs,  for   multi-head \tinyimg with VGG.}%
    \label{fig:tiny_multihead_vgg_auprin}%
    \end{figure*}

\begin{figure*}[h]
  \vspace*{-0.2cm}
    \centering
    \subfloat[{\footnotesize AUPR-IN~($\%$),~ ER 25s}]{{\includegraphics[width=.44\textwidth]{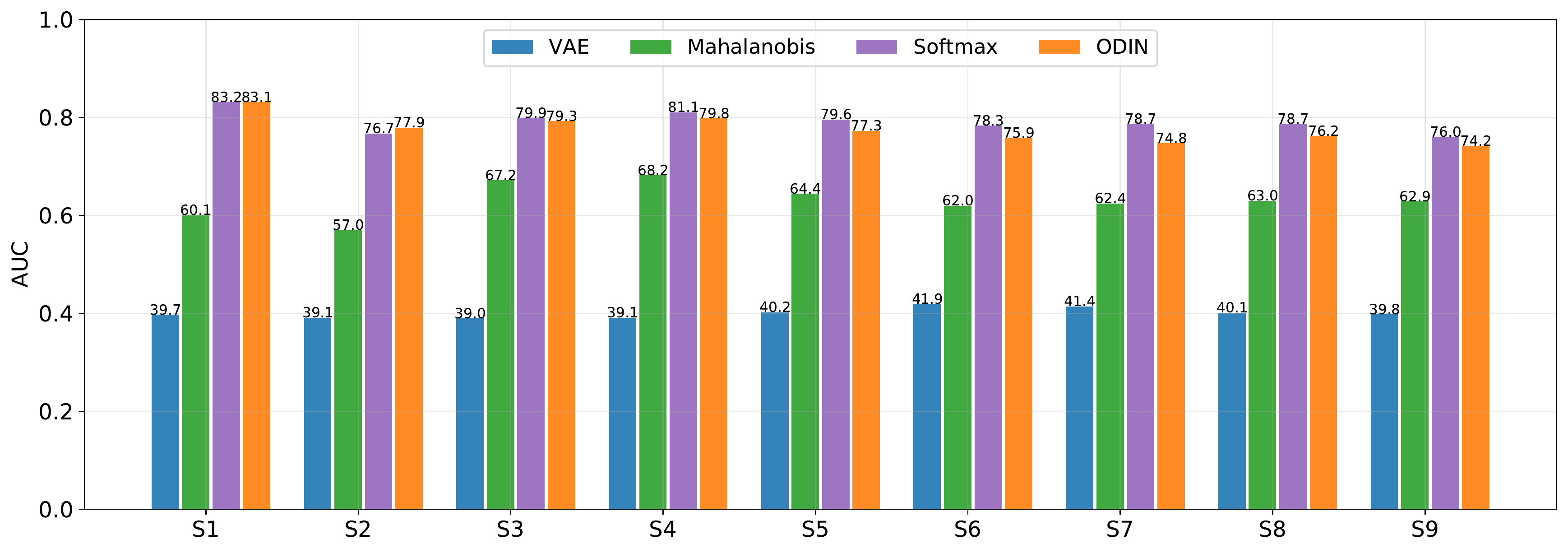}}}
        \hfill
    \subfloat[{\footnotesize AUPR-IN~($\%$),~ER 50s}]{{\includegraphics[width=.44\textwidth]{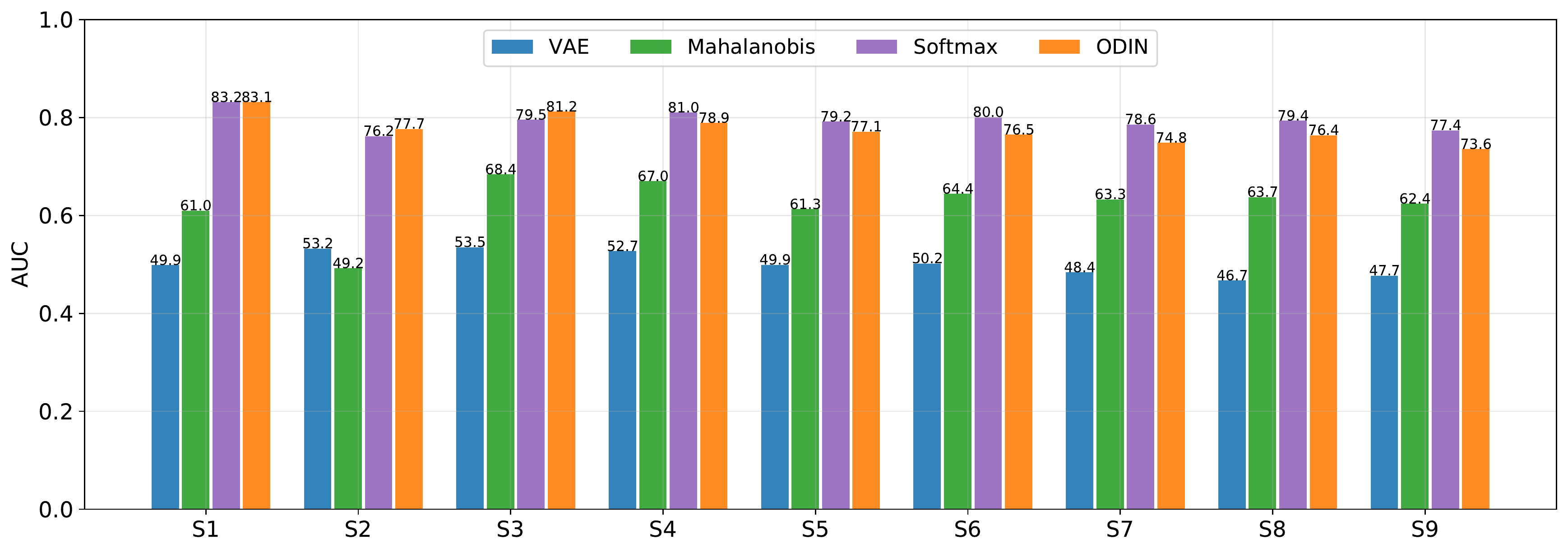}}}
\caption{\footnotesize Novelty detection performance measured by  AUPR-IN estimated after each learning stage and averaged over all possible (\IN,~\Out) pairs,  for    shared-head \tinyimg with ResNet and \ER.}%
    \label{fig:tiny_sharedhead_resnet_auprin}%
    \end{figure*}

\begin{figure*}[h]
  \vspace*{-0.2cm}
    \centering
    \subfloat[{\footnotesize AUPR-IN~($\%$),~ SSIL 25s}]{{\includegraphics[width=.44\textwidth]{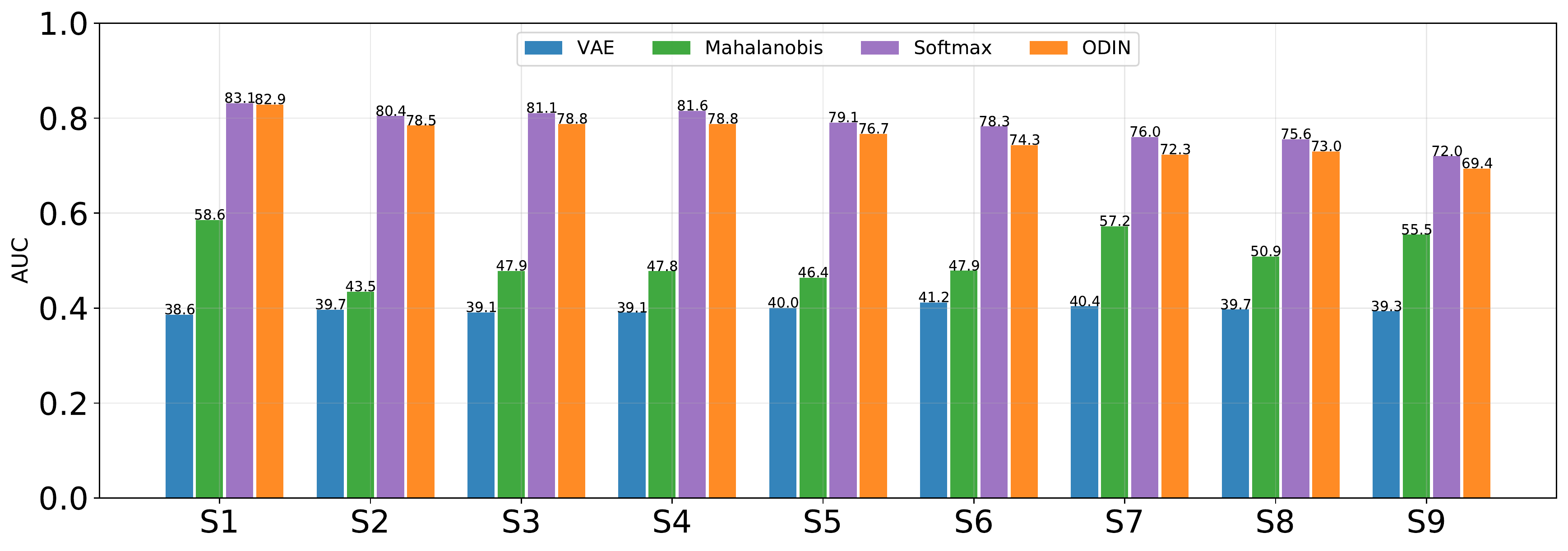}}}
        \hfill
    \subfloat[{\footnotesize AUPR-IN~($\%$),~SSIL 50s}]{{\includegraphics[width=.44\textwidth]{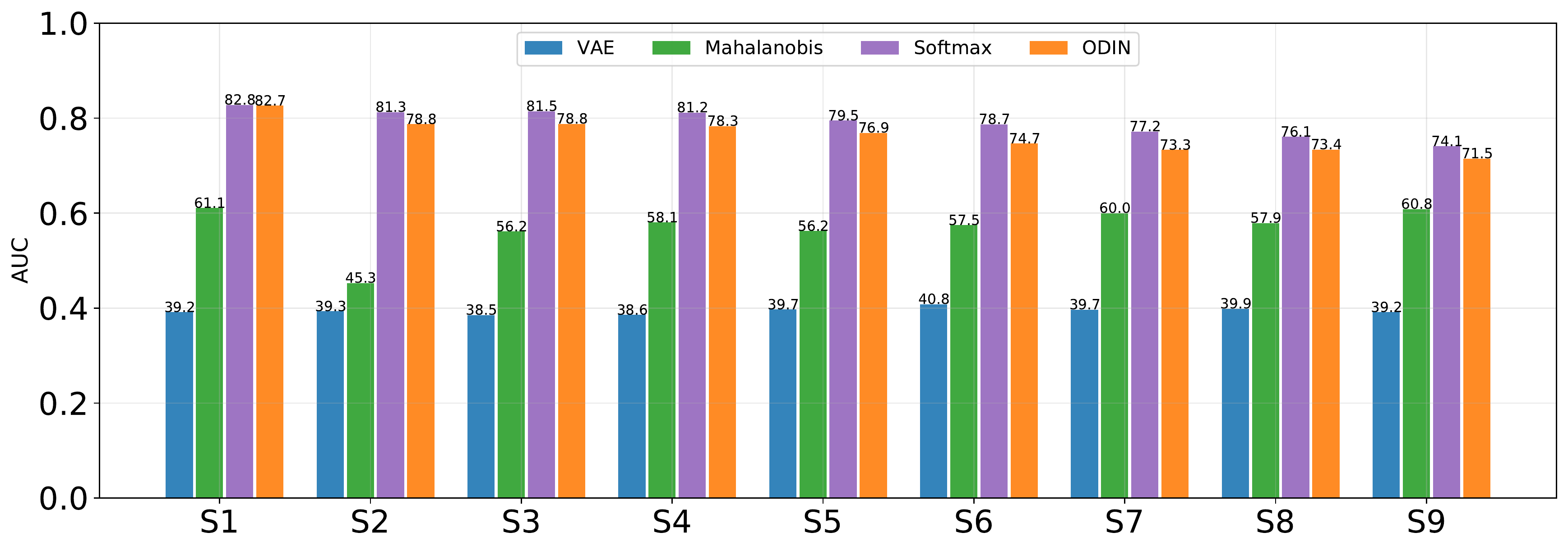}}}
\caption{\footnotesize Novelty detection performance measured by  AUPR-IN and estimated after each learning stage and averaged over all possible (\IN,~\Out) pairs,  for   shared-head \tinyimg with ResNet and \SSIL.}%
    \label{fig:tiny_sharedhead_resnet_auprin_SSIL}%
    \end{figure*}

\begin{figure*}[t]
  \vspace*{-0.2cm}
    \centering
    \subfloat[{\footnotesize DER~($\%$),~ \fine}]{{\includegraphics[width=.33\textwidth]{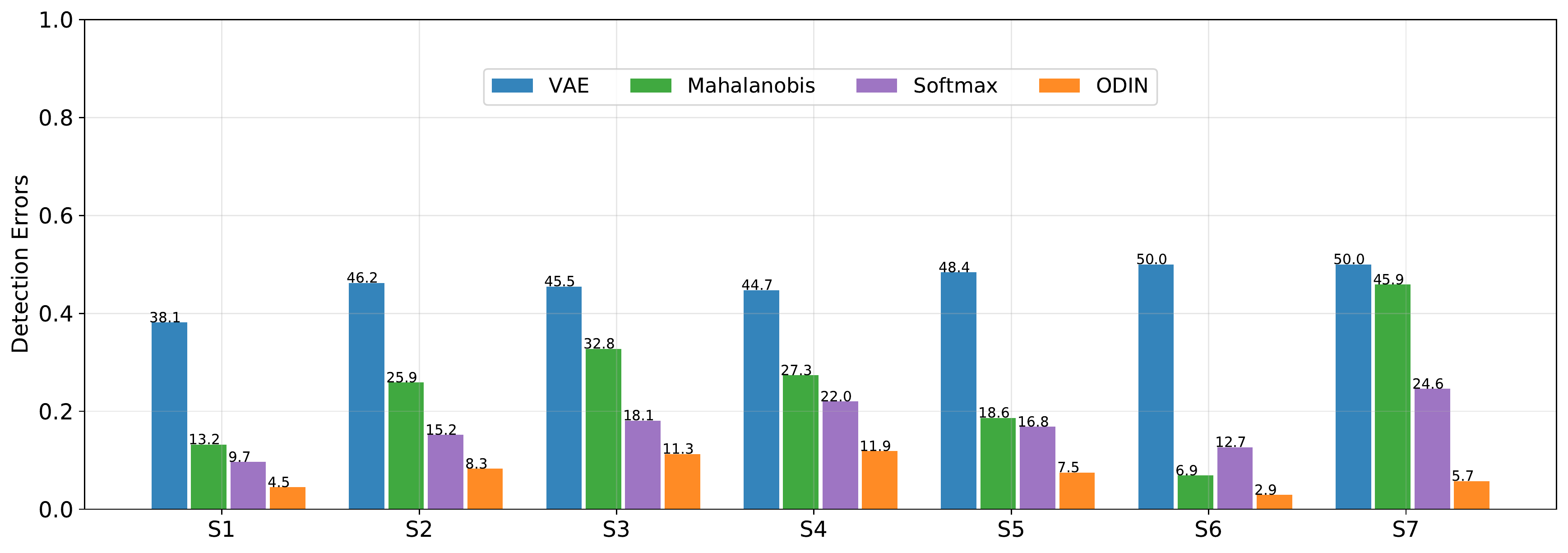}} }%
     \hfill
    \subfloat[{\footnotesize DER~($\%$),~\LwF}]{{\includegraphics[width=.33\textwidth]{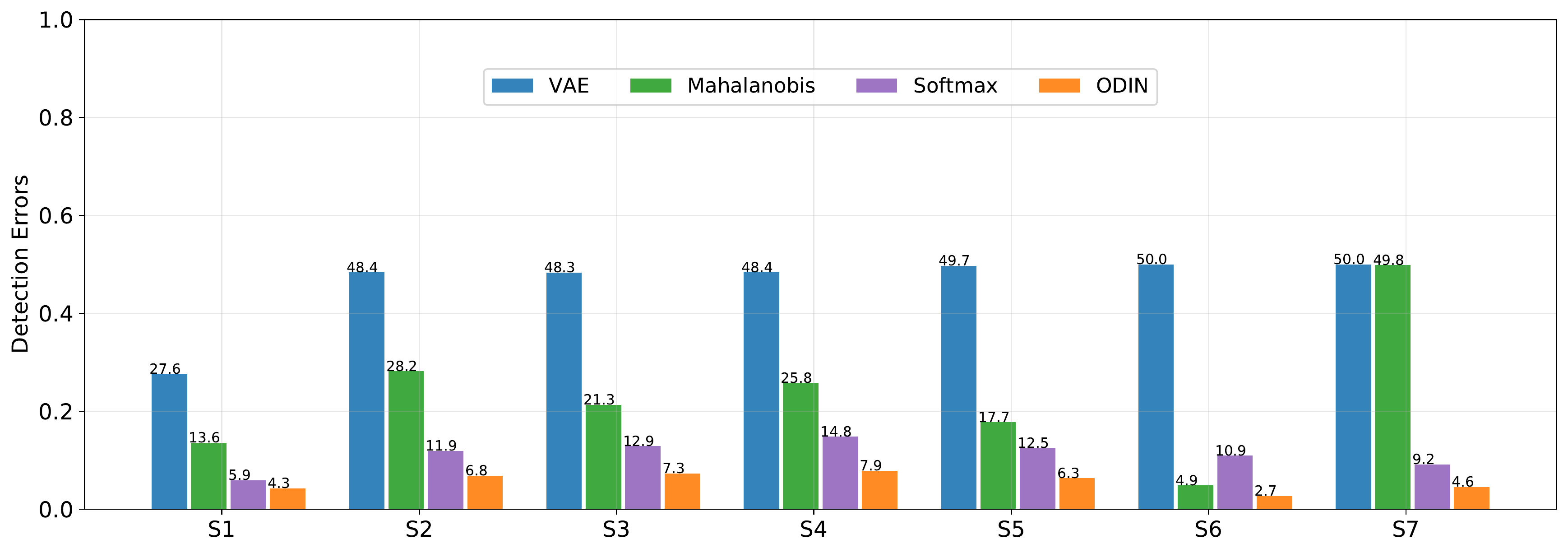}}}
        \hfill
    \subfloat[{\footnotesize DER~($\%$),~ \MAS}]{{\includegraphics[width=.32\textwidth]{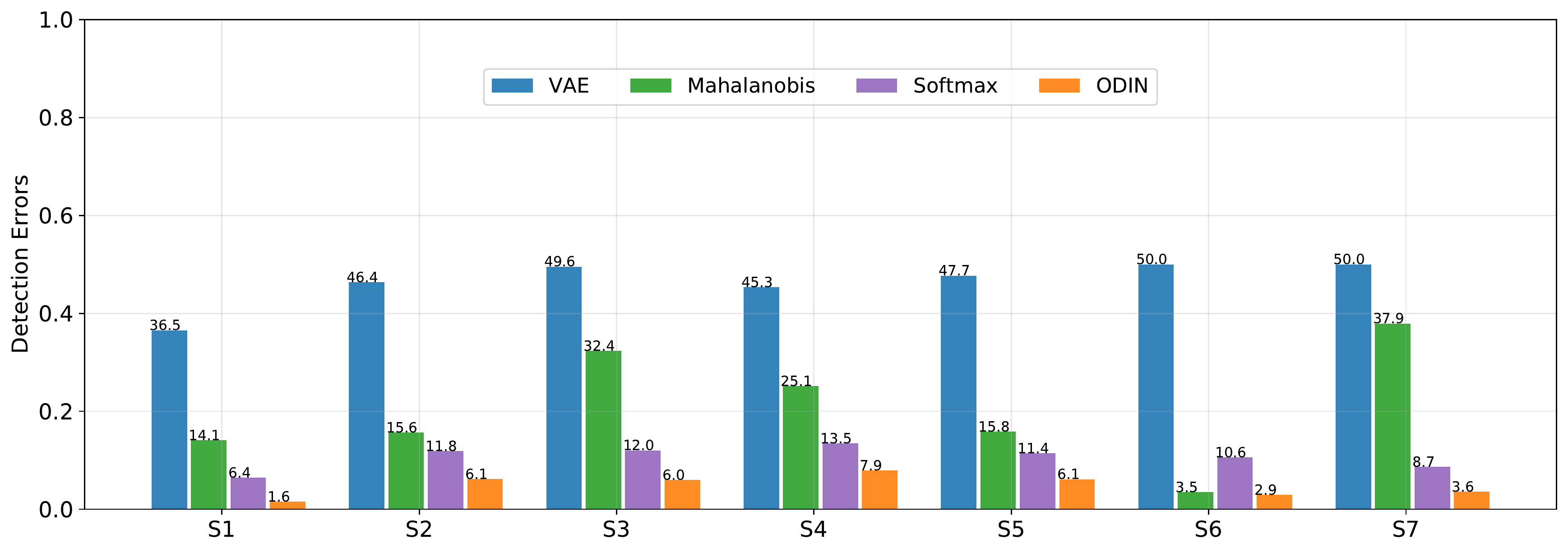}}}
\caption{\footnotesize Novelty detection performance measured by DER (lower is better)  after each learning stage averaged over each possible (\IN,~\Out) pair, for    \tasks.}%
    \label{fig:8task_DER}%
    \end{figure*}

\begin{figure*}[h]
  \vspace*{-0.2cm}
    \centering
    \subfloat[{\footnotesize DER~($\%$),~ \fine}]{{\includegraphics[width=.33\textwidth]{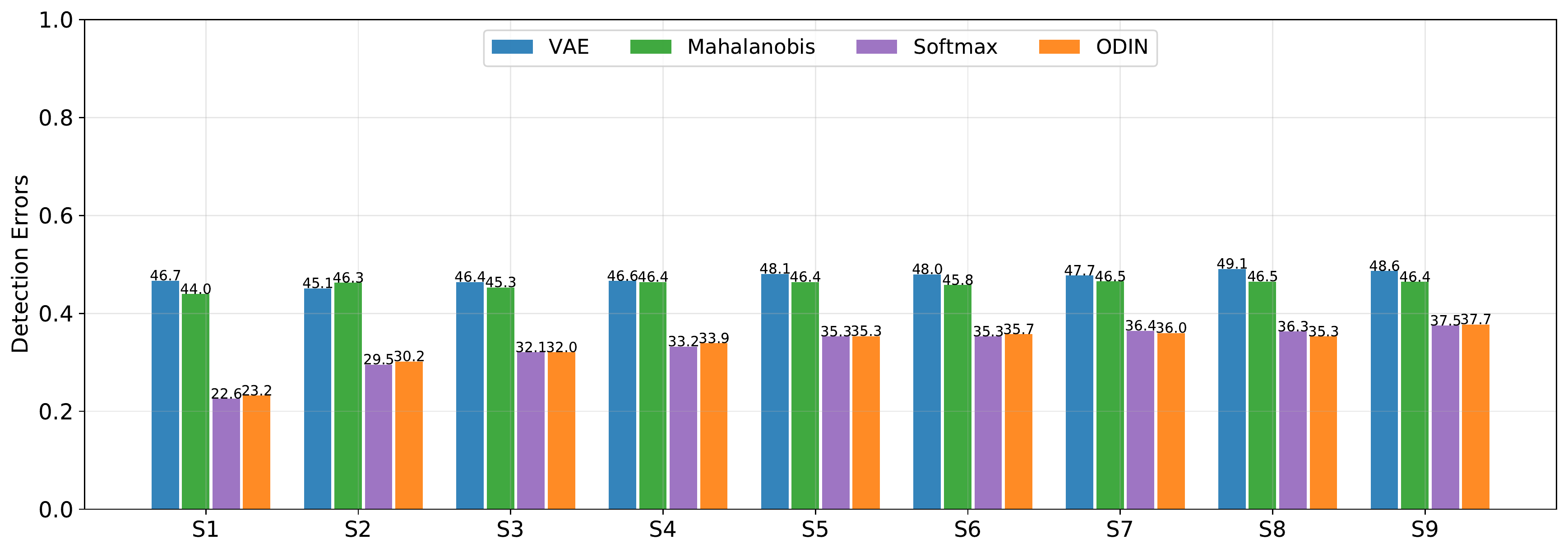}}} %
     \hfill
    \subfloat[{\footnotesize DER~($\%$),~\LwF}]{{\includegraphics[width=.33\textwidth]{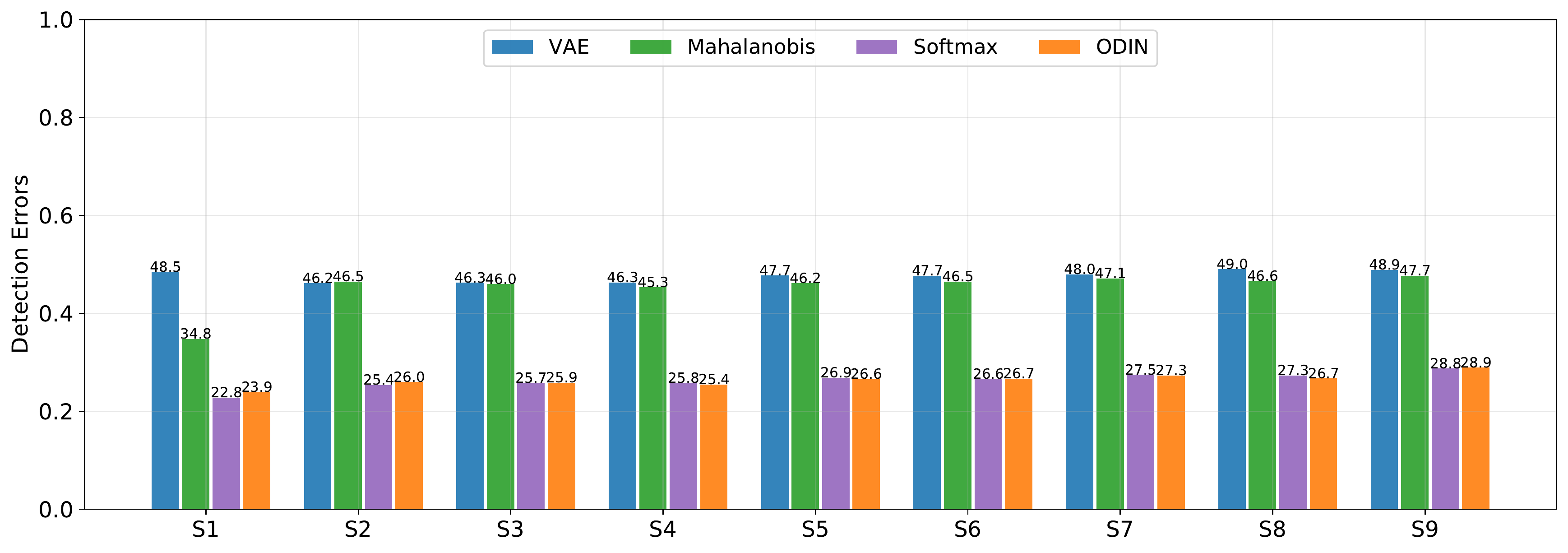}}}
        \hfill
    \subfloat[{\footnotesize DER~($\%$),~ \MAS}]{{\includegraphics[width=.32\textwidth]{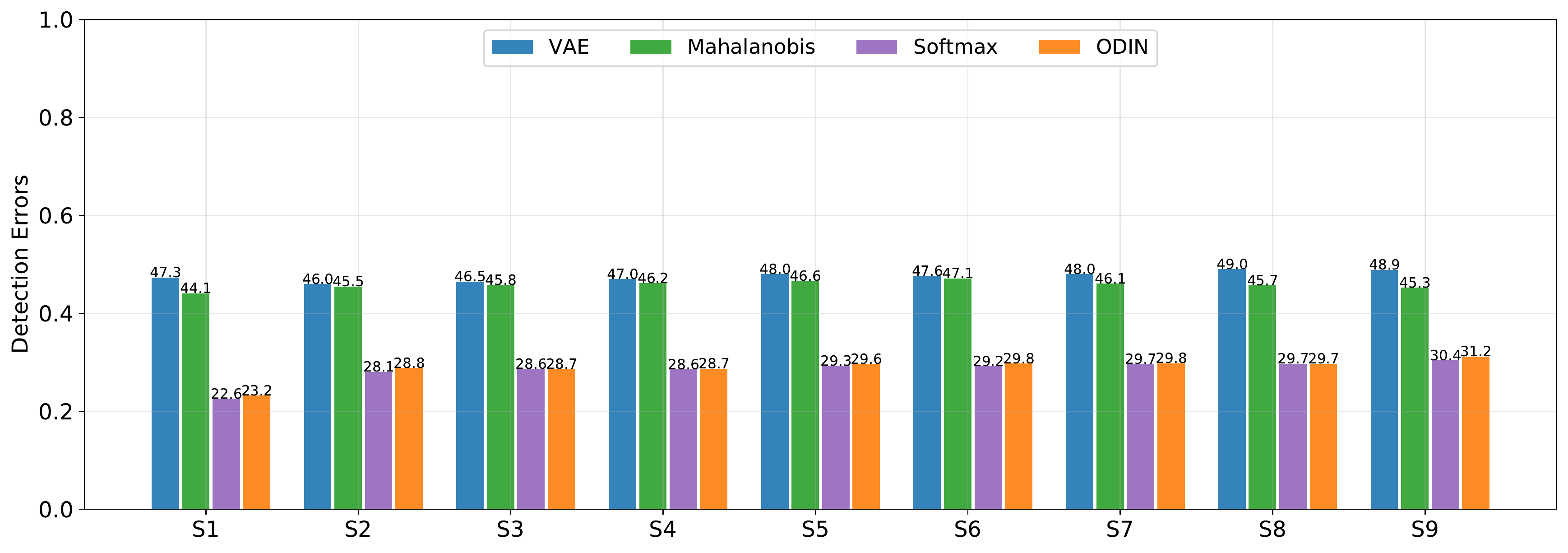}}}
\caption{\footnotesize Novelty detection performance measured by  DER (lower is better)  after each learning stage averaged over each possible (\IN,~\Out) pair, for   multi-head \tinyimg with ResNet.}%
    \label{fig:tiny_multihead_resnet_DER}%
    \end{figure*}

\begin{figure*}[h]
  \vspace*{-0.2cm}
    \centering
    \subfloat[{\footnotesize DER~($\%$),~ \fine}]{{\includegraphics[width=.33\textwidth]{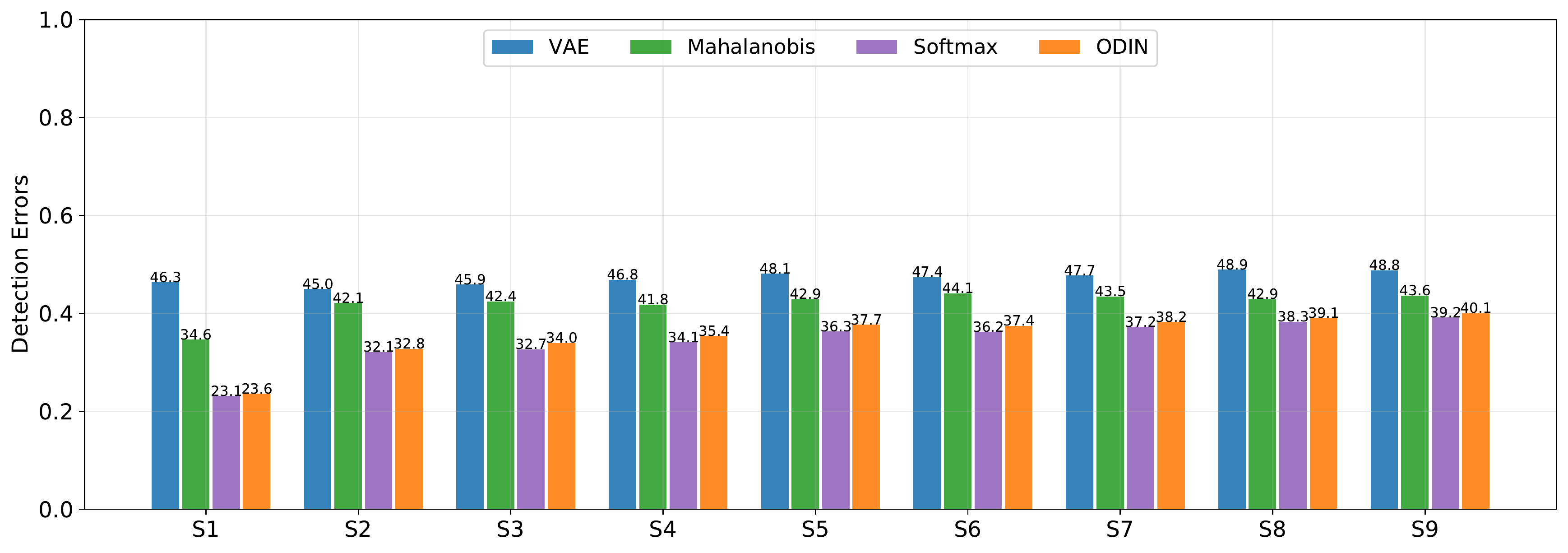}}} %
     \hfill
    \subfloat[{\footnotesize DER~($\%$),~\LwF}]{{\includegraphics[width=.33\textwidth]{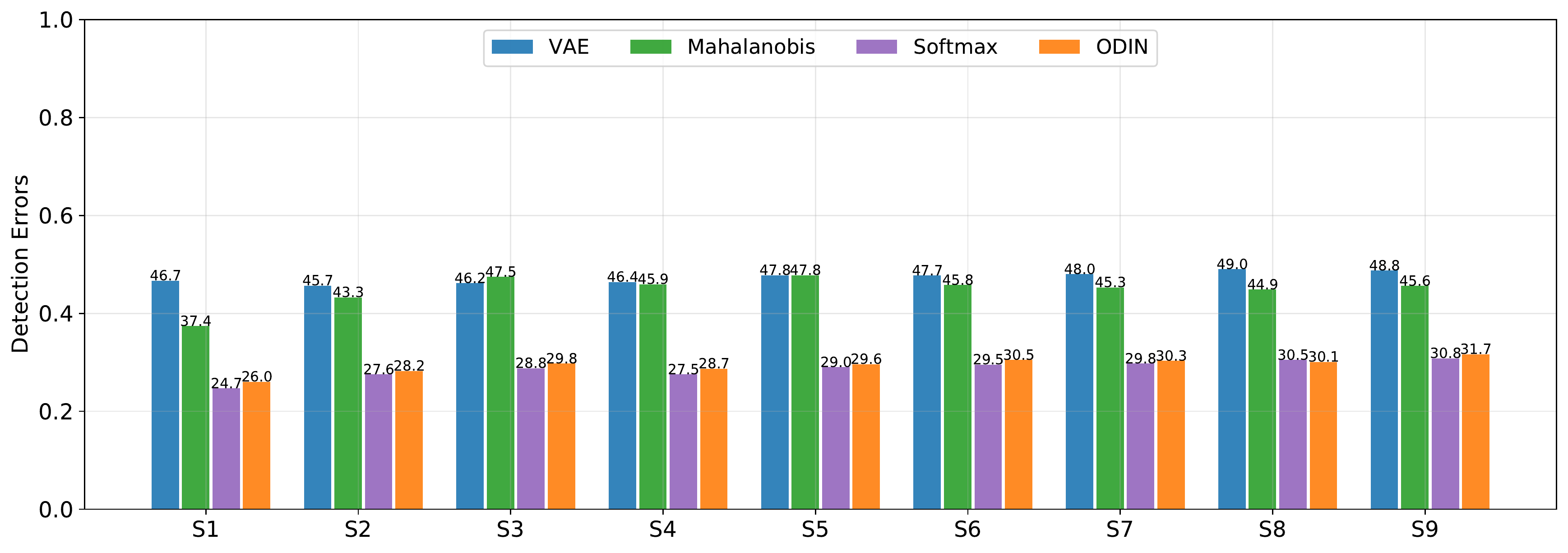}}}
        \hfill
    \subfloat[{\footnotesize DER~($\%$),~ \MAS}]{{\includegraphics[width=.32\textwidth]{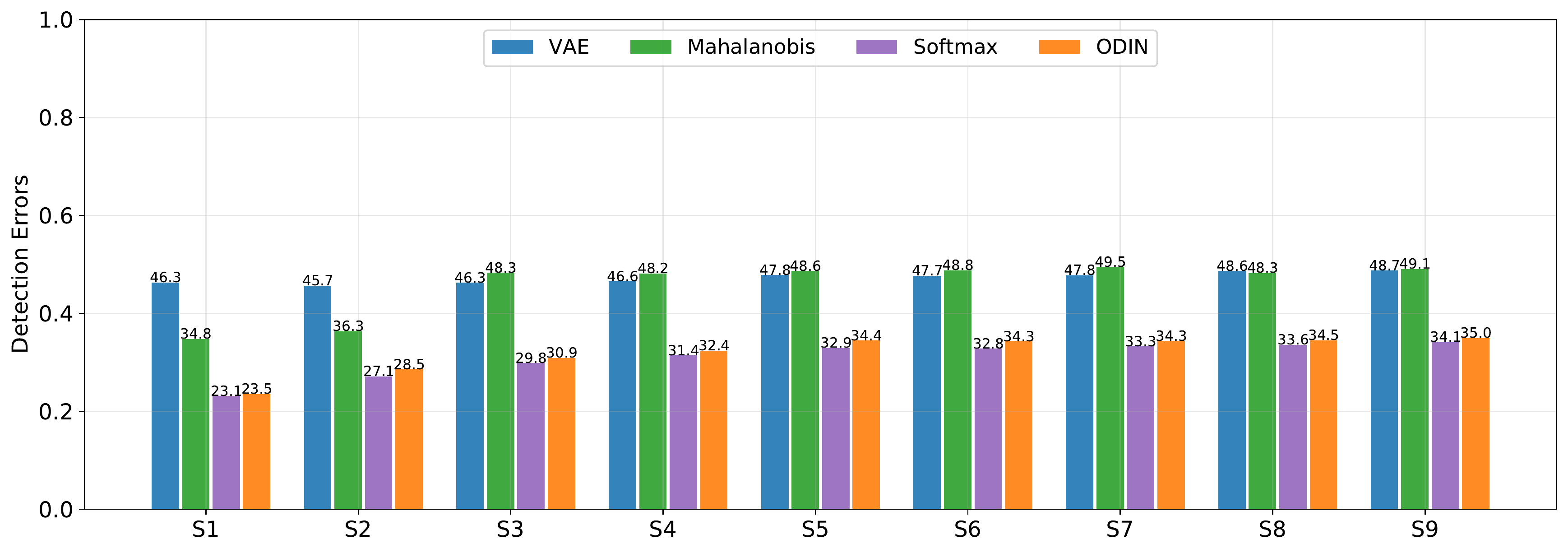}}}
\caption{\footnotesize Novelty detection performance measured by DER (lower is better)  after each learning stage averaged over each possible (\IN,~\Out) pair, for   multi-head \tinyimg with VGG.}%
    \label{fig:tiny_multihead_vgg_DER}%
    \end{figure*}

\begin{figure*}[h]
  \vspace*{-0.2cm}
    \centering
    \subfloat[{\footnotesize DER~($\%$),~ ER 25s}]{{\includegraphics[width=.44\textwidth]{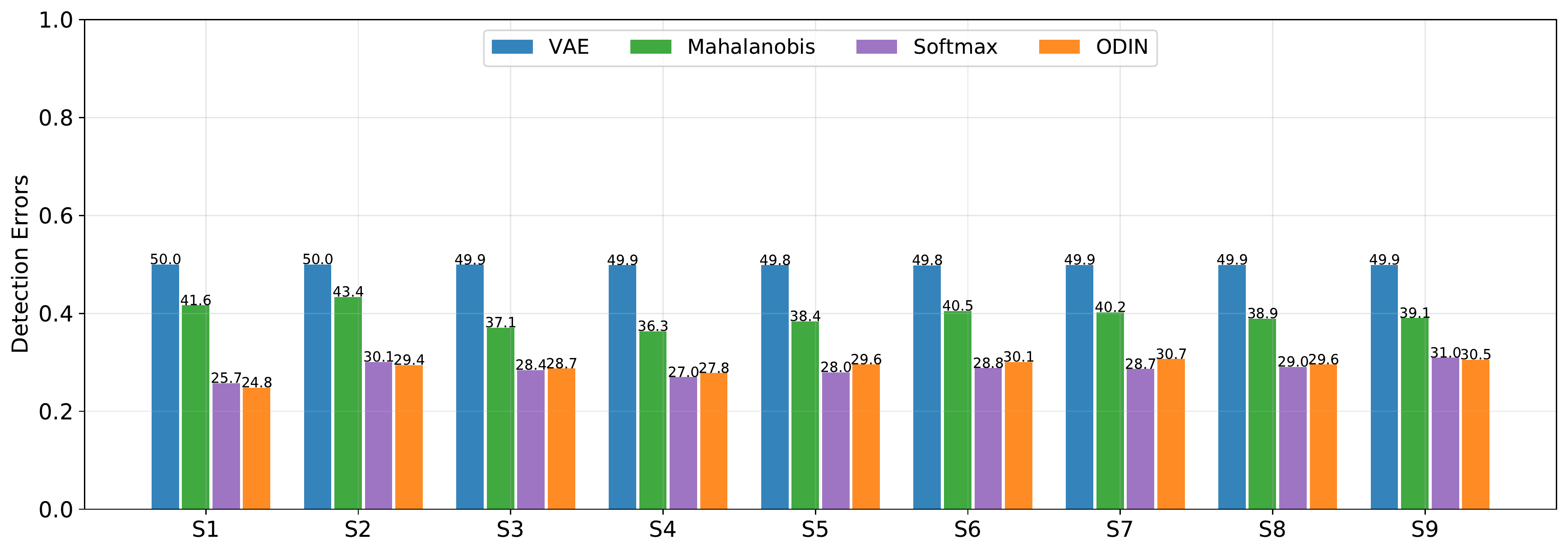}}}
        \hfill
    \subfloat[{\footnotesize DER~($\%$),~ER 50s}]{{\includegraphics[width=.44\textwidth]{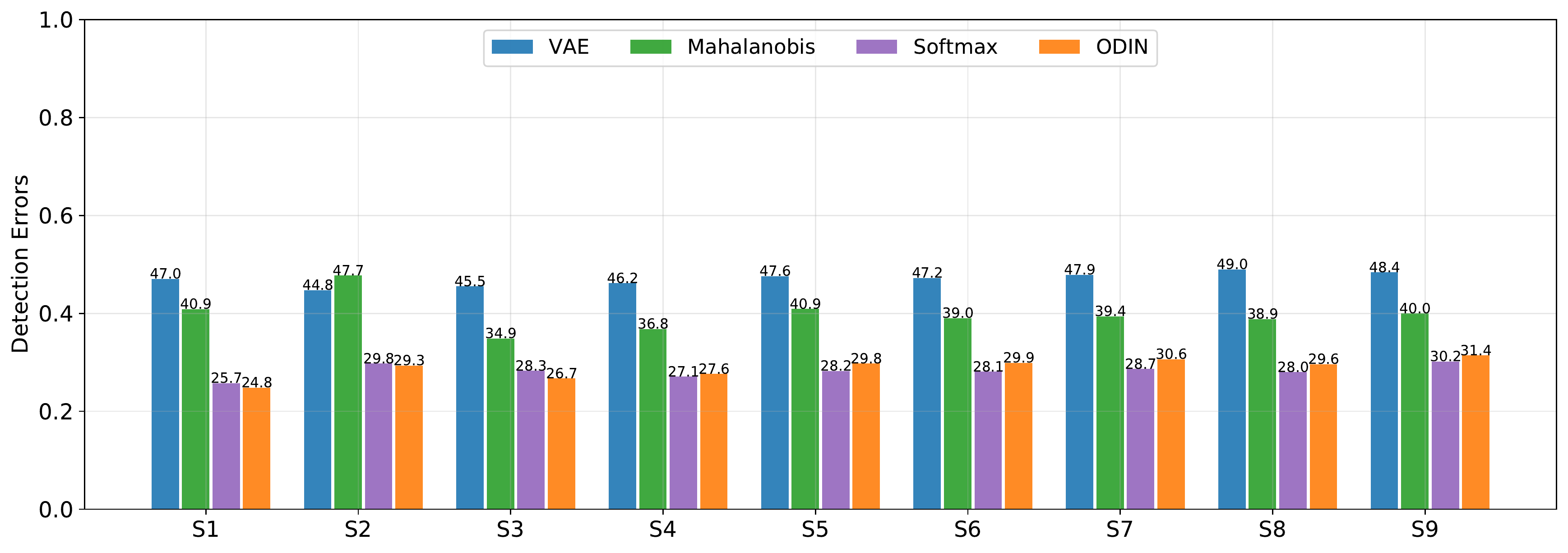}}}
\caption{\footnotesize Novelty detection performance measured by DER performance after each learning stage averaged over each possible (\IN,~\Out) pair,  for   shared-head \tinyimg with ResNet and \ER.}%
    \label{fig:tiny_sharedhead_resnet_DER}%
    \end{figure*}

\begin{figure*}[h]
  \vspace*{-0.2cm}
    \centering
    \subfloat[{\footnotesize DER~($\%$),~ SSIL 25s}]{{\includegraphics[width=.44\textwidth]{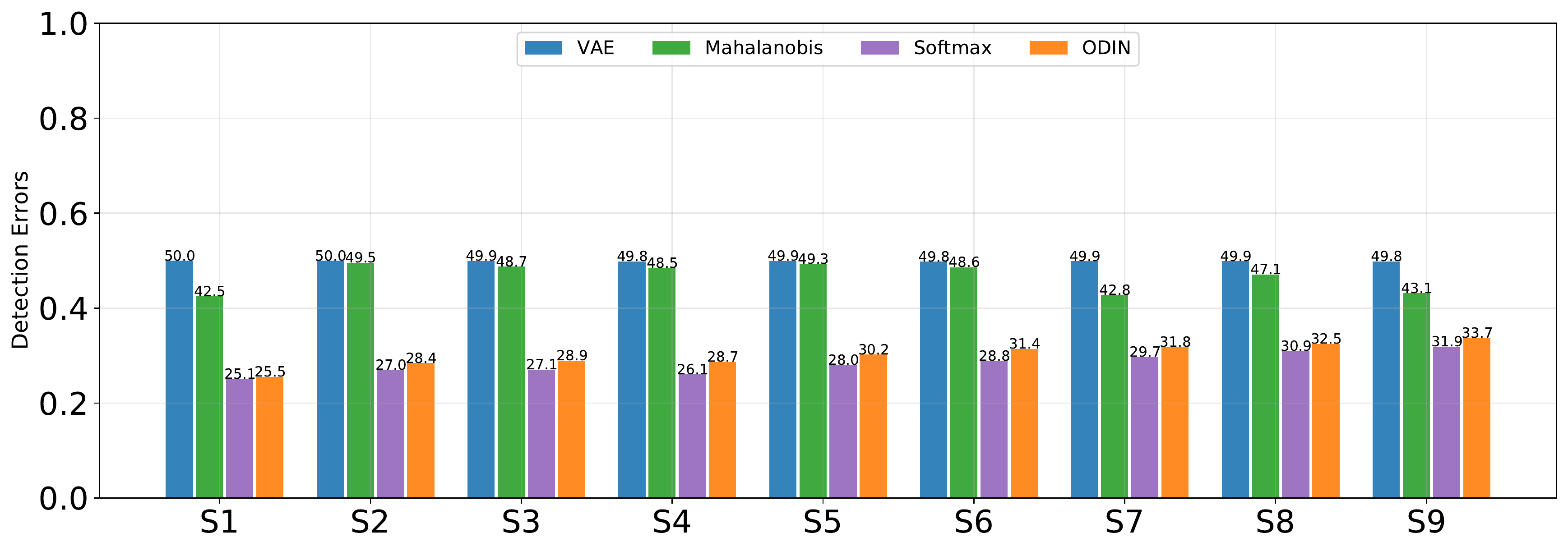}}}
        \hfill
    \subfloat[{\footnotesize DER~($\%$),~SSIL 50s}]{{\includegraphics[width=.44\textwidth]{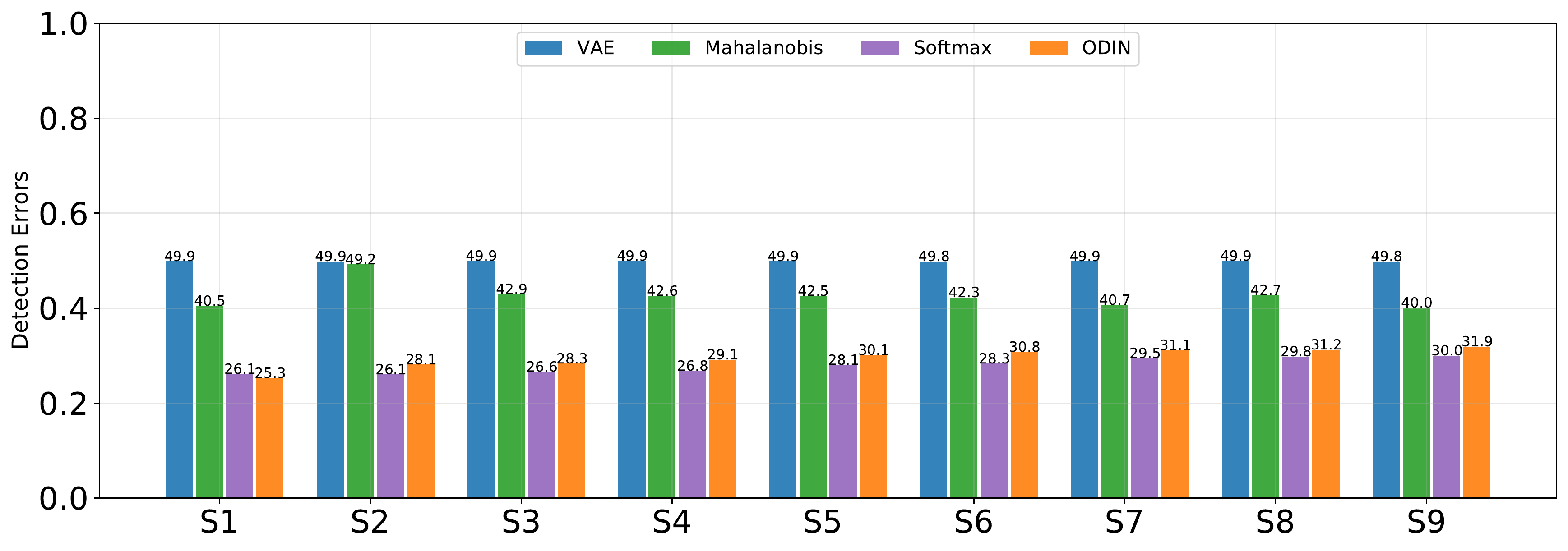}}}
\caption{\footnotesize Novelty detection performance measured by DER performance after each learning stage averaged over each possible (\IN,~\Out) pair,  for   shared-head \tinyimg with ResNet and \SSIL.}%
    \label{fig:tiny_sharedhead_resnet_DER_SSIL}%
    \end{figure*}

 Here, we report the average  DER and AUPR-IN scores of  the \IN vs. \Out  detection problems at each learning stage. 
   Figures~\ref{fig:8task_auprin},~\ref{fig:tiny_multihead_resnet_auprin},~\ref{fig:tiny_multihead_vgg_auprin},~\ref{fig:tiny_sharedhead_resnet_auprin},~\ref{fig:tiny_sharedhead_resnet_auprin_SSIL} show the AUPR-IN scores for \tasks, \tinyimg multi-head with ResNet, \tinyimg multi-head with VGG, \tinyimg shared-head with ResNet and \ER, shared-head with ResNet and \SSIL~ respectively.
      Figures~\ref{fig:8task_DER},~\ref{fig:tiny_multihead_resnet_DER},~\ref{fig:tiny_multihead_vgg_DER},~\ref{fig:tiny_sharedhead_resnet_DER},~\ref{fig:tiny_sharedhead_resnet_DER_SSIL} show the DER scores for \tasks, \tinyimg multi-head with ResNet, \tinyimg multi-head with VGG, \tinyimg shared-head with ResNet and \ER,~\tinyimg shared-head with ResNet and \SSIL~ respectively.
    \subsection{Inspection of the ND Performance on Each \IN Set\label{sec:laststage}}
\begin{figure*}[t]
  \vspace*{-0.2cm}
    \centering
    \subfloat[{\footnotesize Last Stage AUC~($\%$),~ \fine}]{{\includegraphics[width=.33\textwidth]{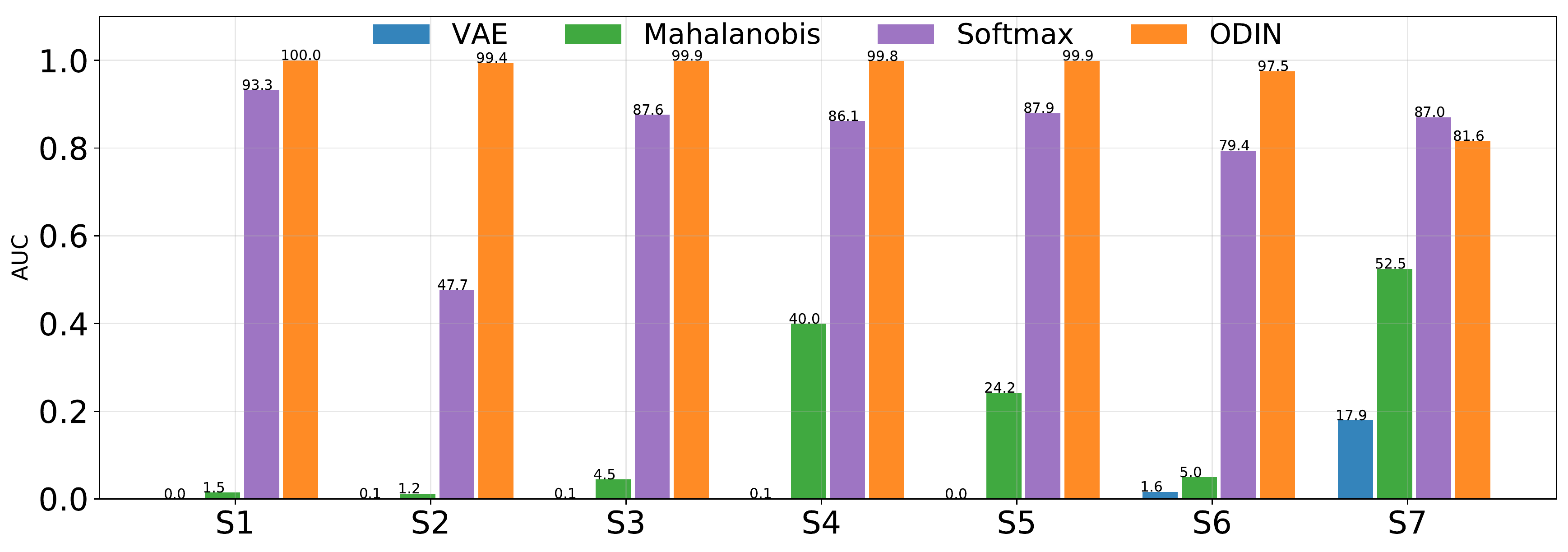}} }%
     \hfill
    \subfloat[{\footnotesize Last Stage AUC~($\%$),~\LwF}]{{\includegraphics[width=.33\textwidth]{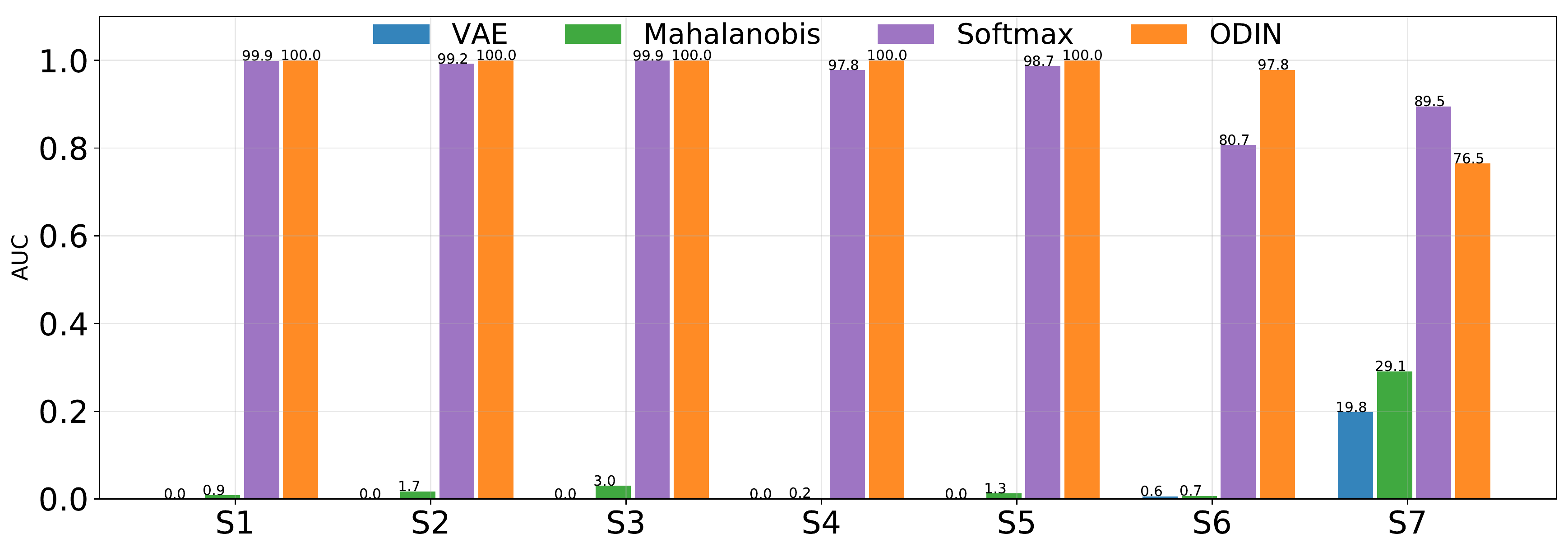}}}
        \hfill
    \subfloat[{\footnotesize Last Stage AUC~($\%$),~ \MAS}]{{\includegraphics[width=.32\textwidth]{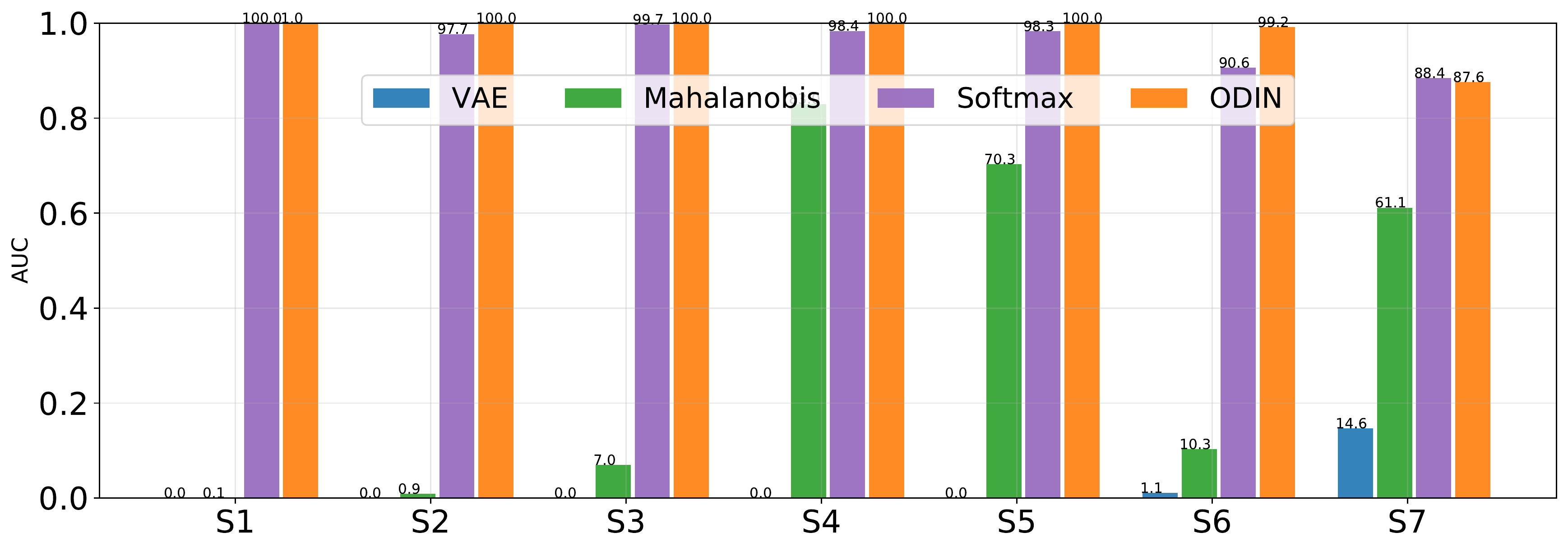}}}
\caption{\footnotesize Novelty Detection Last Stage AUC. The AUC scores for each \IN set (learnt at each previous stage) vs. the last \Out set based on the model before the last stag $\mathcal{M}_{T-1}$  for  \tasks. }%
    \label{fig:8task_laststage}%
    \end{figure*}

\begin{figure*}[h]
  \vspace*{-0.2cm}
    \centering
    \subfloat[{\footnotesize Last Stage AUC~($\%$),~ \fine}]{{\includegraphics[width=.33\textwidth]{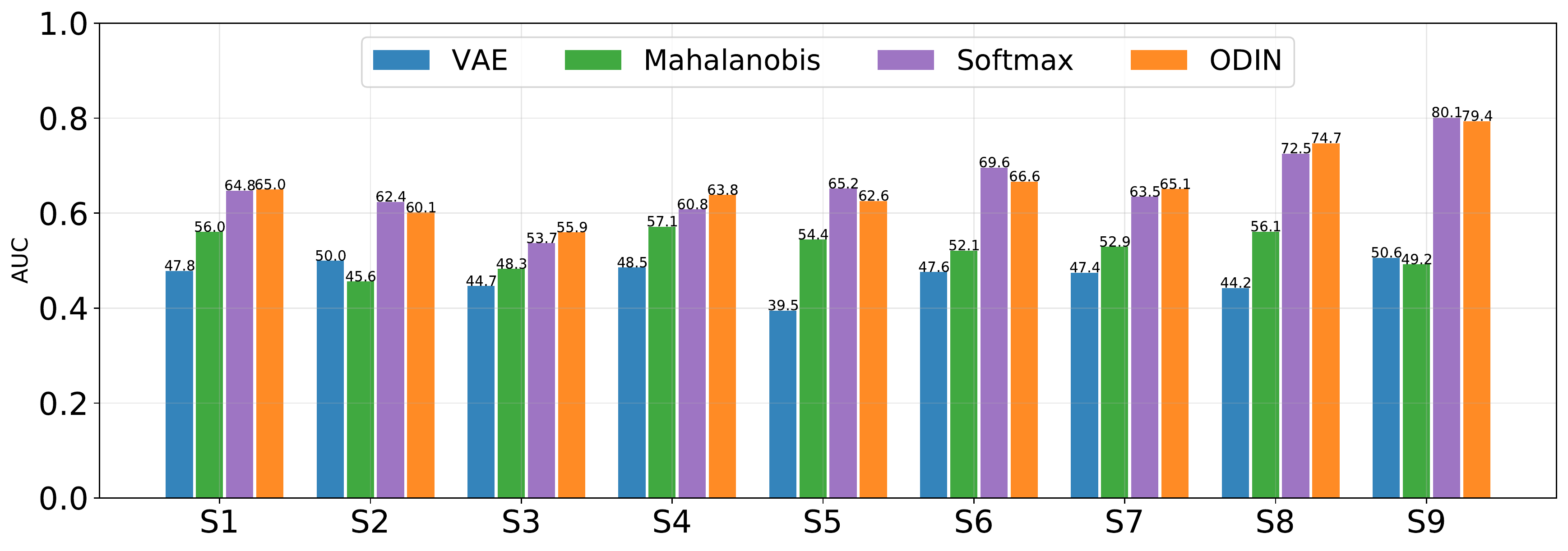}}} %
     \hfill
    \subfloat[{\footnotesize Last Stage AUC~($\%$),~\LwF}]{{\includegraphics[width=.33\textwidth]{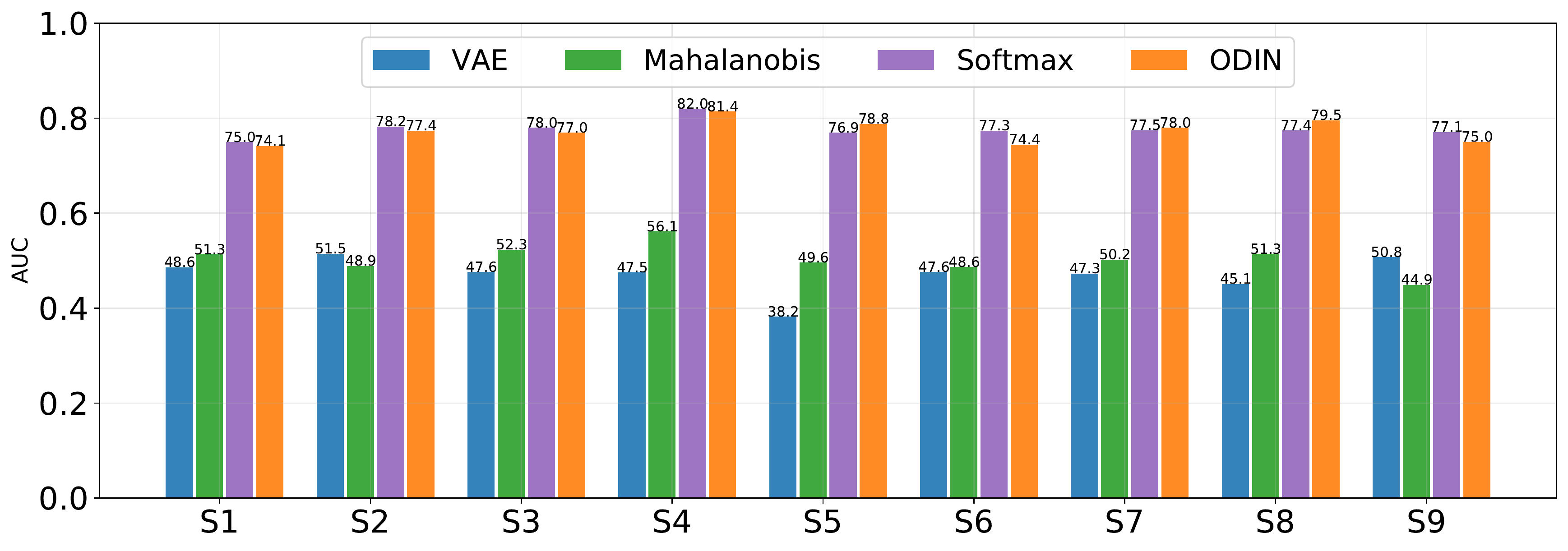}}}
        \hfill
    \subfloat[{\footnotesize Last Stage AUC~($\%$),~ \MAS}]{{\includegraphics[width=.32\textwidth]{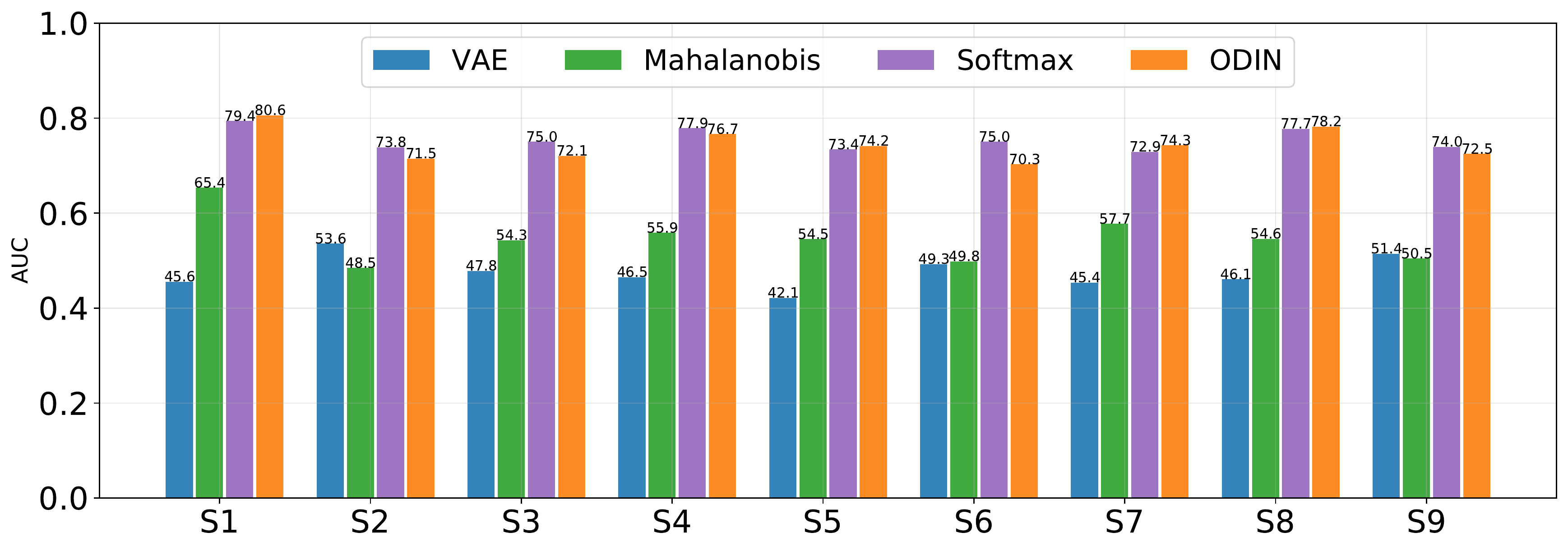}}}
\caption{\footnotesize  Last Stage AUC. The AUC scores for each \IN set (learnt at each previous stage) vs. the last \Out set based on the model before the last stag $\mathcal{M}_{T-1}$ for   multi-head \tinyimg with ResNet.}%
    \label{fig:tiny_multihead_resnet_laststage}%
    \end{figure*}

\begin{figure*}[h]
  \vspace*{-0.2cm}
    \centering
    \subfloat[{\footnotesize Last Stage AUC~($\%$),~ ER 25s}]{{\includegraphics[width=.44\textwidth]{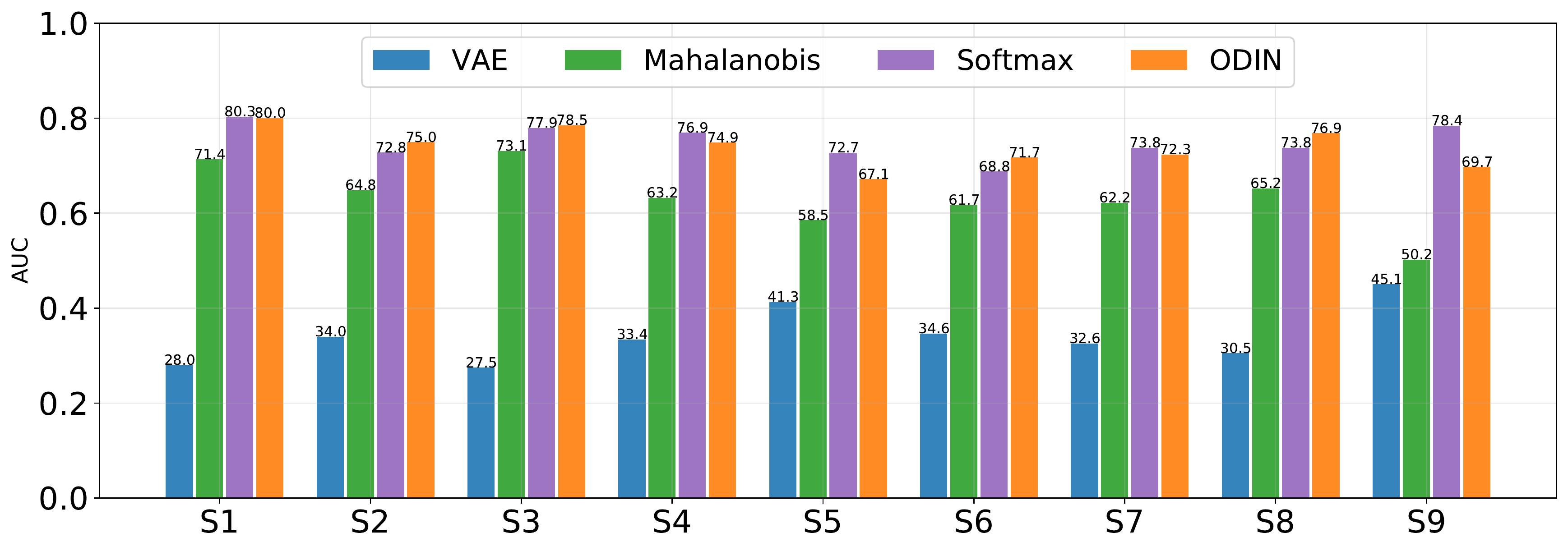}}}
        \hfill
    \subfloat[{\footnotesize Last Stage AUC~($\%$),~ER 50s}]{{\includegraphics[width=.44\textwidth]{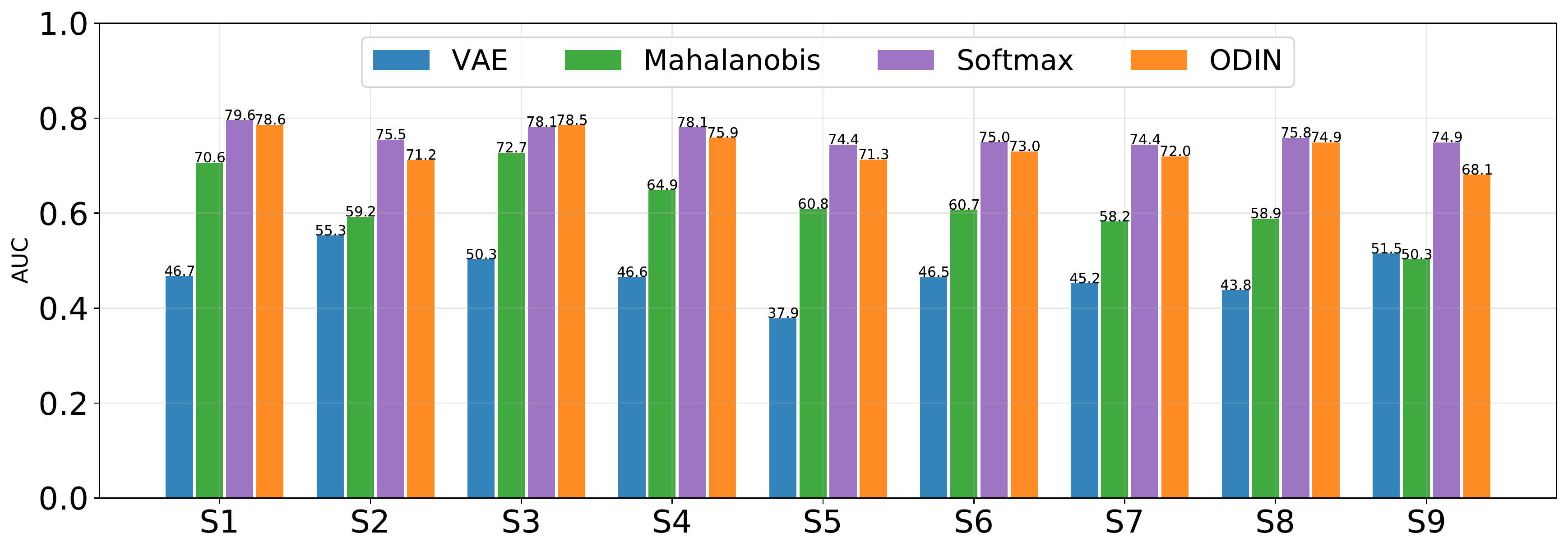}}}
\caption{\footnotesize  Last Stage AUC. The AUC scores for each \IN set (learnt at each previous stage) vs. the last \Out set based on the model before the last stag $\mathcal{M}_{T-1}$ for   shared-head \tinyimg with ResNet with \ER}%
    \label{fig:tiny_sharedhead_resnet_laststage}%
    \end{figure*}

\begin{figure*}[h]
  \vspace*{-0.2cm}
    \centering
    \subfloat[{\footnotesize Last Stage AUC~($\%$),~ SSIL 25s}]{{\includegraphics[width=.44\textwidth]{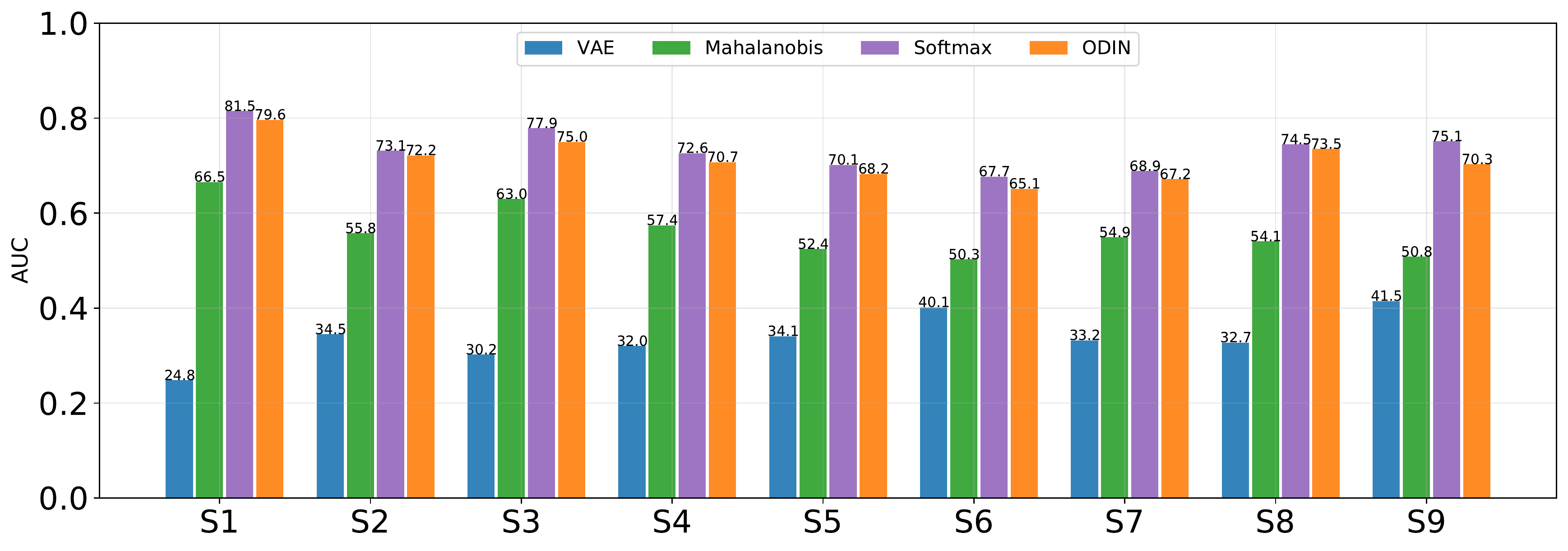}}}
        \hfill
    \subfloat[{\footnotesize Last Stage AUC~($\%$),~SSIL 50s}]{{\includegraphics[width=.44\textwidth]{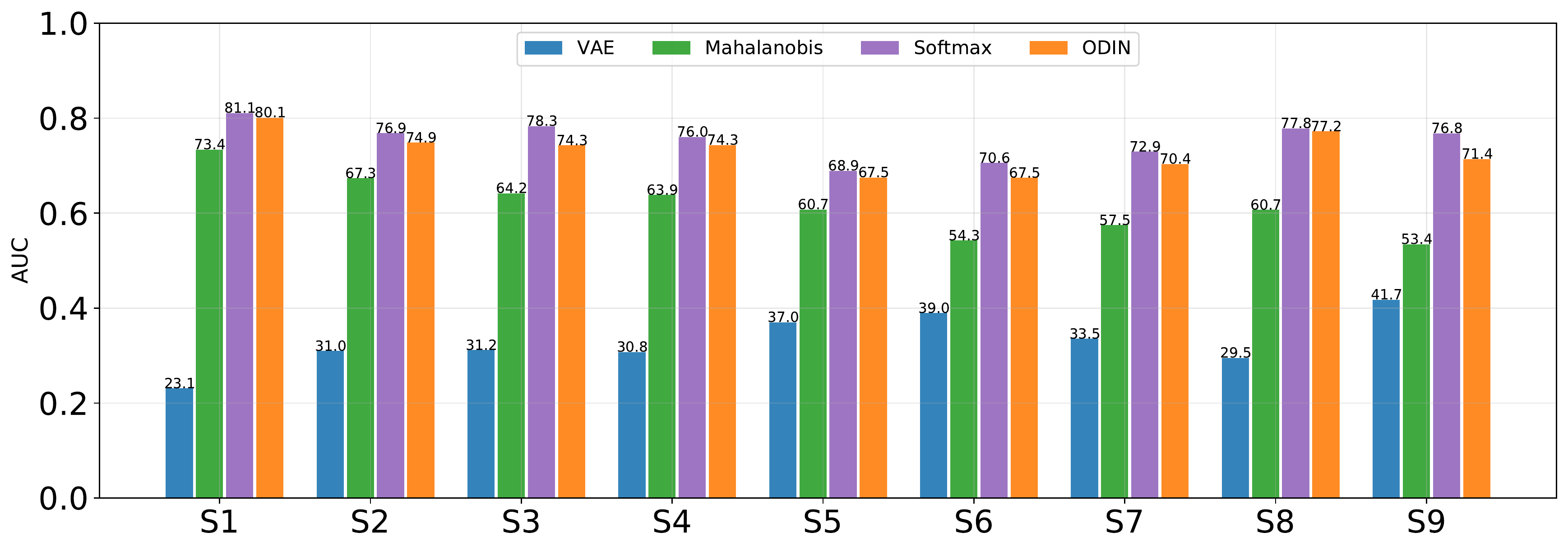}}}
\caption{\footnotesize Last Stage AUC. The AUC scores for each \IN set (learnt at each previous stage) vs. the last \Out set based on the model before the last stag $\mathcal{M}_{T-1}$ for   shared-head \tinyimg with ResNet with \SSIL..}%
    \label{fig:tiny_sharedhead_resnet_laststage_SSIL}%
    \end{figure*}

    In the main paper, we note that the ND performance could differ for the different considered \IN sets. Due to space limit, in the main paper we only report the Previous Stages (P.AUC) and the most Recent Stage (R.AUC) to discriminate between previously learnt sets and the newly learnt one. Here, we report the AUC score for each \IN set separately. To have an overall view, we consider the model before the last learning stage ($\mathcal{M}_{T-1}$) and we  construct  \IN sets for tasks or classes learnt at stages $1 \dots T-1$ while the \Out set represents the samples of the last task in the considered sequences. Figures~\ref{fig:8task_laststage},~\ref{fig:tiny_multihead_resnet_laststage},~\ref{fig:tiny_sharedhead_resnet_laststage},~\ref{fig:tiny_sharedhead_resnet_laststage_SSIL} show the last stage AUC for \tasks, multi-head \tinyimg and shared-head \tinyimg~(\ER~and \SSIL) respectively. In most cases (except from \LwF with \tinyimg), the AUC scores of the first learnt \IN set remain higher than that for the consequently learnt \IN sets (i.e., at stage $2$ or $3$). This indicates that the isolated learning of the classes representing the first set was more stable leading to less forgetting rates compared to the later learnt classes.
    \subsection{Forgotten Set (Additional Results)\label{sec:forg}}
    In the main paper, due to space limit, we have shown the detection scores of \F samples on 3 selected stages only. Here we show the full detection scores at all the different stages. Table~\ref{tab:ForgottenvsInvsOut} report the detection errors DER of \IN vs. \F and \F vs. \Out at the different learning stages and under the different considered settings. Table~\ref{tab:ForgottenvsInvsOutAUC} reports the AUC scores. 
      \begin{table*}[ht]
\centering
\tabcolsep=0.11cm
\small
 \begin{tabular}{ |l|l|l|}
 \cellcolor{green!20} low detection error &\cellcolor{yellow!20}medium detection error &\cellcolor{red!20}high detection error\\
\end{tabular}
\resizebox{\textwidth}{!}{
\begin{tabular}{ |l|l|ll|ll||ll|ll||ll|ll|ll|ll|ll|  }
 \hline
Setting& Method&\multicolumn{2}{|c|}{M-S2}& \multicolumn{2}{|c|}{M-S3}& \multicolumn{2}{|c|}{M-S4}& \multicolumn{2}{|c|}{M-S5}& \multicolumn{2}{|c|}{M-S6}&\multicolumn{2}{|c|}{M-S7} &\multicolumn{2}{|c|}{M-S8} &\multicolumn{2}{|c|}{M-S9} &\multicolumn{2}{|c|}{M-S10}  \\
 \hline
& &F. \vs N.&F \vs O.&F. \vs N.&F \vs O.&F. \vs N.&F \vs O.&F. \vs N.&F \vs O.&F. \vs N.&F \vs O.&F. \vs N.&F \vs O.&F. \vs N.&F \vs O.&F. \vs N.&F \vs O.&F. \vs N.&F \vs O.\\
 \hline
 \multirow{4}{4em}{Eight-Tasks \fine}   &\nav& \cellcolor{green!20}19.2 & \cellcolor{green!20}17.5&26.8 & 18.6&29.3 & 22.8&35.5 & 16.1&32.5 & 9.4&\cellcolor{yellow!20}33.5&\cellcolor{red!20} 35.3&\cellcolor{red!20}36.6 & -  &-&-&-&-   \\
&\odin &\cellcolor{yellow!20}32.5 &\cellcolor{green!20} 7.6&35.8 & 10.9&37.3 & 7.1&37.5 & 4.4&37.1 & 0.7&\cellcolor{red!20}35.7&\cellcolor{green!20} 4.2&\cellcolor{red!20}36.9& -  &-&-&-&-   \\
&\mah  &\cellcolor{red!20} 49.9 & \cellcolor{green!20}16.8&47.7 & 39.5&49.3 & 19.0&49.7 & 12.0&49.3 & 4.9&\cellcolor{red!20}49.1 & \cellcolor{red!20}46.6&\cellcolor{red!20}49.4 & -  &-&-&-&-   \\
&\vae & 50.0 & 50.0&47.6 & 50.0&48.7 & 50.0&48.9 & 50.0&48.8 & 50.0&48.8& 50.0&49.5 & -  &-&-&-&-   \\
\hline
\multirow{4}{4em}{Eight-Tasks MAS} & \nav &\cellcolor{green!20} 7.6 &\cellcolor{green!20} 12.4&12.1 & 15.5&15.0 & 9.8&\cellcolor{yellow!20}21.2 & 7.6&22.1 & 5.5&21.3 & \cellcolor{green!20}14.6&\cellcolor{yellow!20}22.3 & -  &-&-&-&-  \\
  &\odin &\cellcolor{yellow!20} 26.9 &\cellcolor{green!20} 3.2&23.9 & 7.1&27.9 & 5.3&31.9 & 3.8&32.4 & 0.6&31.6 & \cellcolor{green!20}2.1&\cellcolor{yellow!20}31.2 & -  &-&-&-&-  \\
  &\mah &\cellcolor{red!20} 49.9 &\cellcolor{green!20} 10.2&45.9 &\cellcolor{red!20} 40.4&48.0 & \cellcolor{yellow!20}24.6&49.1 &\cellcolor{green!20} 8.4&47.6 & 2.5&48.4 &\cellcolor{red!20} 38.2&\cellcolor{green!20}11.8 & -   &-&-&-&-    \\
  &\vae & 49.8 & 50.0&48.6 & 49.9&49.3 & 50.0&49.3 & 50.0&49.0 & 50.0&48.8 & 50.0&49.0 & -  &-&-&-&-      \\  
\hline
\multirow{4}{4em}{Eight-Tasks \LwF}   &\nav &\cellcolor{green!20} 15.4 & \cellcolor{green!20} 15.3&14.5 & 16.1&17.9 & 12.5&17.6 & 11.9&17.9 & 9.1&\cellcolor{yellow!20}20.0 &\cellcolor{green!20} 16.2&\cellcolor{yellow!20}24.7 & -  &-&-&-&-   \\
  &\odin &\cellcolor{yellow!20} 30.7 & \cellcolor{green!20}10.5&26.8 & 10.2&30.5 & 6.7&32.8 & 4.2&33.3 & 0.4&33.1 & \cellcolor{green!20}1.3&\cellcolor{red!20}35.0 & - &-&-&-&-\\
  &\mah & \cellcolor{red!20}49.8 & \cellcolor{green!20} 19.7&49.4 & 13.2&48.8 & 14.6&48.2 & 8.2&46.7 & 2.7&48.8 & \cellcolor{red!20}49.9&43.3 & - &-&-&-&- \\
  &\vae & 49.8 & 50.0&48.6 & 49.9&49.3 & 50.0&49.3 & 50.0&49.0 & 50.0&48.8 & 50.0&49.0 & -  &-&-&-&-      \\  
\hline
\hline
\multirow{4}{4em}{TinyImag. Seq.  \fine} &\nav  & \cellcolor{yellow!20} 31.3 & 48.3&33.8 & 49.0&34.5 & 49.6&34.5 & 49.5&34.2 & 49.5&\cellcolor{red!20}35.7 & 49.2&35.7 & 49.3&35.6 & 49.4&\cellcolor{red!20}36.4 & - \\ 

& \odin & \cellcolor{yellow!20}33.9 & 46.7&33.7 & 49.1&\cellcolor{red!20}36.0 & 49.1&35.7 & 49.2&36.7 & 48.4&36.9 & 48.6&36.7 & 48.4&36.5 & 49.3&\cellcolor{red!20}36.2 & -    \\ 

&\mah &   \cellcolor{red!20}44.8 & 46.2&46.1 & 47.5&46.3 & 47.8&46.8 & 47.8&46.2 & 47.2&46.9 & 47.5&46.9 & 47.3&45.9 & 48.0&\cellcolor{red!20}45.5 & -  \\ 
&\vae &  46.7 & 47.6&46.2 & 46.1&46.9 & 46.2&46.7 & 46.3&46.4 & 47.3&45.3 & 47.1&46.4 & 47.7&47.2 & 47.9&46.7 & - \\ 
\hline
\multirow{4}{4em}{TinyImag. Seq. \MAS} & \nav  & \cellcolor{green!20} 15.4 & 49.7&16.3 & 49.7&15.7 & 48.0&16.9 & 48.3&15.4 & 49.0&16.0 & 49.2&16.2 & 48.6&18.9 & 48.6&\cellcolor{green!20} 16.3 & -   \\ 

& \odin  &\cellcolor{yellow!20} 24.4 & 48.0&24.5 & 48.8&27.2 & 45.6&25.3 & 47.3&24.9 & 48.0&25.9 & 47.3&26.5 & 47.0&29.0 & 47.0&\cellcolor{yellow!20} 27.0 & -   \\ 

&\mah & \cellcolor{red!20}42.9 & 43.8&42.7 & 46.2&43.4 & 46.3&41.8 & 47.0&43.0 & 46.6&43.0 & 46.1&43.4 & 45.4&43.2 & 44.4&\cellcolor{red!20} 43.0 & -  \\ 
&\vae &  46.7 & 47.6&46.2 & 46.1&46.9 & 46.2&46.7 & 46.3&46.4 & 47.3&45.3 & 47.1&46.4 & 47.7&47.2 & 47.9&46.7 & -  \\ 
\hline
\multirow{3}{4em}{TinyImag. Seq. \LwF} & \nav&\cellcolor{yellow!20} 20.6 & 49.5&\cellcolor{green!20}19.5 & 48.9&18.0 & 48.4&19.3 & 47.6&17.3 & 48.8&18.0 & 47.9&16.7 & 48.4&18.5 & 48.3&\cellcolor{green!20}17.9 & -   \\ 

& \odin & \cellcolor{yellow!20}24.2 & 49.4&24.2 & 45.7&25.6 & 46.4&26.8 & 42.3&26.6 & 43.6&25.2 & 44.0&24.9 & 45.5&26.6 & 46.1&\cellcolor{yellow!20}26.5 & -  \\ 

&\mah &  \cellcolor{red!20}45.7 & 46.0&43.2 & 47.5&42.4 & 46.9&44.5 & 46.3&44.9 & 45.4&43.9 & 46.4&43.0 & 45.1&43.2 & 47.8&\cellcolor{red!20}44.9 & -   \\ 
&\vae &  46.7 & 47.6&46.2 & 46.1&46.9 & 46.2&46.7 & 46.3&46.4 & 47.3&45.3 & 47.1&46.4 & 47.7&47.2 & 47.9&46.7 & - \\
\hline
\hline
\multirow{4}{4em}{TinyImag. Seq.  \ER 25s}&\nav &\cellcolor{yellow!20} 21.3 & 46.3&29.5 & 46.3&27.2 & 47.2&26.9 & 46.9&25.7 & 48.6&27.2 & 47.7&27.0 & 47.2&30.6 & 46.5&\cellcolor{yellow!20}30.8 & -    \\  
&\odin &\cellcolor{red!20} 29.6 & 40.3&35.4 & 41.6&31.9 & 44.2&31.1 & 45.4&31.0 & 46.1&31.2 & 45.6&31.8 & 45.1&32.3 & 45.5&\cellcolor{yellow!20}32.1 & -   \\  
&\mah & \cellcolor{red!20}37.5 & 46.1&36.3 & 44.0&37.3 & 44.7&37.3 & 47.2&40.1 & 47.9&40.7 & 46.9&40.4 & 45.5&39.4 & 47.0&\cellcolor{red!20}38.5 & -\\
&\vae & 45.3 & 45.1&47.5 & 44.4&46.0 & 45.6&46.5 & 47.1&46.0 & 46.9&47.9 & 46.4&46.9 & 48.8&47.2 & 48.6&46.8 & -   \\ 
\hline
\multirow{4}{4em}{TinyImag. Seq.  \ER 50s} & \nav &\cellcolor{green!20} 19.9 & 46.4&\cellcolor{yellow!20}30.5 & 44.3&23.3 & 46.3&25.3 & 45.5&27.4 & 45.4&26.1 & 46.0&25.9 & 45.8&27.3 & 46.9&\cellcolor{yellow!20}29.5 & -   \\ 
&\odin & \cellcolor{yellow!20}28.2 & 45.5&32.4 & 40.5&31.9 & 43.6&30.9 & 43.8&34.4 & 40.9&32.7 & 43.7&32.1 & 42.2&32.4 & 45.2&\cellcolor{yellow!20}31.9 & -   \\
&\mah & \cellcolor{red!20}47.3 & 43.8&\cellcolor{yellow!20}32.8 & 43.9&\cellcolor{red!20}38.7 & 43.9&41.7 & 44.2&39.0 & 43.6&40.6 & 44.0&40.1 & 43.3&40.0 & 45.4&\cellcolor{red!20}39.3 & -  \\  
&\vae & 45.3 & 45.1&47.5 & 44.4&46.0 & 45.6&46.5 & 47.1&46.0 & 46.9&47.9 & 46.4&46.9 & 48.8&47.2 & 48.6&46.8 & -   \\ 
\hline
\multirow{4}{4em}{TinyImag. Seq.  \SSIL 25s}&\nav & \cellcolor{yellow!20}24.0 & 46.3&30.7 & 39.6&28.9 & 41.8&31.1 & 41.2&32.4 & 41.1&31.5 & 42.5&34.1 & 42.0&35.8 & 42.6&\cellcolor{yellow!20}34.1 & - \\ 
&\odin & \cellcolor{yellow!20}28.7 & 41.5&\cellcolor{red!20}36.4 & 37.9&34.7 & 40.4&36.1 & 39.6&38.7 & 39.2&37.4 & 40.6&39.8 & 39.4&40.3 & 39.7&\cellcolor{red!20}38.3 & - \\ 
&\mah & 49.1 & 48.4&46.4 & 48.0&46.4 & 47.9&47.9 & 49.0&47.0 & 48.3&44.1 & 45.5&45.1 & 47.9&43.9 & 45.6&42.3 & - \\ 
&\vae & 49.9 & 49.1&48.5 & 49.2&49.3 & 49.7&49.3 & 49.6&49.2 & 49.6&49.1 & 49.7&49.1 & 49.6&49.3 & 49.6&49.2 & - \\ 
\hline
\multirow{4}{4em}{TinyImag. Seq.  \SSIL 50s}&\nav &\cellcolor{yellow!20} 24.7 & 43.9&31.9 & 37.9&31.0 & 40.0&30.1 & 43.4&32.2 & 41.9&29.8 & 43.4&33.9 & 42.3&33.3 & 42.4&\cellcolor{yellow!20}33.2 & - \\ 
&\odin &\cellcolor{yellow!20} 29.4 & 40.2&\cellcolor{red!20}38.6 & 36.0&37.1 & 37.9&34.9 & 40.6&37.3 & 39.3&35.3 & 41.4&39.0 & 39.8&37.8 & 40.5&\cellcolor{red!20}37.7 & - \\ 
&\mah & 48.5 & 47.7&38.5 & 48.4&42.7 & 44.3&42.2 & 45.1&42.4 & 45.5&42.2 & 44.4&43.9 & 44.1&42.6 & 42.4&42.5 & - \\ 
&\vae & 50.0 & 49.1&49.2 & 49.6&49.4 & 49.8&49.4 & 49.5&49.4 & 49.5&49.5 & 49.3&49.3 & 49.6&49.4 & 49.7&49.2 & - \\ 
\hline
\end{tabular}}
\caption{\small Detection errors of Forgotten samples (F) versus \IN samples (N)   and versus \Out samples (O) under the different setting. }
 \label{tab:ForgottenvsInvsOut}
 \end{table*}
   \begin{table*}[ht]
\centering
\tabcolsep=0.11cm
\small
\resizebox{\textwidth}{!}{
\begin{tabular}{ |l|l|ll|ll||ll|ll||ll|ll|ll|ll|ll|  }
 \hline
Setting& Method&\multicolumn{2}{|c|}{M-S2}& \multicolumn{2}{|c|}{M-S3}& \multicolumn{2}{|c|}{M-S4}& \multicolumn{2}{|c|}{M-S5}& \multicolumn{2}{|c|}{M-S6}&\multicolumn{2}{|c|}{M-S7} &\multicolumn{2}{|c|}{M-S8} &\multicolumn{2}{|c|}{M-S9} &\multicolumn{2}{|c|}{M-S10}  \\
 \hline
& &F. \vs N.&F \vs O.&F. \vs N.&F \vs O.&F. \vs N.&F \vs O.&F. \vs N.&F \vs O.&F. \vs N.&F \vs O.&F. \vs N.&F \vs O.&F. \vs N.&F \vs O.&F. \vs N.&F \vs O.&F. \vs N.&F \vs O.\\
 \hline
\multirow{4}{4em}{Eight-Tasks \fine} &\nav & 86.0 & 89.0&81.3 & 85.5&73.9 & 89.4&70.8 & 91.3&72.1 & 96.3&71.7 & 68.8&68.6 & -  &-&-&-&-  \\
&\odin & 71.5 & 98.7&70.0 & 94.0&67.4 & 96.4&66.7 & 97.8&66.9 & 99.7&68.8 & 95.0&67.0 & -  &-&-&-&-  \\ 
&\mah & 83.6 & 95.7&87.1 & 97.8&51.1 & 70.7&54.6 & 87.8&46.2 & 94.0&49.1 & 2.0&40.4 & - &-&-&-&-  \\
&\vae & 48.4 & 3.0&49.5 & 7.3&48.7 & 2.4&46.9 & 3.5&48.3 & 0.1&46.9 & 0.5&46.3 & -  &-&-&-&-  \\ 
\hline
\multirow{4}{4em}{Eight-Tasks MAS} & \nav & 95.4 & 90.1&92.7 & 86.3&90.3 & 91.7&84.7 & 95.6&83.8 & 97.9&84.5 & 90.3&83.4 & - &-&-&-&-  \\
&\odin & 77.8 & 99.1&81.7 & 95.0&76.7 & 97.1&73.1 & 98.4&72.5 & 99.9&73.5 & 99.4&73.4 & &-&-&-&-  \\
&\mah & 80.3 & 93.6&78.5 & 92.2&41.7 & 77.8&69.3 & 87.7&59.9 & 97.8&45.2 & 5.7&57.4 & -  &-&-&-&-  \\
&\vae & 47.2 & 3.5&43.4 & 9.3&42.0 & 2.2&44.2 & 2.8&44.9 & 0.1&42.7 & 0.2&42.0 & -  \\  
\hline
\multirow{4}{4em}{Eight-Tasks \LwF}   &\nav & 90.4 & 90.3&90.6 & 85.8&88.5 & 90.1&88.1 & 92.3&87.6 & 95.8&86.1 & 85.7&81.4 & -  &-&-&-&-  \\
&\odin & 75.9 & 95.8&78.1 & 95.1&74.2 & 97.6&72.2 & 98.3&71.2 & 100.0&71.0 & 99.2&68.9 & -                   &-&-&-&-  \\
&\mah & 61.6 & 87.3&72.9 & 92.3&56.5 & 81.7&50.9 & 90.3&51.8 & 96.6&42.2 & 0.9&6.7 & - &-&-&-&-  \\
&\vae & 42.8 & 7.9&41.4 & 12.7&45.4 & 4.2&46.6 & 2.8&45.8 & 0.0&46.3 & 0.1&46.6 & -  &-&-&-&-  \\
\hline
\hline
\multirow{4}{4em}{TinyImag. Seq.  \fine} &\nav & 73.5 & 48.0&70.7 & 47.6&70.2 & 46.5&69.1 & 47.3&70.1 & 46.1&68.1 & 47.1&67.7 & 47.0&68.5 & 45.3&67.1 & - 
\\
&\odin & 70.4 & 50.7&69.8 & 48.1&68.2 & 47.7&67.1 & 48.4&66.5 & 49.2&66.3 & 49.5&66.6 & 49.7&67.3 & 46.6&67.3 & -\\
&\mah & 52.7 & 53.3&53.1 & 51.2&52.1 & 50.4&51.5 & 50.3&52.5 & 51.3&51.3 & 50.9&51.4 & 51.4&53.4 & 49.9&53.7 & - \\
&\vae & 50.2 & 47.8&51.9 & 52.1&52.6 & 51.2&51.7 & 50.1&50.7 & 50.2&52.3 & 49.2&51.2 & 47.5&51.8 & 45.9&49.6 & - \\ 
\hline
\multirow{4}{4em}{TinyImag. Seq. \MAS} &\nav & 88.0 & 42.4&85.4 & 41.2&88.2 & 37.2&86.3 & 39.1&86.7 & 38.0&87.3 & 36.2&87.5 & 35.5&86.1 & 37.7&85.9 & -   \\
&\odin & 77.7 & 56.1&74.1 & 53.5&77.9 & 50.4&76.3 & 51.8&77.0 & 48.4&77.6 & 48.6&76.8 & 49.6&76.9 & 48.0&76.5 & -     \\
&\mah & 61.1 & 55.2&54.8 & 50.9&54.4 & 50.3&54.7 & 47.9&53.4 & 48.7&49.9 & 52.8&49.0 & 55.0&50.5 & 53.3&52.1 & -   \\
&\vae & 51.4 & 45.7&54.1 & 48.5&54.0 & 48.5&53.1 & 48.9&50.8 & 50.4&50.5 & 47.4&48.3 & 49.4&51.7 & 46.2&52.5 & - \\ 
\hline
\multirow{3}{4em}{TinyImag. Seq. \LwF} &\nav & 85.5 & 45.1&86.1 & 42.2&87.1 & 40.1&86.0 & 42.5&87.4 & 39.9&87.0 & 40.7&88.2 & 38.4&87.1 & 38.4&87.3 & - \\    &\odin & 83.7 & 48.0&80.1 & 51.2&80.3 & 50.9&77.8 & 55.4&78.6 & 53.2&79.1 & 52.1&79.8 & 51.4&78.8 & 50.0&78.1 & -  \\   
&\mah & 52.3 & 53.4&56.0 & 49.7&56.7 & 49.4&54.1 & 50.8&52.9 & 51.0&53.1 & 49.8&53.5 & 50.3&54.5 & 46.5&52.9 & -   \\         
&\vae & 50.2 & 47.1&51.8 & 51.2&50.7 & 52.0&49.2 & 52.8&50.6 & 52.0&49.3 & 52.8&49.5 & 50.5&50.4 & 47.2&50.5 & -  \\
\hline
\hline
\multirow{4}{4em}{TinyImag. Seq.  \ER 25s}&\nav & 84.3 & 47.8&75.3 & 51.9&77.6 & 50.6&79.0 & 50.4&80.2 & 47.7&78.5 & 50.0&78.8 & 48.7&74.3 & 51.4&73.5 & -     \\  &\odin & 74.0 & 57.2&69.0 & 59.8&72.4 & 55.9&74.0 & 54.9&73.3 & 53.0&72.9 & 53.5&73.0 & 54.8&71.3 & 54.3&72.1 & -             \\
&\mah & 63.9 & 50.8&67.5 & 53.7&65.0 & 54.7&65.5 & 52.1&61.2 & 50.2&60.9 & 52.4&60.6 & 54.0&62.8 & 52.4&63.0 & - \\
&\vae & 54.1 & 43.6&49.8 & 51.4&50.6 & 51.4&52.4 & 49.1&52.1 & 49.5&50.8 & 50.3&52.7 & 46.7&51.9 & 46.0&52.0 & -\\ 
\hline
\multirow{4}{4em}{TinyImag. Seq.  \ER 50s} & \nav & 85.7 & 48.4&73.8 & 54.7&81.9 & 48.9&80.2 & 51.3&77.5 & 53.0&78.6 & 50.8&78.9 & 51.5&77.7 & 49.9&75.7 & -   \\
& \odin & 77.7 & 52.5&71.5 & 61.9&71.2 & 56.7&72.0 & 55.6&68.5 & 59.7&70.8 & 56.1&70.9 & 58.0&71.4 & 53.9&72.2 & -    \\
& \mah & 63.5 & 51.4&71.7 & 55.0&61.5 & 55.0&60.0 & 54.5&60.8 & 55.8&59.7 & 54.0&59.6 & 56.3&60.0 & 53.6&62.1 & -   \\
& \vae & 56.9 & 41.7&47.3 & 54.0&50.3 & 52.3&51.5 & 50.1&49.0 & 51.5&46.4 & 53.5&50.2 & 47.8&49.3 & 48.0&49.8 & -   \\ 
\hline
\multirow{4}{4em}{TinyImag. Seq.  \SSIL 25s}&\nav & 82.7 & 52.8&72.0 & 61.1&76.0 & 58.4&73.7 & 58.0&70.6 & 59.5&72.2 & 56.7&69.0 & 57.9&67.6 & 57.3&69.7 & - \\ 
&\odin & 76.5 & 58.8&65.3 & 64.6&69.1 & 61.6&67.2 & 61.5&62.7 & 62.4&64.2 & 60.4&61.4 & 62.6&60.7 & 61.1&62.7 & - \\ 
&\mah & 30.7 & 47.8&50.9 & 43.7&48.9 & 46.1&46.0 & 45.5&47.0 & 46.4&54.6 & 53.2&53.1 & 48.0&54.7 & 52.3&55.4 & - \\ 
&\vae & 28.0 & 45.1&36.1 & 42.5&35.9 & 41.3&37.6 & 42.7&41.0 & 43.1&42.5 & 41.9&41.4 & 41.7&42.6 & 39.9&42.1 & - \\ 
\hline
\multirow{4}{4em}{TinyImag. Seq.  \SSIL 50s}&\nav & 79.9 & 54.8&71.4 & 63.4&72.8 & 61.1&75.0 & 57.0&72.2 & 58.6&74.5 & 55.6&69.4 & 58.8&70.1 & 57.8&70.6 & - \\ 
&\odin & 74.6 & 60.1&62.4 & 67.6&65.3 & 65.0&68.4 & 60.6&64.2 & 62.3&67.3 & 59.5&61.7 & 63.1&63.5 & 61.1&64.0 & - \\ 
&\mah & 33.5 & 49.4&62.5 & 48.1&54.7 & 55.5&54.8 & 53.2&57.8 & 52.8&56.1 & 55.2&54.3 & 54.8&56.4 & 57.4&56.2 & - \\ 
&\vae & 28.9 & 43.5&37.8 & 39.6&37.7 & 39.6&38.1 & 42.0&40.3 & 42.4&39.2 & 43.1&41.9 & 40.9&40.8 & 40.0&40.9 & - \\ 
\hline
\end{tabular}}
\caption{\small AUC scores in the detection of Forgotten samples (F) versus \IN samples (N)   and versus \Out samples (O) under the different setting. }
 \label{tab:ForgottenvsInvsOutAUC}
 \end{table*}
 

    \subsection{Additional Analysis~\label{sec:discussion}}
    In the main paper, in an attempt to gain more insights and open the door for further possible direction to tackle this problem, we present an analogy where the last layer classes weights are treated as prototypes for the different learnt categories.
        We retrained the models in \tinyimg with no bias term and obtained similar CL performance to that when using a bias term in the last layer.
However, for shared-head setting, we note an imbalance issue in the last layer weights, a problem spotted in previous works~\citep{zhao2020maintaining,wu2019large,hou2019learning} thus we train shared-head networks with a normalized classification layer as in~\cite{qi2018low}. 

\begin{figure*}[h!]
  \vspace*{-0.2cm}
    \centering
    \subfloat[{\footnotesize Features Norm with~ \fine}]{{\includegraphics[width=.33\textwidth]{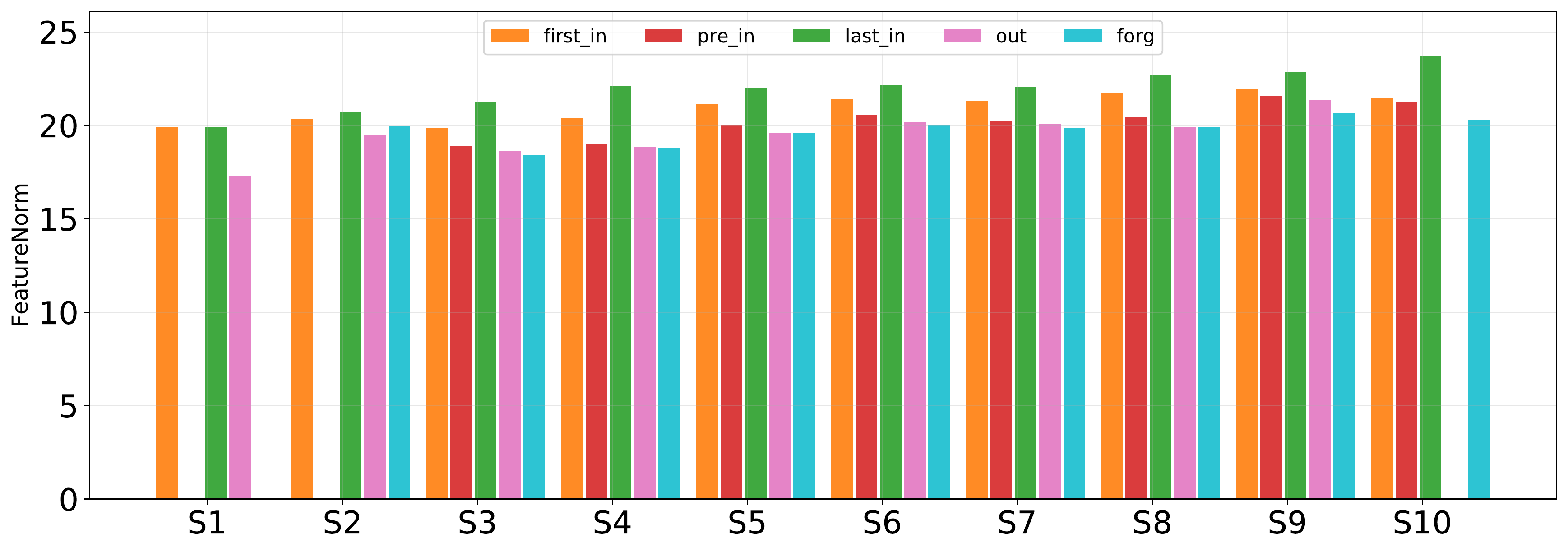}}} %
     \hfill
    \subfloat[{\footnotesize Features Norm~\LwF}]{{\includegraphics[width=.33\textwidth]{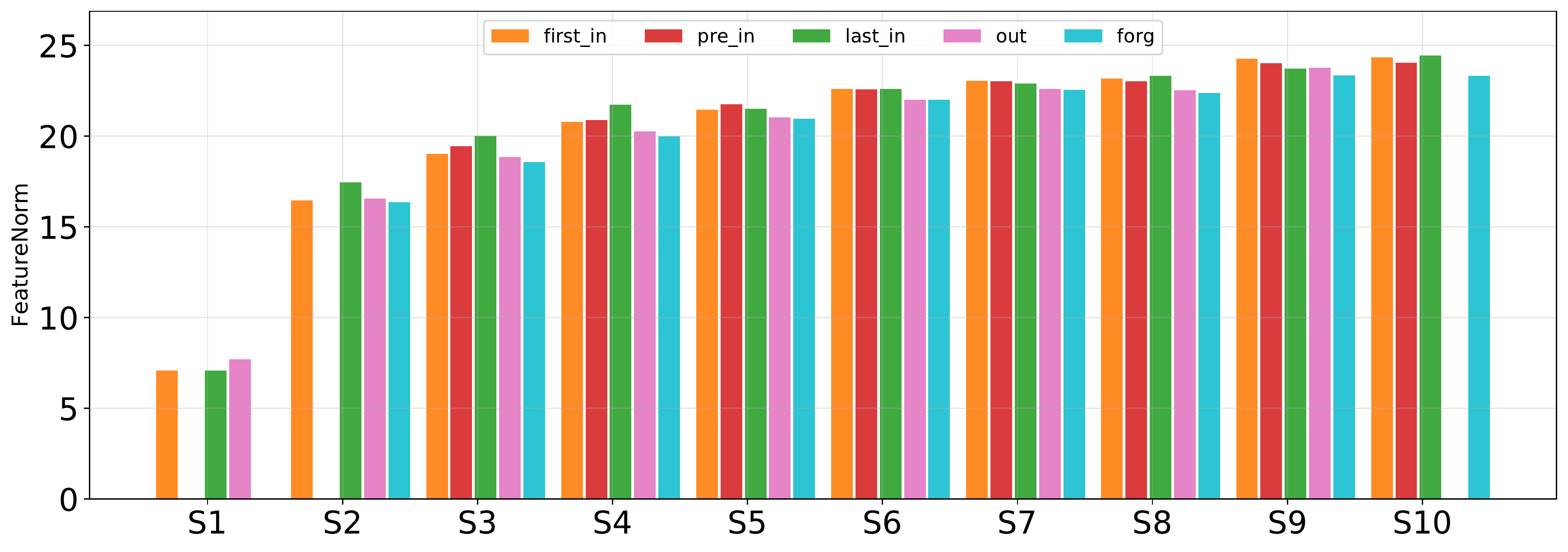}}}
        \hfill
    \subfloat[{\footnotesize Features Norm~ \MAS}]{{\includegraphics[width=.32\textwidth]{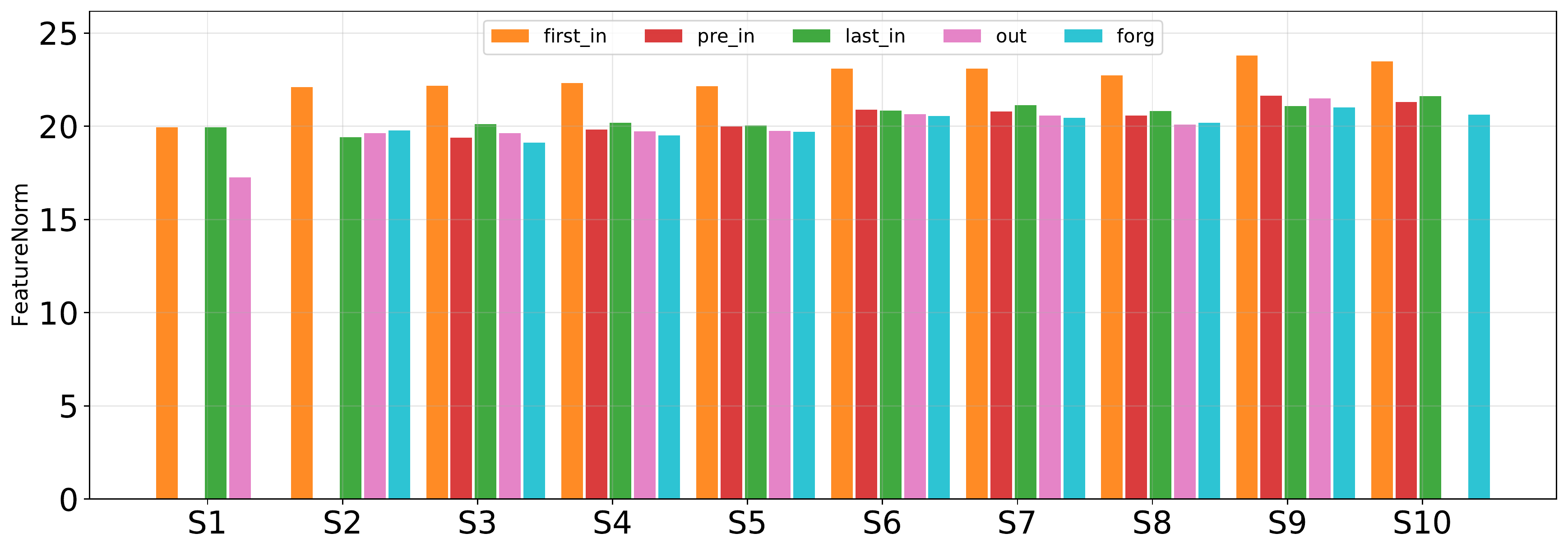}}}
\caption{\footnotesize Features norm of first \IN set, last \IN set and the in between \IN sets along with the features norm of \F set and \Out set for  multi-head \tinyimg with ResNet}%
    \label{fig:tiny_multihead_featnorm}%
    \end{figure*}

\begin{figure*}[h]
  \vspace*{-0.2cm}
    \centering
    \subfloat[{\footnotesize Features Norm with~ ER 25s}]{{\includegraphics[width=.49\textwidth]{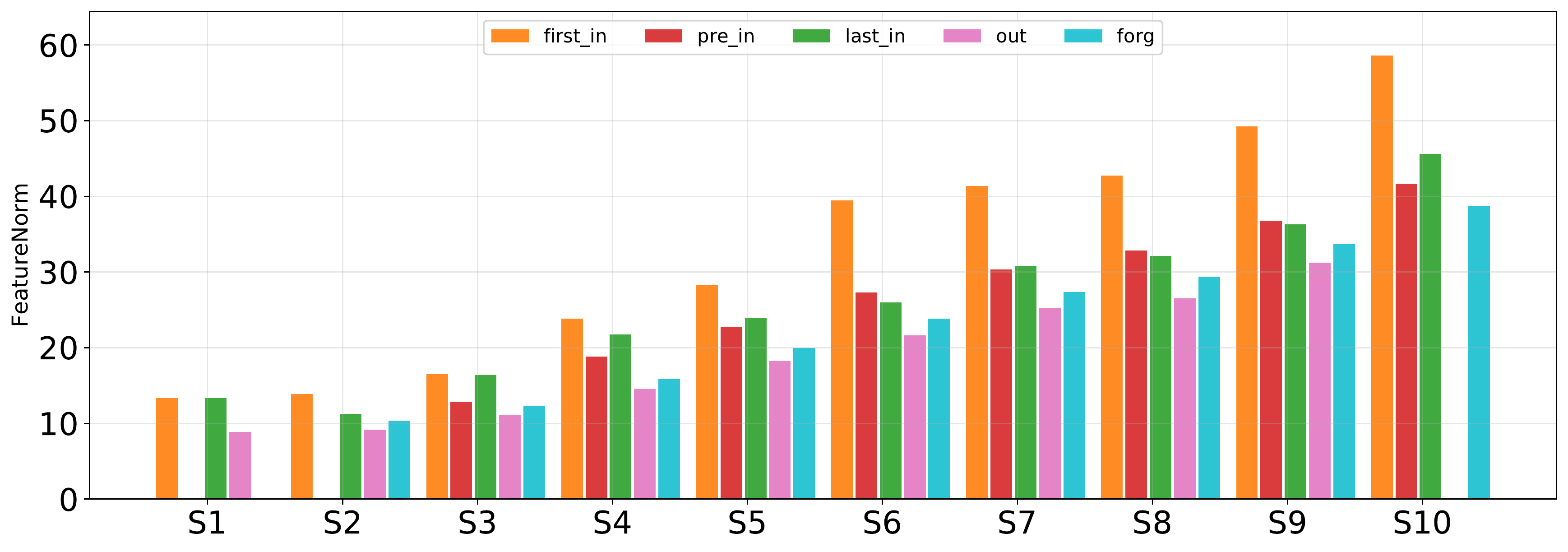}}}
        \hfill
    \subfloat[{\footnotesize Features Norm with~ER 50s}]{{\includegraphics[width=.49\textwidth]{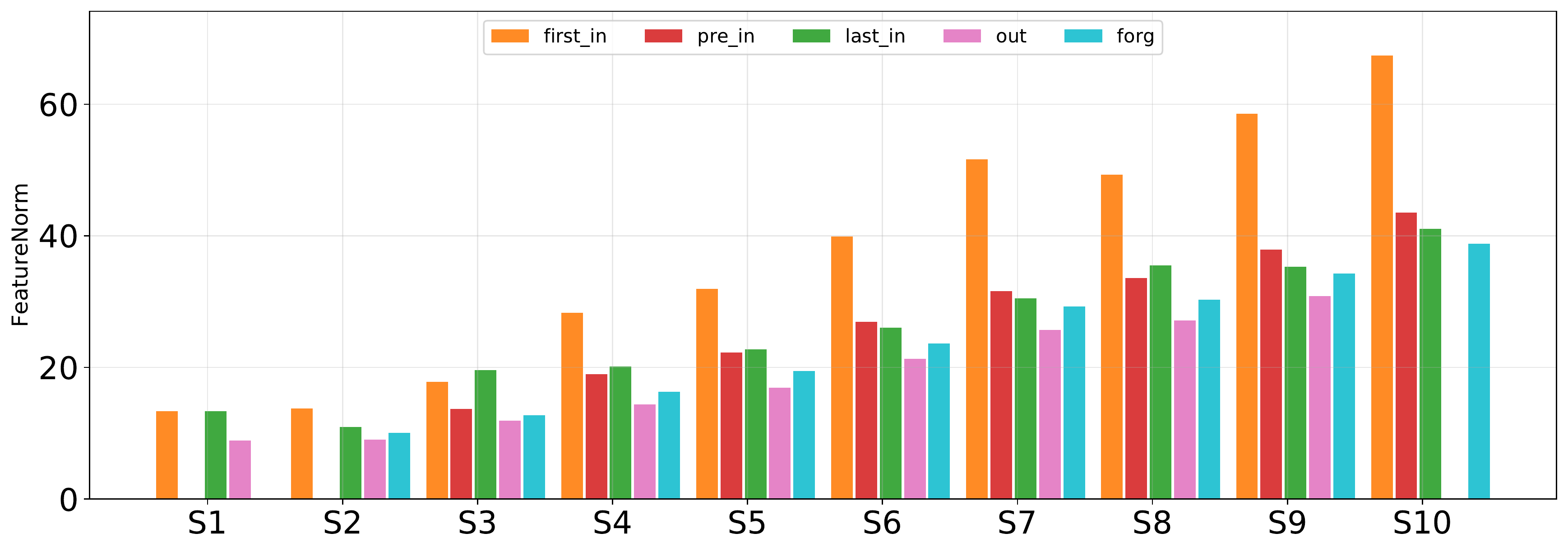}}}
\caption{\footnotesize Features norm of first \IN set, last \IN set and the in between \IN sets along with the features norm of \F set and \Out set for \ER~ under shared-head on \tinyimg with ResNet}%
    \label{fig:tiny_sharedhead_featnorm}%
    \end{figure*}

\begin{figure*}[h]
  \vspace*{-0.2cm}
    \centering
    \subfloat[{\footnotesize Features Norm with~ SSIL 25s}]{{\includegraphics[width=.49\textwidth]{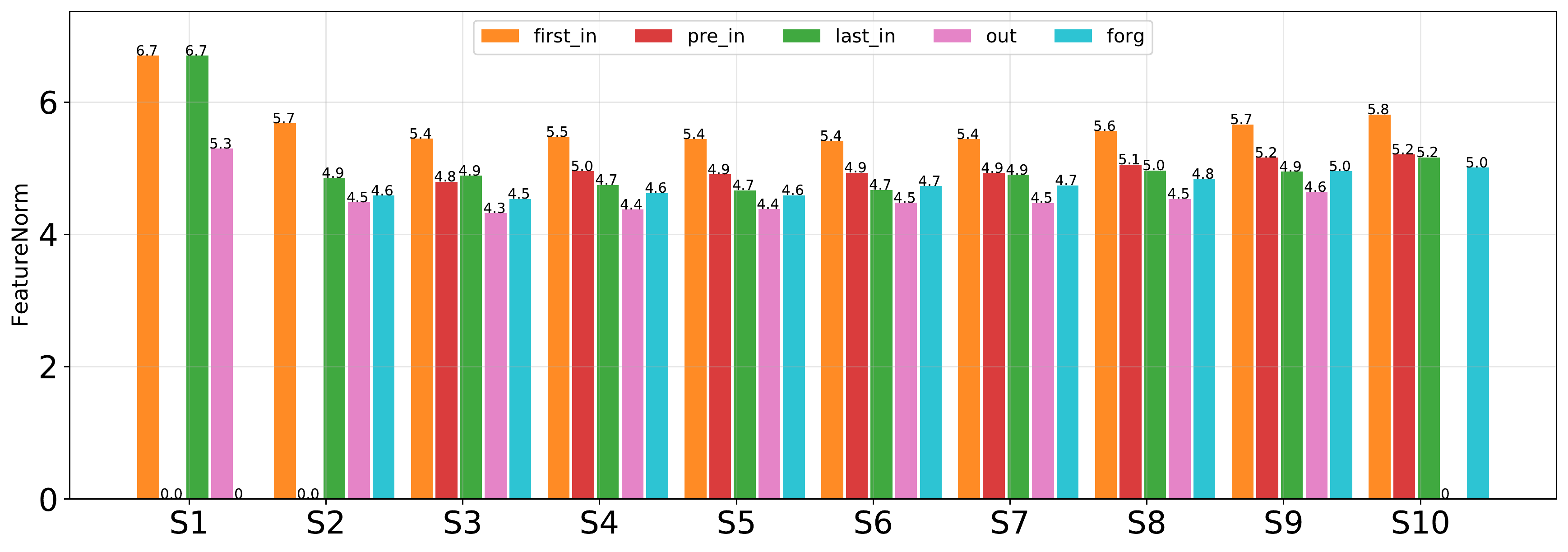}}}
        \hfill
    \subfloat[{\footnotesize Features Norm with~SSIL 50s}]{{\includegraphics[width=.49\textwidth]{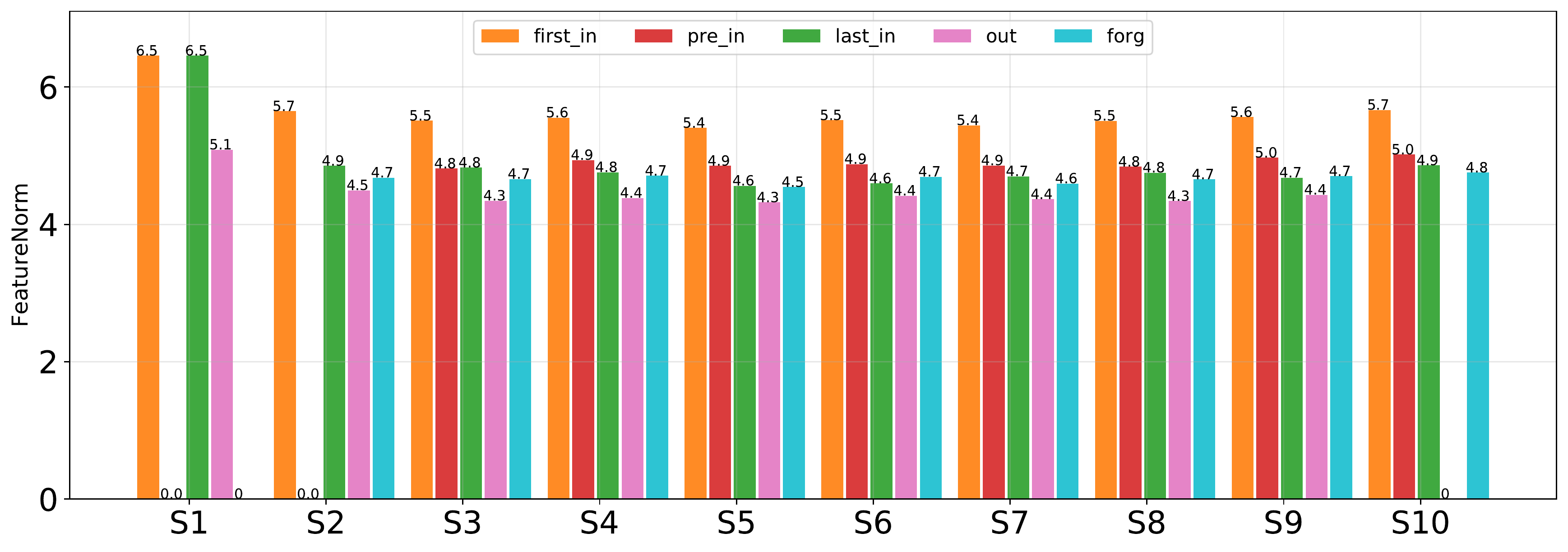}}}
\caption{\footnotesize Features norm of first \IN set, last \IN set and the in between \IN sets along with the features norm of \F set and \Out set for \SSIL~ under shared-head  on \tinyimg with ResNet.}%
    \label{fig:tiny_sharedhead_featnorm_SSIL}%
    \end{figure*}

\begin{figure*}[h]
  \vspace*{-0.2cm}
    \centering
    \subfloat[{\footnotesize Features Mean with~ \fine}]{{\includegraphics[width=.33\textwidth]{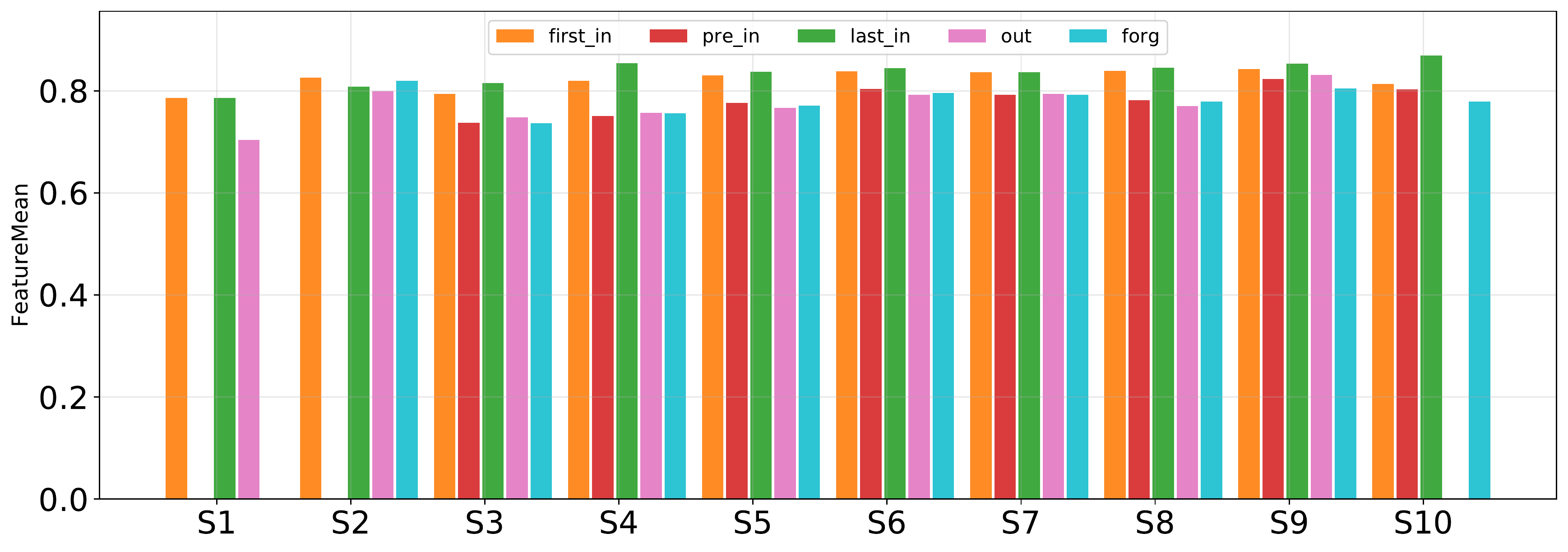}}} %
     \hfill
    \subfloat[{\footnotesize Features Mean~\LwF}]{{\includegraphics[width=.33\textwidth]{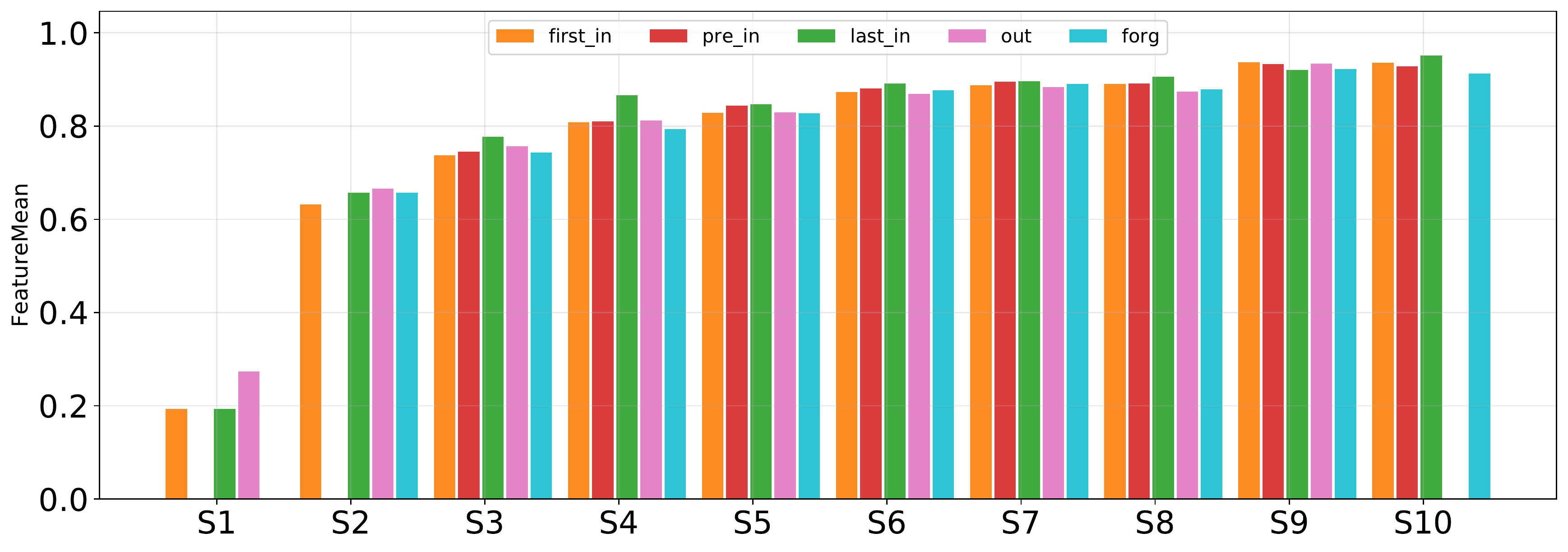}}}
        \hfill
    \subfloat[{\footnotesize Features Mean~ \MAS}]{{\includegraphics[width=.32\textwidth]{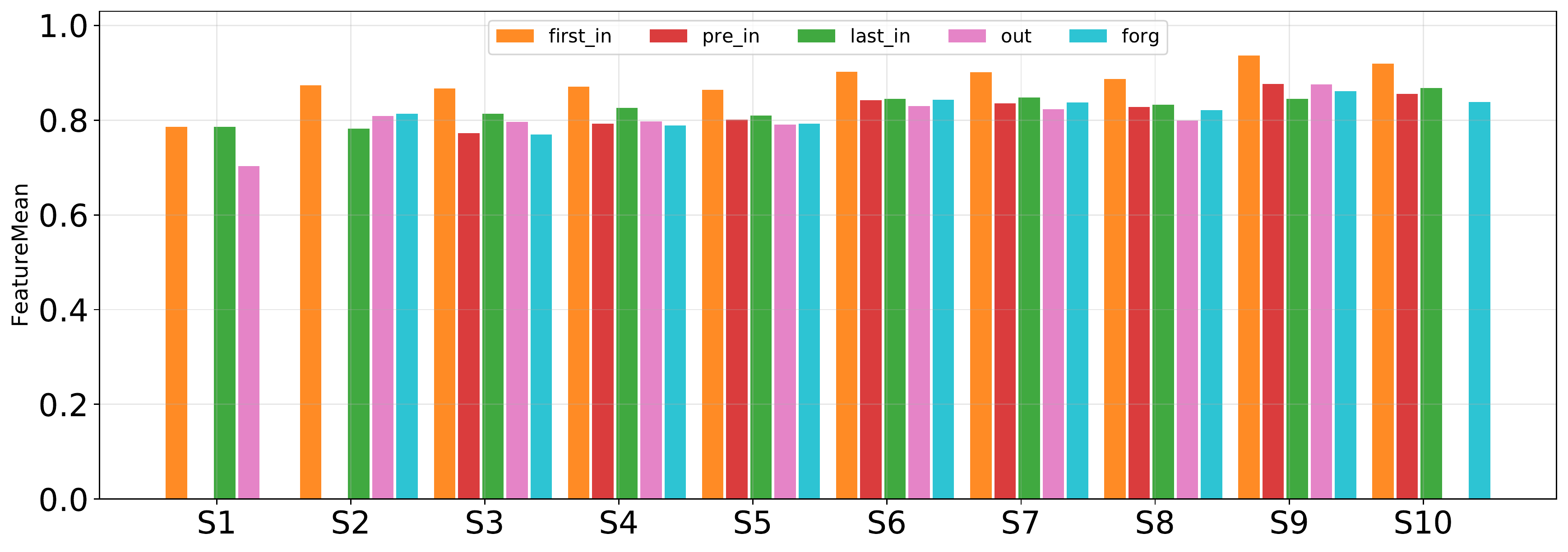}}}
\caption{\footnotesize Features mean of first \IN set, last \IN set and the in between \IN sets along with the features mean of \F set and \Out set for  multi-head \tinyimg with ResNet.}%
    \label{fig:tiny_multihead_featmean}%
    \end{figure*}

\begin{figure*}[h]
  \vspace*{-0.2cm}
    \centering
    \subfloat[{\footnotesize Features Mean with~ ER 25s}]{{\includegraphics[width=.49\textwidth]{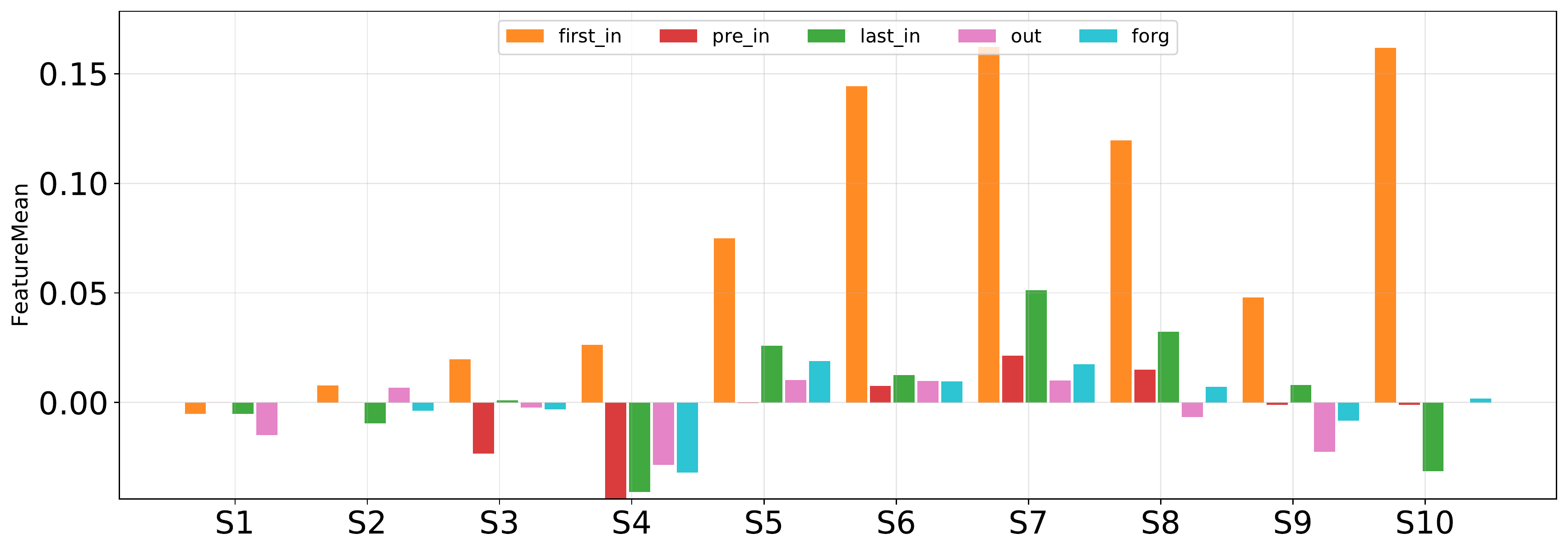}}}
        \hfill
    \subfloat[{\footnotesize Features Mean with~ER 50s}]{{\includegraphics[width=.49\textwidth]{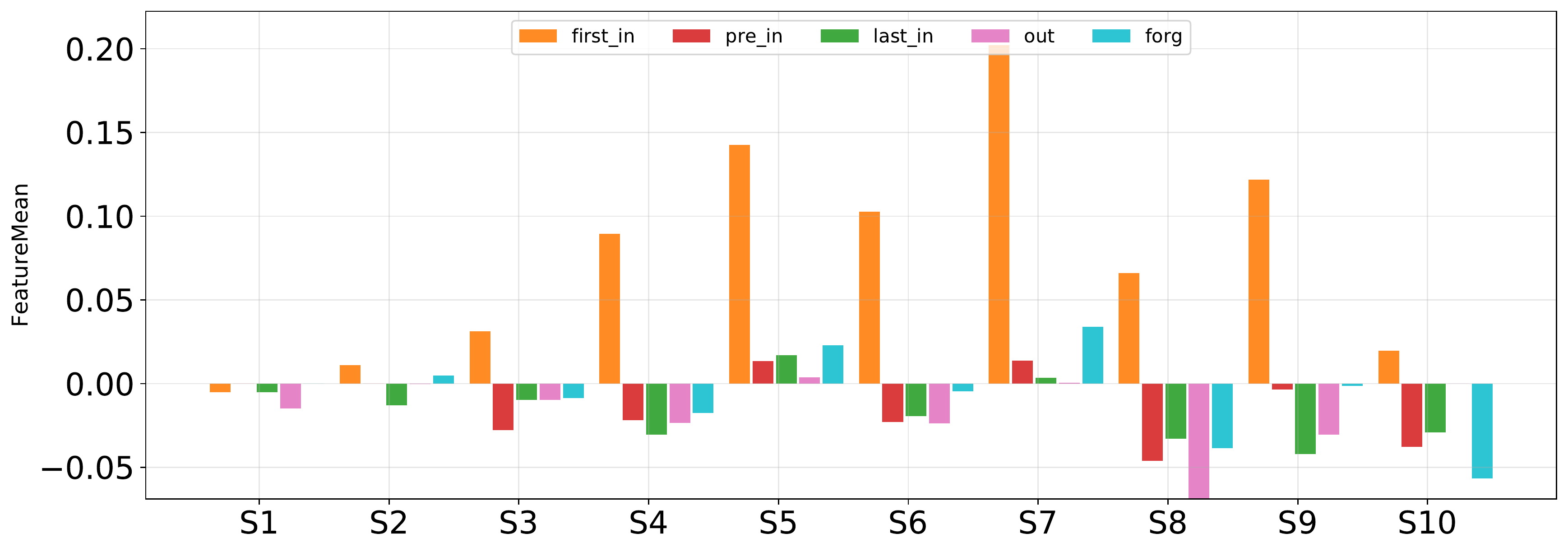}}}
\caption{\footnotesize Features mean of first \IN set, last \IN set and the in between \IN sets along with the features mean of \F set and \Out set for \ER~ under shared-head on \tinyimg with ResNet.}%
    \label{fig:tiny_sharedhead_featmean}%
    \end{figure*}

\begin{figure*}[h]
  \vspace*{-0.2cm}
    \centering
    \subfloat[{\footnotesize Features Mean with~ SSIL 25s}]{{\includegraphics[width=.49\textwidth]{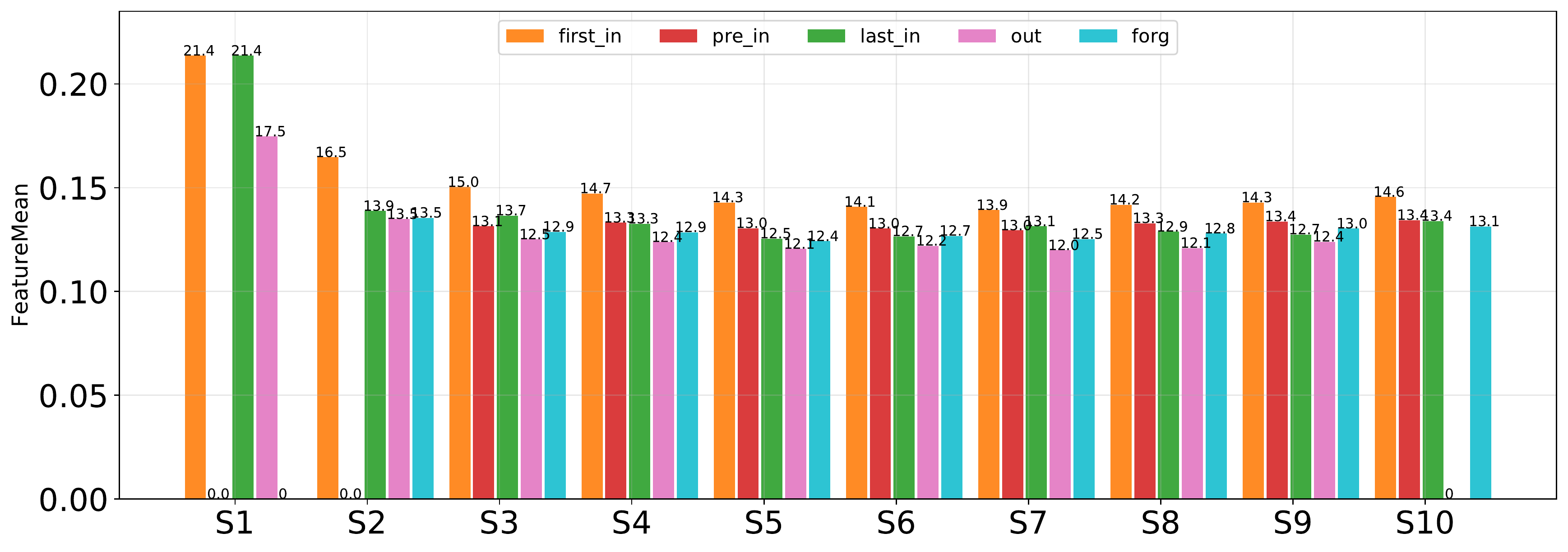}}}
        \hfill
    \subfloat[{\footnotesize Features Mean with~SSIL 50s}]{{\includegraphics[width=.49\textwidth]{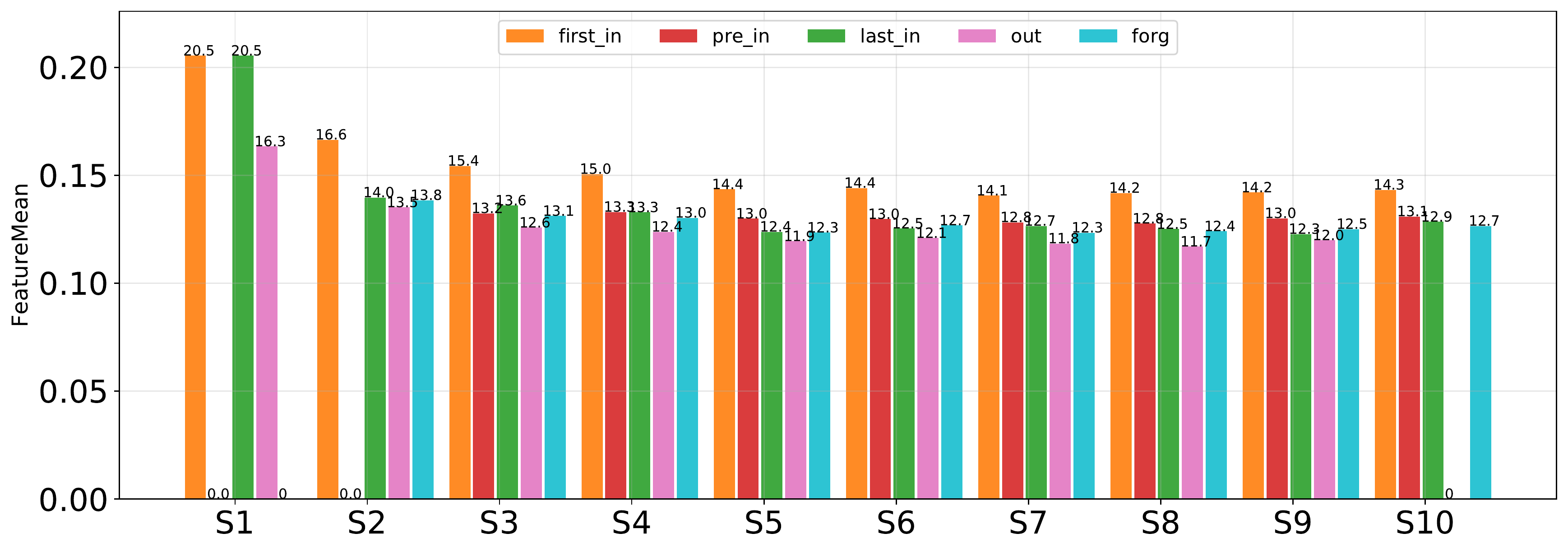}}}
\caption{\footnotesize Features mean of first \IN set, last \IN set and the in between \IN sets along with the features mean of \F set and \Out set for \SSIL~ under shared-head on \tinyimg with ResNet.}%
    \label{fig:tiny_sharedhead_featmean_SSIL}%
    \end{figure*}

\begin{figure*}[h]
  \vspace*{-0.2cm}
    \centering
    \subfloat[{\footnotesize Features Std. with~ \fine}]{{\includegraphics[width=.33\textwidth]{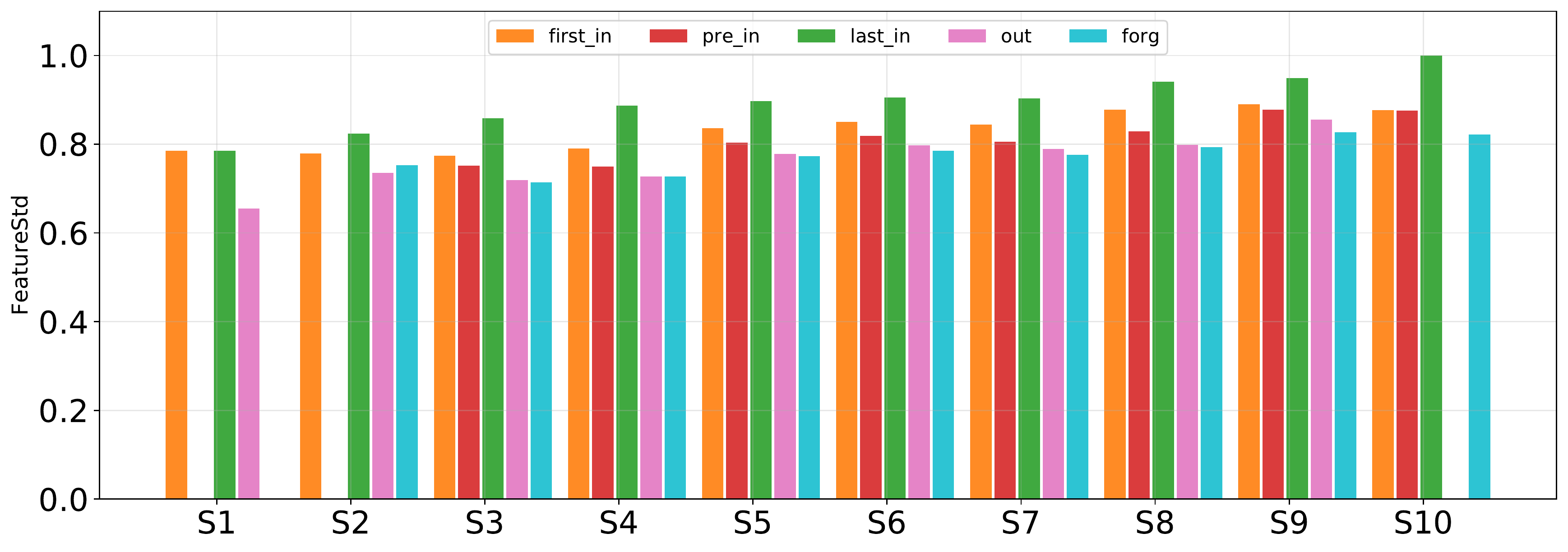} }}%
     \hfill
    \subfloat[{\footnotesize Features Std.~\LwF}]{{\includegraphics[width=.33\textwidth]{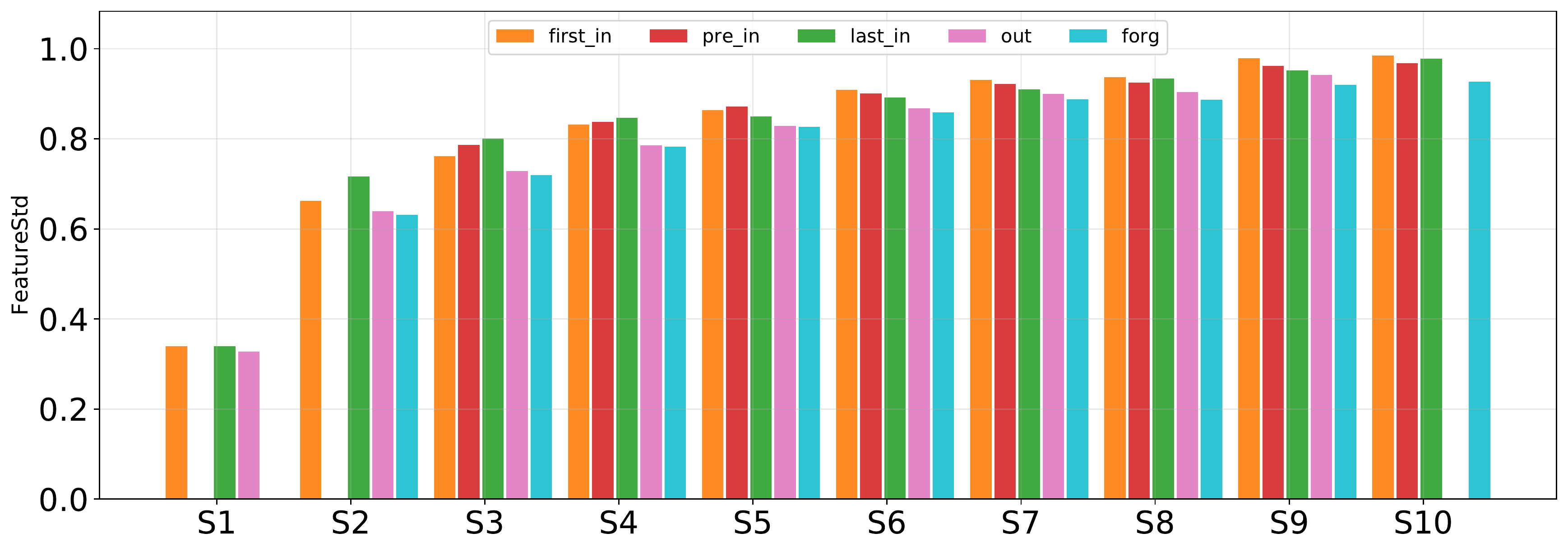}}}
        \hfill
    \subfloat[{\footnotesize Features Std.~ \MAS}]{{\includegraphics[width=.32\textwidth]{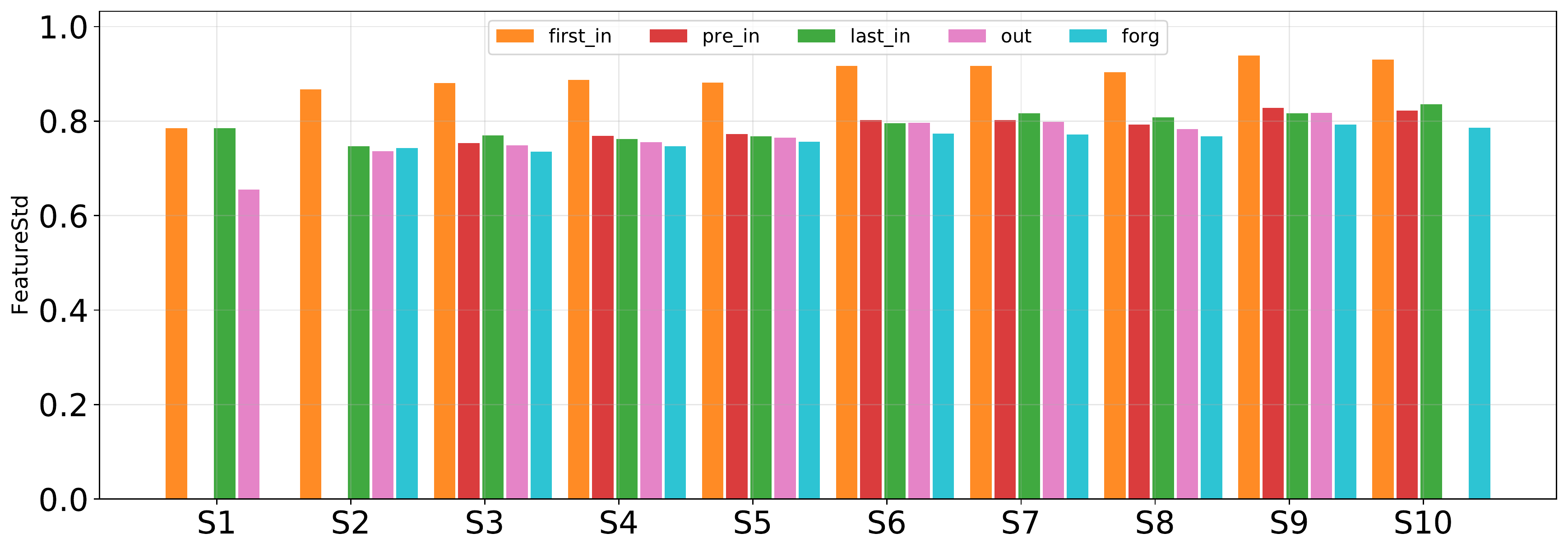}}}
\caption{\footnotesize Features std. of first \IN set, last \IN set and the in between \IN sets along with the features std. of \F set and \Out set for  multi-head \tinyimg with ResNet.}%
    \label{fig:tiny_multihead_featstd}%
    \end{figure*}

\begin{figure*}[h]
  \vspace*{-0.2cm}
    \centering
    \subfloat[{\footnotesize Features Std. with~ ER 25s}]{{\includegraphics[width=.49\textwidth]{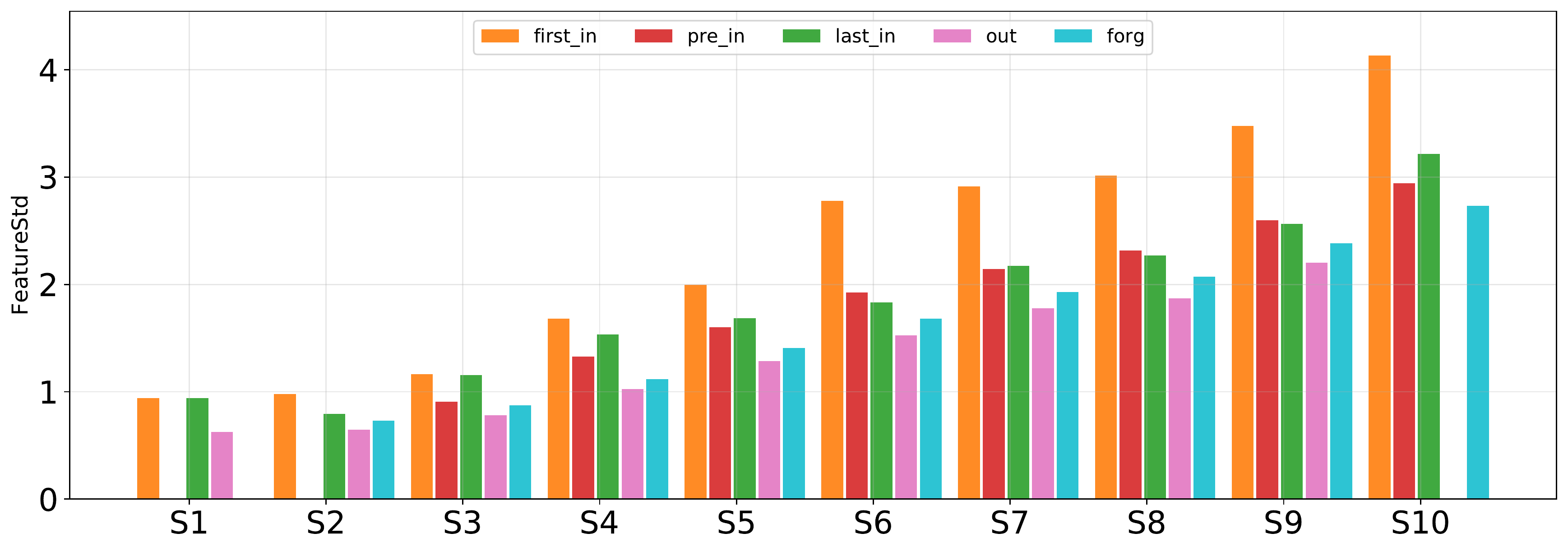}}}
        \hfill
    \subfloat[{\footnotesize Features Std. with~ER 50s}]{{\includegraphics[width=.49\textwidth]{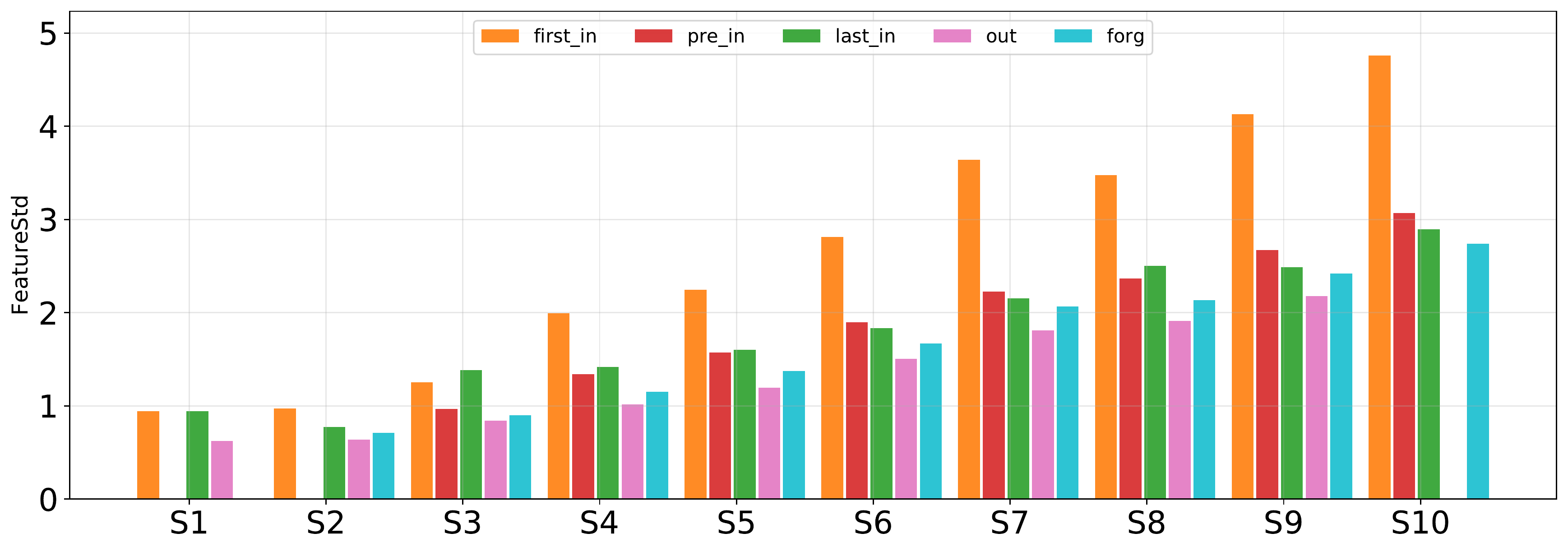}}}
\caption{\footnotesize Features std. of first \IN set, last \IN set and the in between \IN sets along with the features std. of \F set and \Out set for \ER~ under shared-head on \tinyimg with ResNet.}%
    \label{fig:tiny_sharedhead_featstd}%
    \end{figure*}

\begin{figure*}[h]
  \vspace*{-0.2cm}
    \centering
    \subfloat[{\footnotesize Features Mean with~ SSIL 25s}]{{\includegraphics[width=.49\textwidth]{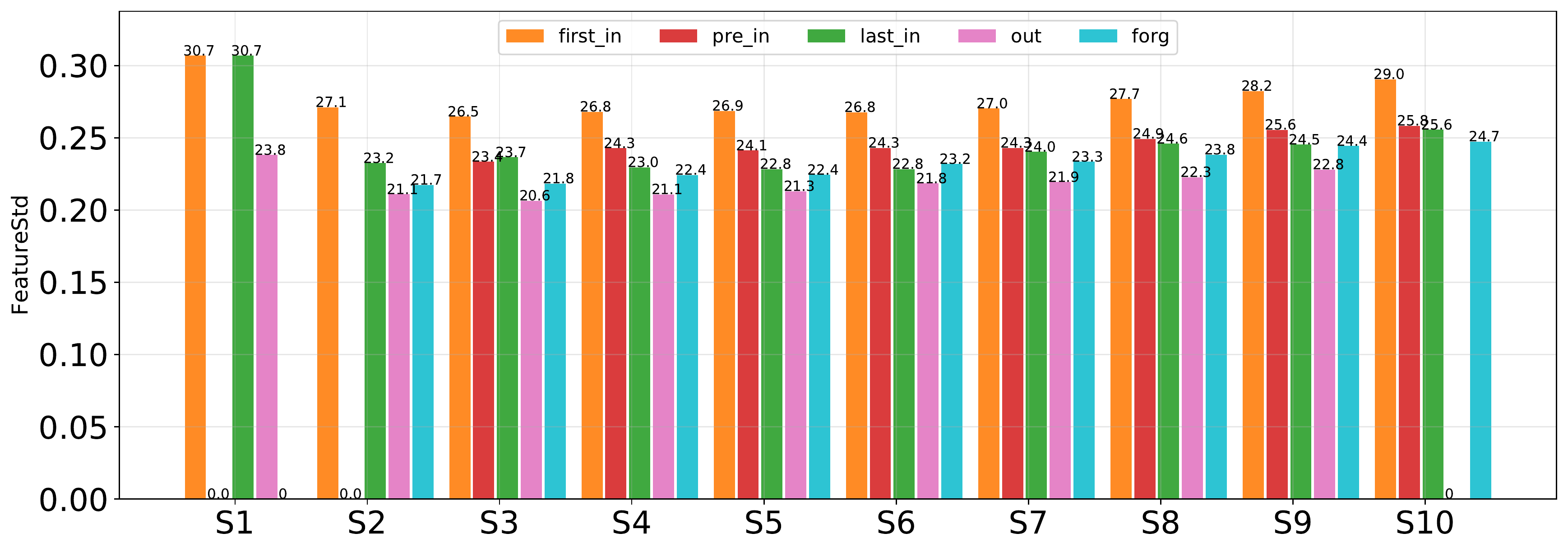}}}
        \hfill
    \subfloat[{\footnotesize Features Mean with~SSIL 50s}]{{\includegraphics[width=.49\textwidth]{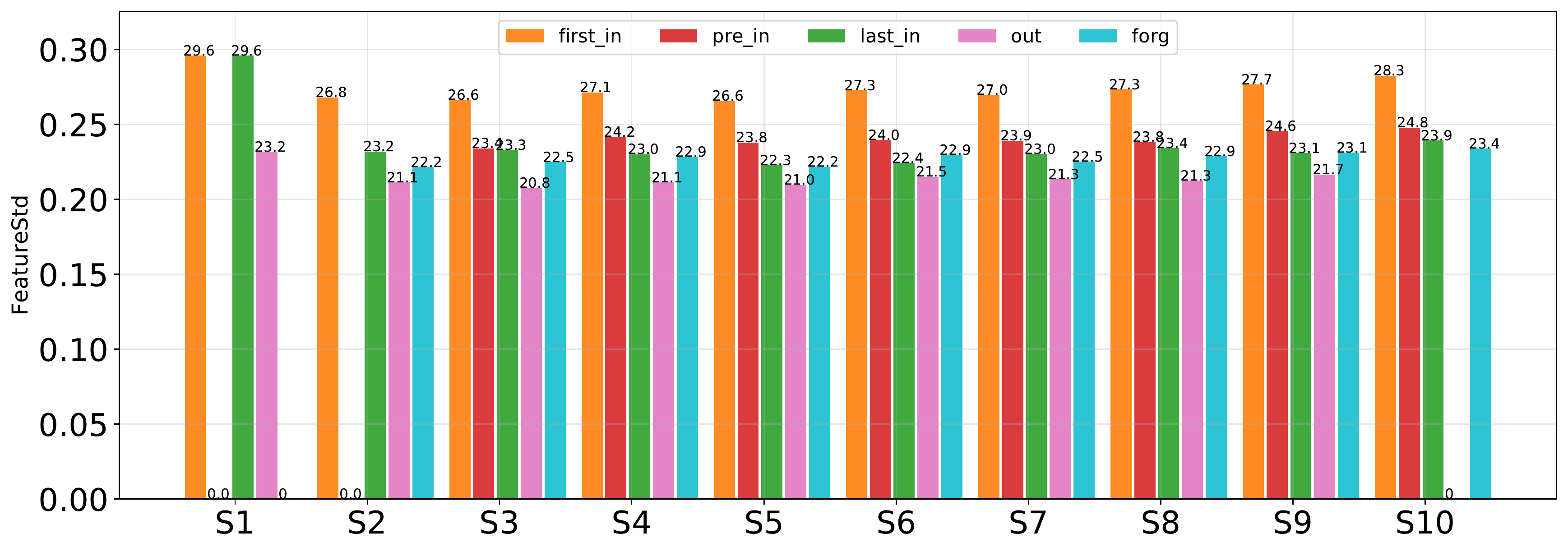}}}
\caption{\footnotesize Features std. of first \IN set, last \IN set and the in between \IN sets along with the features mean of \F set and \Out set for \SSIL~ under shared-head on \tinyimg with ResNet.}%
    \label{fig:tiny_sharedhead_featstd_SSIL}%
    \end{figure*}

    We have argued that the norm of the last layer features and the angle between the feature vector and the predicted class weight (prototype) are the main data dependent factors for a given sample prediction. We mention that the feature norm  differs significantly between the samples of the \IN set and those of the \Out set at the first learning stage, however, the difference diminishes as learning stages are passing. More specifically, \F samples tend to have smaller features norm than that of \IN samples. In Figure~\ref{fig:tiny_multihead_featnorm} and Figures (~\ref{fig:tiny_sharedhead_featnorm},~\ref{fig:tiny_sharedhead_featnorm_SSIL})  we report the average features norm of the different studied sets for \tinyimg with multi-head and shared-head (\ER~and \SSIL) settings respectively. We divided the concerned \IN sets into 3 groups: the first \IN set corresponding to the first learnt task, the last \IN set corresponding to the last learnt task and the remaining \IN sets corresponding to remaining previous tasks (pre-in). It can be seen that at the first stage the average features norm of the \IN set is higher than that of the \Out set in all considered settings. The features norm of the \F set tends to be smaller than \IN set features norm and similar or slightly larger than that of the \Out set in the considered settings. In all cases (except when deploying \LwF as a CL method) the features norm of the first \IN set samples remain higher than other previous \IN sets. When multi-head setting is considered, the features norm of previous \IN sets (pre-in) don't exhibit any significant differences with that of the \Out set. We also report the features mean   ( Figure~\ref{fig:tiny_multihead_featmean} and  Figures~\ref{fig:tiny_sharedhead_featmean},~\ref{fig:tiny_sharedhead_featmean_SSIL})) and features std. (Figure~\ref{fig:tiny_multihead_featstd},  Figures~\ref{fig:tiny_sharedhead_featstd},~\ref{fig:tiny_sharedhead_featstd_SSIL}) for~\tinyimg multi-head and shared-head (~\ER~ and ~\SSIL~) respectively. The std. of the features show similar trends to the features norm, however the mean is less expressive apart from the first \IN set large features mean in the shared-head setting.

\clearpage



